\documentclass[11pt]{article}

\usepackage[final]{acl}

\usepackage{times}
\usepackage{latexsym}

\usepackage[T1]{fontenc}

\usepackage[utf8]{inputenc}

\usepackage{microtype}

\usepackage{inconsolata}

\usepackage{graphicx}

\usepackage{booktabs}
\usepackage{multirow}
\usepackage{amsmath}
\usepackage{xcolor}
\usepackage{enumitem}
\usepackage{comment}

\newcommand{\todo}[1]{\textcolor{red}{#1}}
\newcommand{\arr}[1]{#1}

\newcommand{\mh}{MultiHiertt}
\newcommand{\fldqa}{FinLongDocQA}

\usepackage{pifont}

\newcommand{\cmark}{\textcolor{green!60!black}{\ding{51}}}
\newcommand{\xmark}{\textcolor{red!75!black}{\ding{55}}}
\usepackage[most]{tcolorbox}

\usepackage{array}
\newcolumntype{I}{!{\vrule width 0.7pt}}

\usepackage{algorithm}
\usepackage{algpseudocode}

\definecolor{boxaccent}{RGB}{0,150,136}   
\newtcolorbox{takeaway}[1][]{
  enhanced, breakable,
  colback=boxaccent!4,          
  colframe=boxaccent!70,        
  boxrule=0.5pt, arc=3pt,
  left=6pt, right=6pt, top=4pt, bottom=4pt,
  fonttitle=\bfseries, coltitle=boxaccent!80!black,
  #1
}
\title{A Table Is Worth 64 Tokens: Pixel-level Compression for Multi-Table Document Question Answering}

\author{
  Iñigo Alonso \and Mirella Lapata \\
  School of Informatics \\
  University of Edinburgh \\
  Edinburgh, UK \\
  \texttt{\{ialonso, mlap\}@ed.ac.uk}
}

\begin{document}
\maketitle
\begin{abstract}
Answering questions over real-world documents requires processing long
inputs that interleave text with tables.  Optical context compression,
which represents context as images, promises to reduce token cost, but
its effect on table understanding remains unclear. We study
pixel-level table compression for question answering over documents
with multiple tables, evaluating seven VLMs across two benchmarks and
five visual-token budgets.  \arr{At moderate budgets, table images
  match \arr{\textsc{html}} in accuracy with fewer tokens, but aggressive downscaling
  makes models reason longer without recovering accuracy, eroding the
  savings.}  Yet even at 64 visual tokens per table, tables preserve
enough signal to identify whether they are relevant to a question.  We
exploit this asymmetry with a training-free, two-stage method: the
model first \emph{identifies} the tables needed to answer a question
from a pixel-compressed context, and then \emph{reasons} over those
tables at native resolution.  On long financial documents, it uses
41\% fewer tokens than single-step QA over native-resolution tables
while gaining 5 accuracy points, and 21\% fewer than the most
efficient compressed configuration at no accuracy cost.  It also still
saves tokens when applied to retrieved context.\footnote{Code and data are available at \url{https://github.com/alonsoapp/A-Table-Is-Worth-64-Tokens}.}
\end{abstract}

\section{Introduction}
\label{sec:intro}

Real-world documents are long and heterogeneous, mixing free-form text
with structured elements such as tables.  Despite advances in
efficient attention, processing long
inputs remains computationally costly \citep{liu2025comprehensivesurveylongcontext}, and performance degrades as
contexts grow \citep{hsieh2024ruler}, especially when the relevant
evidence sits in the middle of the input \citep{liu-etal-2024-lost}. Representing the input
efficiently is therefore not just an optimisation goal but a
requirement for handling documents at scale.

\begin{figure*}[t]
  \centering
\includegraphics[width=.8\linewidth]{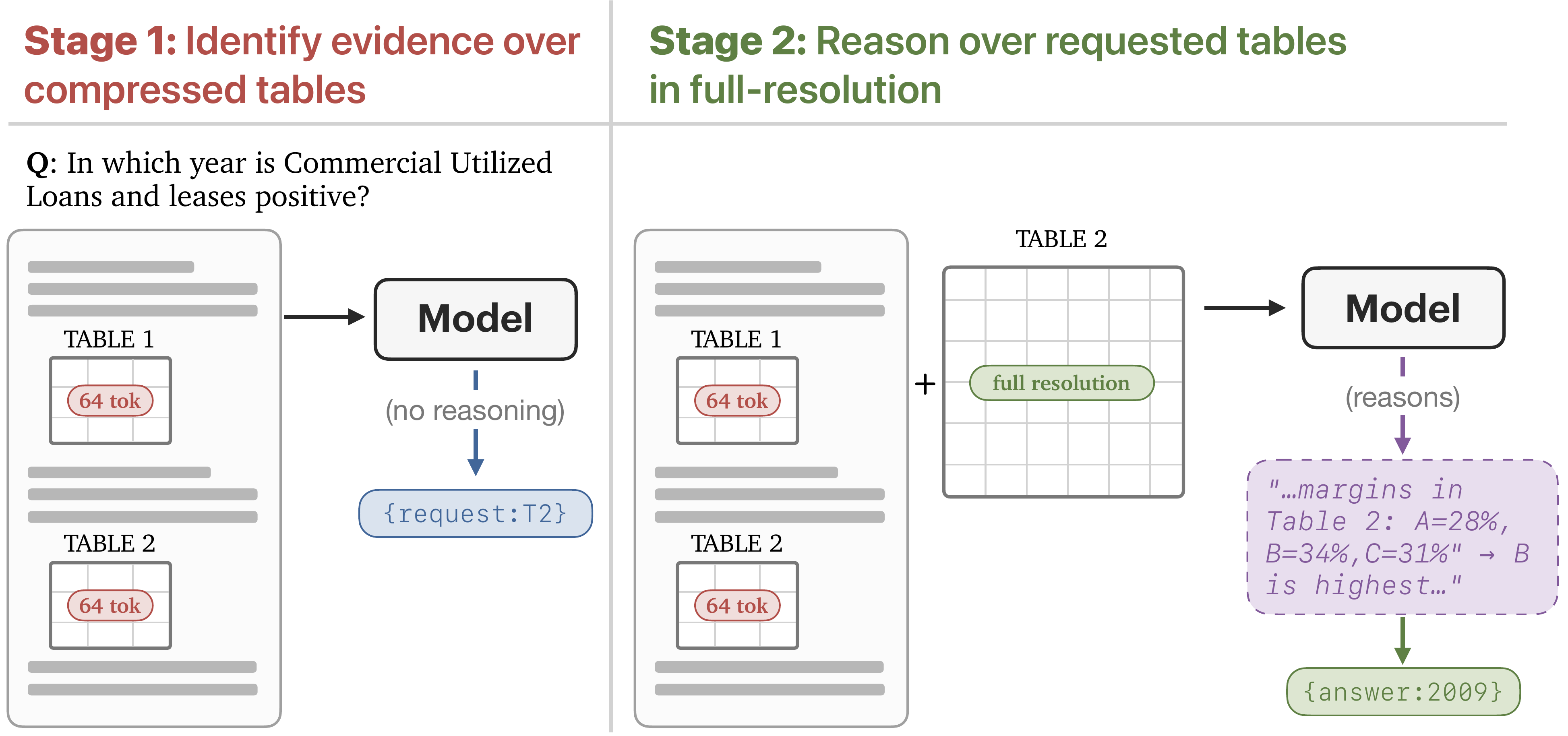}
  \caption{Our two-stage \emph{identify first, reason later}
    method. \textbf{Stage 1}: the model receives the document with
    every table compressed to a small visual-token budget and, instead
    of answering, identifies the tables it needs. \textbf{Stage 2}:
    the identified tables are supplied at native resolution and the
    model reasons over them to answer.}
  \label{fig:hero}
  \vspace{-10pt}
\end{figure*}
%

Recent work on optical context compression has shown that text
rendered as an image typically consumes fewer tokens than the same
content tokenised as text \citep{wei2025deepseek}. Originally
tested on OCR-style reconstruction, these findings have since been
extended to downstream tasks, both with dedicated training
\citep{xing2025vision, cheng-etal-2026-glyph} and with off-the-shelf
VLMs \citep{li-etal-2025-text}. This evidence, however, is limited to
plain text, which motivates our central question: \emph{can
  pixel-level table compression improve efficiency at no performance
  cost?}

To answer it, we focus on question answering (QA) over financial
documents \arr{(\mh{} and \fldqa{})}, a task representative of real-world challenges: long
context, heterogeneous information (text and tables), and questions
requiring numerical reasoning. We evaluate \arr{seven VLMs}
spanning \arr{fixed- and variable-length visual tokenisation}, under a
common framework of five visual-token budgets (64, 128, 256, 512, and
1,024 tokens per image), using the number of tokens processed by the
decoder backbone as a proxy for efficiency.  Our experiments show that
simply lowering the resolution is not enough: once cell values become
illegible, models reason longer and accuracy drops. Yet at the same
budgets, they can still tell which tables are needed to answer a
question.

Drawing on these findings, we propose a simple two-stage method (see
Figure~\ref{fig:hero}). The model first receives the document with
every table compressed to a given budget and, instead of answering
directly, identifies the relevant tables; these are then
supplied as images at native resolution, and the model reasons over them
to produce an answer. Concurrent work shows that decoupling evidence
identification from reasoning can improve long-context reasoning
\citep{guan2026evidence, zhao2026recontext}. We show that
compression makes the identification step cheap, since it preserves
the coarse information needed to locate relevant tables. Our
contributions can be summarised as follows:

\begin{itemize}[leftmargin=*,itemsep=2pt]
 \item We characterise the accuracy--efficiency trade-off
   of pixel-level table compression in multi-table document QA across
   \arr{seven} VLMs, two datasets, and five visual-token budgets,
   showing that on \mh{}, table images at moderate budgets match their
   \textsc{html} serialisation in accuracy while using fewer input tokens.

 \item We identify an asymmetry in how compression affects model
   behaviour. Lower resolutions degrade fine-grained reading and induce
   longer reasoning traces that offset the input-token savings,
   yet even heavily downscaled tables retain enough coarse signal for
   question-conditioned identification of the relevant tables.

 \item We exploit this asymmetry with a two-stage method that follows an \emph{identify first, reason later} recipe: it first
      identifies relevant tables in a compressed context and then
      reasons over only those tables at native resolution.
  \arr{On the long \fldqa{} documents,} it saves 41\% of total
   tokens while gaining \arr{5} accuracy points over single-step QA with
   native-resolution tables, and matches the accuracy of the most
   efficient single-step compressed configuration using \arr{21}\% fewer
   tokens.

 \item We situate the method against external retrieval. A dedicated
   retriever identifies evidence tables more accurately than a model
   reading a compressed context, but this advantage does not carry
   through to a better accuracy--efficiency trade-off; our
   method matches or exceeds retrieval at lower token cost, and continues
   to save tokens when applied on top of retrieved context.
\end{itemize}

\section{Related Work}
\label{sec:related}

\paragraph{Efficient Inference for Long Contexts.}
Long-context models remain costly and often struggle to use evidence
distributed across long inputs \citep{liu2025comprehensivesurveylongcontext,
liu-etal-2024-lost}. Complementary approaches reduce the context
processed in each forward pass through prompt compression
\citep{yoon-etal-2024-compact}, memory
\citep{chen2024generativeadapter}, agentic reading
\citep{zhang2024chainagents}, or retrieval
\citep{gunther2025latechunking}. \arr{Retrieval-augmented generation is
the dominant and most token-efficient of these, at the cost of building
an index per document and fixing the number of retrieved units in
advance; current retrievers index both text \citep{chen-etal-2024-m3}
and rendered document images \citep{faysse2024colpali}.} \arr{Concurrent} work shows that
separating evidence identification from reasoning improves
long-context performance \citep{guan2026evidence,zhao2026recontext}.
Our method adopts this separation but uses pixel compression to make
evidence identification over multi-table documents inexpensive; it
can be applied to either full or retrieved contexts.

\paragraph{Optical Context Compression.}
\arr{Representing language through rendered pixels is a well-established
line of work} \citep{rust2023pixels,lee2023pix2struct,lotz-etal-2023-text,
  gao2024improving}. A related line treats this representation as a
form of context compression, since visual encoders may represent
rendered text with fewer tokens \citep{wei2025deepseek}. This idea has
been extended to downstream tasks through specialised training
\citep{wang2024leveraging,xing2025vision,cheng-etal-2026-glyph} and,
more recently, with off-the-shelf VLMs
\citep{li-etal-2025-text}. Related approaches prune visual tokens
\citep{chen2024fastv,shang2024PruMerge,choi2026docprune}, but are
often tied to particular architectures. In contrast, we study
training-free pixel compression for structured, information-dense tables
across models with \arr{fixed- and variable-length visual tokenisation}.

\paragraph{Table Representation and Reasoning.}
Tables are commonly linearised for language models
\citep{herzig-etal-2020-tapas,wang2024chain,zou2025tattootoo},
e.g.,~in markdown and \textsc{html} \citep{sui2023table}, although
linearisation can impair structural reasoning over long tables
\citep{wang2025needleinatable}. Other work compares visual and textual
table representations
\citep{deng-etal-2024-tables,zheng-etal-2024-multimodal,
  alonso-etal-2024-pixt3,kim2024tablevqabench,
  singh-etal-2025-mtabvqa} or dynamically routes between them
\citep{xing2025tabledart,wu2025vtoolr1vlmslearnthink}.  \arr{Document
  VQA instead renders whole pages as images \citep{mathew2021docvqa}; we
  render each table separately, so that compression can be controlled
  per table while the surrounding text stays as text. Multi-table QA
  \citep{zhao-etal-2022-multihiertt,singh-etal-2025-mtabvqa} further
  requires locating the relevant tables among many, the step our
  method makes inexpensive.} Most closely related,
\citet{kwok2026tabqaworld} dynamically select visual or textual
representations for portions of a table during reasoning. Their method
targets latency in single-table, multi-turn QA at a fixed image
resolution; we instead study token efficiency under controlled
compression in documents containing multiple interleaved tables and
introduce a training-free method that separates table identification
from full-resolution reasoning.

\section{Experimental Framework}
\label{sec:framework}

\subsection{Datasets}
\label{sec:datasets}

\begin{table}[t]
  \centering
  \small
  \begin{tabular}{@{}l@{}rr@{}}
    \toprule
     & \textbf{\mh{}} & \textbf{\fldqa{}} \\
    \midrule
    Examples             & 1{,}044            & 400 \\
    Tables / doc         & 3 [2, 7]           & 80 [3, 122] \\
    Pages / doc          & 4 [2, 8]           & 100 [16, 156] \\
    Max. Tokens (\textsc{html})   & 12{,}264           & 127{,}816 \\
    Evidence tables      & 1 [0, 3]           & 2 [1, 8] \\
    Evidence pages       & 1 [1, 4]           & 2 [2, 3] \\
    Answer in Text       & 11.0\%             & 0.0\%  \\
    Answer in Table      & 32.4\%             & 100.0\%  \\
    Answer in Text and Table       & 56.6\%             & 0.0\%  \\
    Questions / doc      & 3 [2, 6]           & 1 [1, 6] \\
    \bottomrule
  \end{tabular}
  \caption{\label{tab:datasets} Statistics of the evaluation subsets (mode, with [min, max]
    intervals, except `Max. Tokens', which reports the true
    maximum\arr{, counted with the Qwen 3.5 tokeniser}). \fldqa{} annotates evidence at the page level; we treat
    every table on an evidence page as an evidence table. Answer in
    `Text/Table/Text and Table' gives the percentage of questions
    answerable from text alone, from a table alone, or both\arr{; on
    \fldqa{}, all questions are answered from tables because we sample
    from its \emph{table} category}.}
    \vspace{-10pt}
\end{table}

Existing table-understanding benchmarks \citep{2019TabFactA,
  parikh-etal-2020-totto, kim2024tablevqabench, alonso2026tablet}
often present tables in isolation, whereas real-world document QA
requires models to identify and reason over relevant tables embedded
within surrounding text. Our study also requires tables that can be
represented faithfully as both text and images, so that differences
between modalities reflect the representation rather than differences
in content. We therefore focus on benchmarks that contain multiple
tables interleaved with text and provide tables in a structured form
that can be rendered into images with content parity. \arr{Rendering
  \textsc{html} tables into images preserves their content exactly,
  whereas serialising image-sourced tables into text is costly and
  error-prone.}

These criteria lead us to two benchmarks: \mh{}
\citep{zhao-etal-2022-multihiertt} and \fldqa{}
\citep{wang2026finlongdocqa}. Both combine multiple tables with 
surrounding text, but differ substantially in scale. We use \mh{} as a
\emph{short-context setting}, with typically three tables and four
pages per document, and \fldqa{} as a \emph{long-context setting},
with typically~80 tables and 100~pages per document. Dataset
statistics are shown in Table~\ref{tab:datasets}.

\mh{} test labels are private, so we evaluate on the development set.
Although current models may have encountered this split during
pre-training, our analysis compares configurations within each model
rather than absolute performance across models. For \fldqa{}, we
evaluate 400 randomly sampled questions from the \emph{table}
category, ensuring that the answer evidence is grounded in tables.%
\footnote{At the time of dataset selection, some evidence annotations
in \fldqa{} were incorrect. We therefore restricted evaluation to the
dataset's \emph{table} category, where questions are explicitly
designed to be answered from tables. The annotations were subsequently
corrected by the dataset creators, and we use these
for all reported results\arr{; the errors affected only evidence
annotations, not questions or answers}.} \arr{Documents reach 715k
tokens, so evaluating the full test set across all our configurations
would be prohibitively expensive.} We sample from documents whose
\textsc{html} serialisation
fits within 128k tokens under the Qwen 3.5 tokeniser, allowing all
configurations to be evaluated within the models' supported context
windows\arr{; this excludes the longest documents in the dataset}.

\subsection{Input Settings}
\label{sec:settings}

We compare two \emph{input settings} that differ only in how tables
are represented; the document, question, and the prompt are
always provided as text.

\paragraph{Text (\textsc{html}).}
Tables are serialised as \arr{\textsc{html}}, which is a strong textual representation that
preserves hierarchical structure, including multi-level headers and
spanning cells \citep{sui2023table}. This setting serves as the
performance reference for both accuracy and token usage.
\arr{We additionally report results with Markdown in
Appendix~\ref{sec:app-markdown}.}

\paragraph{Hybrid (Text $+$ Images).}
Document text is provided as text, while tables are rendered as images
from the same \arr{\textsc{html}} used in the textual setting. This gives the two
modalities identical content and structure, allowing us to isolate the
effect of representation. \arr{All experiments in which tables are
represented as images, including our two-stage method and the
single-step baselines, are conducted in this setting.} Tables are rendered in black and white using
a uniform typeface.%
\footnote{Table images were rendered using Firefox in headless mode
(version 142.0.1, GeckoDriver 0.36.0) at an effective density of 96
pixels per inch.}  

\subsection{Models}
\label{sec:models}

\paragraph{Fixed-length Visual Tokenisation.}
Gemma\,4 \citep{gemmateam2026gemma4} represents each image using a
fixed number of visual tokens selected from a discrete grid: 64, 121,
256, 529, or 1{,}089 tokens, independent of the image's dimensions.
We evaluate Gemma\,4 E4B and Gemma\,4 26B-A4B, with context windows of
128k and 256k tokens, respectively. The latter is a mixture-of-experts
model with 4B active parameters (see Table~\ref{tab:models}).

\paragraph{Variable-length Visual Tokenisation.}
Qwen 3 VL \citep{bai2025qwen3vltechnicalreport} and Qwen 3.5 \citep{qwen35blog} represent
images with a number of visual tokens that scales with their pixel
area. Qwen 3 VL pairs a Qwen~3 backbone with a SigLIP\,2 encoder
\citep{tschannen2025siglip2} \arr{further trained} at dynamic input
resolutions; Qwen 3.5 is natively multimodal, pre-trained end-to-end on
text interleaved with visual data, but retains an encoder of the same
family. In both models, a visual token corresponds to a $32 \times
32$ pixel region, determined by the encoder's patch size ($16$) and
spatial merge factor ($2$)\arr{,} so an image of size $h \times w$ yields
$\lceil h/32 \rceil \times \lceil w/32 \rceil$ tokens. We evaluate
the Instruct and Thinking variants of Qwen 3 VL-8B and 
Qwen 3.5-9B. All three configurations support 256k-token contexts. Table~\ref{tab:models} summarises our models; for the
precise mapping from image dimensions to visual-token budgets, see Section~\ref{sec:budgets}.

\paragraph{Additional Models.}
\arr{To test whether our findings extend beyond these two families, we
  also evaluate GLM-4.6V
  \citep{vteam2025glm45vglm41vthinkingversatilemultimodal}, an
  open-weight mixture-of-experts model with 106B total and 12B active
  parameters, and Gemini 3.8 Flash \citep{gemini38flash2026}, a proprietary model
  accessed through its API. \mbox{GLM-4.6V} represents images with a
  variable number of visual tokens. Gemini's architecture is
  undisclosed, but, as with Gemma, its API sets the visual tokens of
  each image by a resolution level (280, 560, or 1{,}120 tokens)
  rather than by the image's dimensions.}  \arr{Three models are
  evaluated on \mh{} only (Appendix~\ref{sec:app-all-results}). The
  128k context windows of Gemma\,4 E4B and GLM-4.6V cannot fit
  \fldqa{} documents alongside reasoning traces, and Qwen 3 VL-8B is
  considerably more expensive to run at this length.}

\begin{table}[t]
\centering
\small
\resizebox{\linewidth}{!}{%
\begin{tabular}{@{}l@{~}c@{~}c@{~}c@{~}c@{}}
\toprule
\textbf{Model} & \arr{\textbf{Params}} & \textbf{Tokenisation} & \textbf{Context} &
\textbf{Reasoning} \\
\midrule
Gemma\,4 E4B       & \arr{4B\textsuperscript{e}} & Fixed       & 128k & \cmark \\
Gemma\,4 26B-A4B   & \arr{26B (4B)} & Fixed       & 256k & \cmark \\
Qwen 3 VL-8B Instruct & \arr{8B} & Variable & 256k & \xmark \\
Qwen 3 VL-8B Thinking & \arr{8B} & Variable & 256k & \cmark \\
Qwen 3.5-9B         & \arr{9B} & Variable & 256k & \cmark \\
\arr{GLM-4.6V}            & \arr{106B (12B)} & Variable    & 128k & \cmark \\
\arr{Gemini 3.8 Flash}    & \arr{---} & Fixed & 1M   & \cmark \\
\bottomrule
\end{tabular}}
\caption{The \arr{seven} \arr{VLMs} we evaluate. \arr{\emph{Params}
    gives total parameters, with active parameters in parentheses for
    mixture-of-experts models (\textsuperscript{e}effective
    parameters); Gemini's are undisclosed.} \emph{Tokenisation} is
  whether the visual encoder emits a fixed or variable number of
  visual tokens per image; \emph{Context} is the supported
  context-window length; \emph{Reasoning} (\cmark/\xmark) marks
  test-time reasoning mode.}
\vspace{-15pt}
\label{tab:models}
\end{table}



\subsection{\arr{Visual-Token Budgets}}
\label{sec:budgets}

To compare compression across visual encoders, we define five nominal
visual-token budgets for each table image: $B \in \{64, 128, 256, 512,
1{,}024\}$. \arr{Budgets are a per-table cap, applied to every table
image in the document.} For the Qwen models, \arr{if an image's token
count (Section~\ref{sec:models}) exceeds $B$}, we downscale the image,
preserving its aspect ratio, to the largest resolution whose token count
does not exceed the budget; images already below the budget are not
upscaled. For Gemma\,4, which supports only a discrete set of
visual-token configurations, we map the nominal budgets to 64, 121,
256, 529, and 1{,}089 tokens, respectively.
\arr{GLM-4.6V is downscaled in the same way as the Qwen models,
according to its own patch size.} \arr{For Gemini,
whose API exposes only fixed resolution levels, we map the three
largest nominal budgets to its 1{,}120, 560, and 280 tokens per image;
its lowest level therefore lies above our two most aggressive budgets.}
All reported token counts use the actual number of
visual tokens entering the language decoder rather than the nominal
budget. \arr{We refer to the largest budget (for Gemma and Gemini, their
largest configuration) as \emph{native resolution}, as it amounts to no
compression on our datasets.}

We measure efficiency by the total number of tokens processed by the
language decoder: textual input tokens, visual tokens, and generated
tokens. Counting both input and output captures not only the savings
from compressing tables but also any additional test-time reasoning
induced by compression. Decoder-token count provides a common measure
across model families, although it does not include the cost of visual
encoding\arr{; wall-clock measurements (Appendix~\ref{sec:app-runtime})
confirm that our method is consistently faster than single-step QA
for Qwen 3.5-9B on \mh{}}.

\arr{For our two-stage method (Section~\ref{sec:method}), the second stage
continues the first-stage conversation, so tokens shared by both
stages are counted once, as under the prefix caching that serving
systems such as vLLM apply automatically
(Appendix~\ref{sec:app-token-accounting}).}


\section{Does Pixel-Level Compression Affect Table Comprehension?}
\label{sec:probing}

We examine how pixel-level compression affects two capabilities
required for table question answering: fine-grained reading and
downstream reasoning. We first use transcription to test whether
models can recover table structure and individual cell values, and
then evaluate whether compressed tables remain useful for answering
numerical questions. For each configuration, we plot task performance
against the median number of input or generated tokens across
examples. Later sections combine these quantities into total token
cost.

\subsection{Table Transcription}
\label{sec:ocr}

Table transcription tests whether the model can read every detail of a
table, including its structure and all cell values, without requiring
reasoning beyond format conversion. Models receive the full document,
including its multiple tables and surrounding text, and are asked to
transcribe every table into \arr{\textsc{html}}\arr{, without test-time
reasoning}. We evaluate the similarity
between generated and reference tables using TEDS
\citep{zhong2020image}\arr{, a tree-edit-distance similarity in~$[0,1]$
that reflects both structure and cell content.}

We compare three settings. In \textsc{html}$\rightarrow$\textsc{html},
models copy the original \arr{\textsc{html}}, providing a control for the generation
task itself. In \textsc{latex}$\rightarrow$\textsc{html}, models
convert \arr{a \LaTeX{} serialisation of the same table into \textsc{html}}, allowing us to measure
the difficulty of format conversion without visual
perception. Finally, \textsc{img}$\rightarrow$\textsc{html} evaluates
transcription from table images across the five visual-token
budgets. An example is provided in
Appendix~\ref{sec:ocr-example}.

Figure~\ref{fig:ocr} shows TEDS against input tokens. The
\textsc{html}$\rightarrow$\textsc{html} control is nearly perfect,
confirming that copying is not the bottleneck.
\textsc{latex}$\rightarrow$\textsc{html} already incurs a measurable
drop \arr{for most models}, showing that part of the transcription error arises from format
conversion rather than visual perception. Transcription from images
\arr{changes little down to 256 tokens but drops sharply below it},
indicating that \arr{heavily} compressed representations
lose fine-grained information about table structure and cell values.

For Gemma\,4, the largest visual-token setting also substantially
increases input cost without improving over the 512-token budget\arr{:
Gemma\,4 26B-A4B uses 50\% more input tokens at 1{,}024 than at 512
for similar TEDS}. 

\begin{figure}[t]
  \centering
  \includegraphics[width=0.88\linewidth]{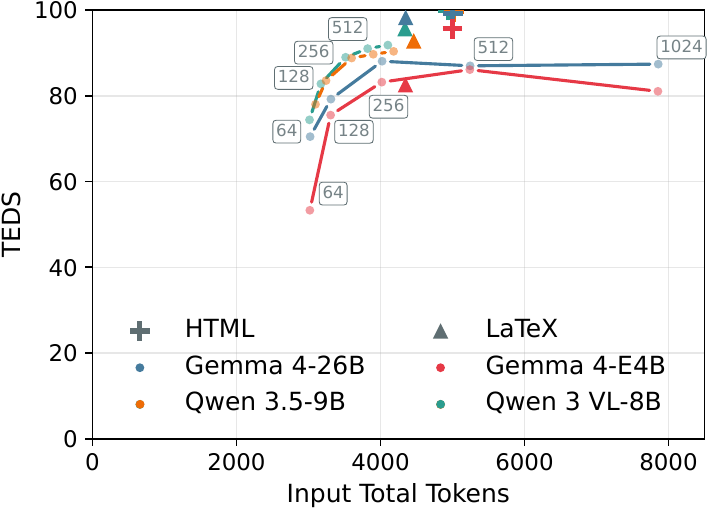}
  \caption{
  Table transcription performance against input tokens on \mh{}.
  The \textsc{html}$\rightarrow$\textsc{html} control is nearly
  perfect, while \textsc{latex}$\rightarrow$\textsc{html} shows that
  format conversion introduces error even without visual perception.
  Transcription from images degrades progressively as the visual-token
  budget decreases.}
  \label{fig:ocr}
 \vspace{-10pt}
\end{figure}

\subsection{Question Answering}
\label{sec:qa}

Reduced transcription fidelity does not necessarily imply lower
downstream performance: answering a question may require reading only
a small portion of a table rather than reconstructing it in full. We
therefore evaluate both input settings
(Section~\ref{sec:settings}) on the \mh{} and \fldqa{} \arr{datasets}. In this analysis, the model receives
the complete document; contexts assembled by an external retriever
are considered in Section~\ref{sec:rag}. Because both datasets emphasise
numerical reasoning, we enable test-time reasoning in the main
experiments and report results without it in
Appendix~\ref{sec:app-nothink}.

\begin{figure}[t]
  \centering
  \includegraphics[width=0.5\linewidth]{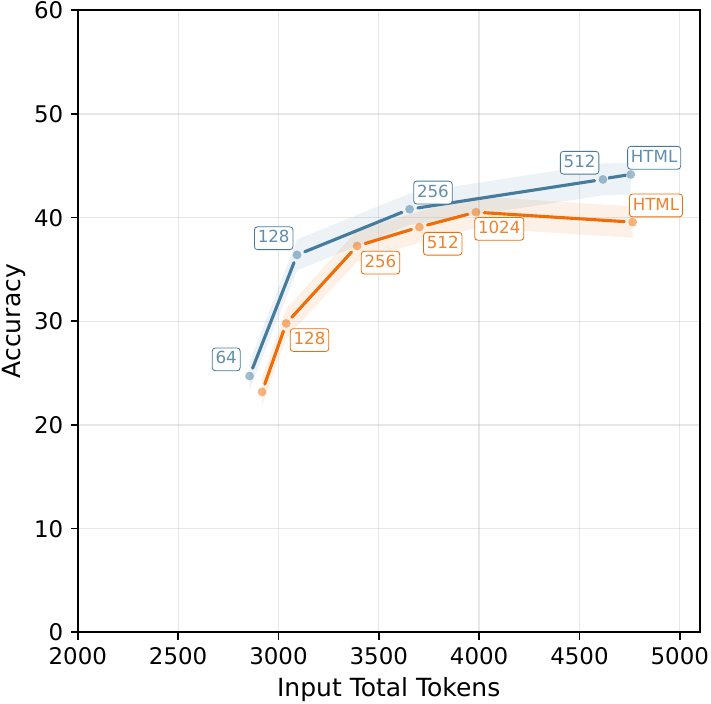}%
  \includegraphics[width=0.5\linewidth]{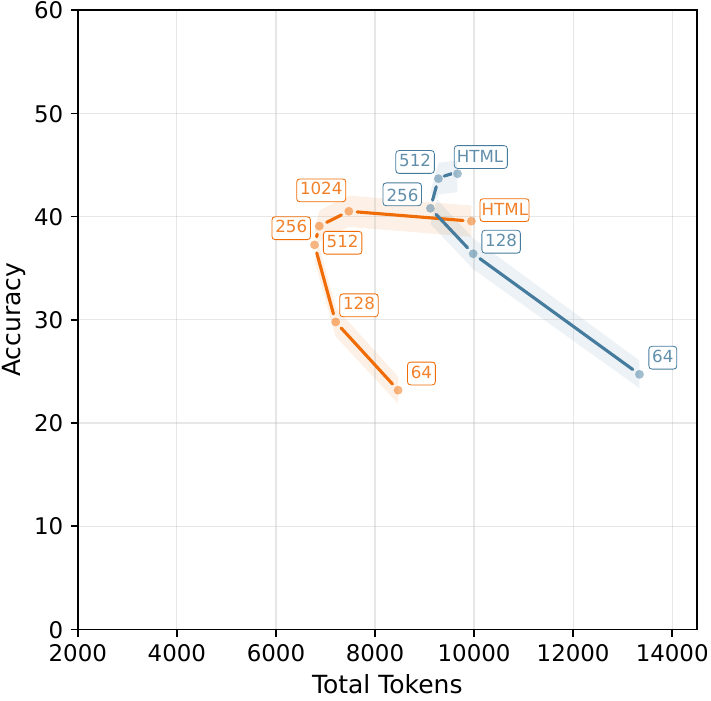}

  \textcolor[HTML]{457b9d}{$\bullet$} \small{Gemma\,4 26B-A4B} \hspace{2em}
  \textcolor[HTML]{EF6C06}{$\bullet$} \small{Qwen 3.5-9B}
  
  \caption{QA accuracy on \mh{} (test-time reasoning enabled) for
  Gemma\,4 26B-A4B and Qwen 3.5-9B, plotted
  against median input tokens (left) and generated tokens (right). At large budgets,
  table images match \arr{\textsc{html}} accuracy (left) using fewer input tokens; stronger
  compression reduces accuracy and produces longer reasoning traces (right).
  Results for all models and \fldqa{} are in
  Appendix~\ref{sec:app-all-results}.}
  \label{fig:qa}
  \vspace{-10pt}
\end{figure}

Figure~\ref{fig:qa} reports QA accuracy against input tokens and
generated tokens on \mh{} for the \arr{largest Gemma and Qwen models we evaluate}.
At sufficiently large visual-token budgets, tables represented as
images match the accuracy of their \arr{\textsc{html}}
serialisation \arr{with fewer input tokens (Qwen 3.5-9B at 1{,}024
tokens, Gemma\,4 26B-A4B at 512)}. Below this point,
however, accuracy generally falls as the token budget tightens. This
pattern is consistent with the transcription results: once compression removes
the detail needed to read cell values reliably, models can
no longer perform the numerical operations required to answer the
question.

Generated tokens reveal a second, less visible cost of compression:
lower visual-token budgets produce longer reasoning traces \arr{than at
native resolution}, without improving performance. Inspection of
the traces shows repeated misreadings of cell values and uncertainty
over illegible entries, causing models to deliberate for longer. This pattern also holds for \fldqa{}, although its much larger number of tables means that input-token
savings still outweigh the additional generation. Even so, measuring
input tokens alone overstates the efficiency gains of aggressive
compression.

Taken together, the results expose an important limitation of direct
QA over compressed tables. Moderate compression can provide useful
input savings without sacrificing accuracy, but aggressive
compression removes the fine-grained information needed for reasoning
and induces additional generation that offsets those savings.
This raises a different possibility: heavily compressed tables may
still preserve enough coarse information to identify which tables are
relevant, even when they are no longer legible enough to answer from
directly. We test this possibility in the next section.

\setcounter{topnumber}{1}

\begin{figure}[t]
  \centering
  \includegraphics[width=1.0\linewidth]{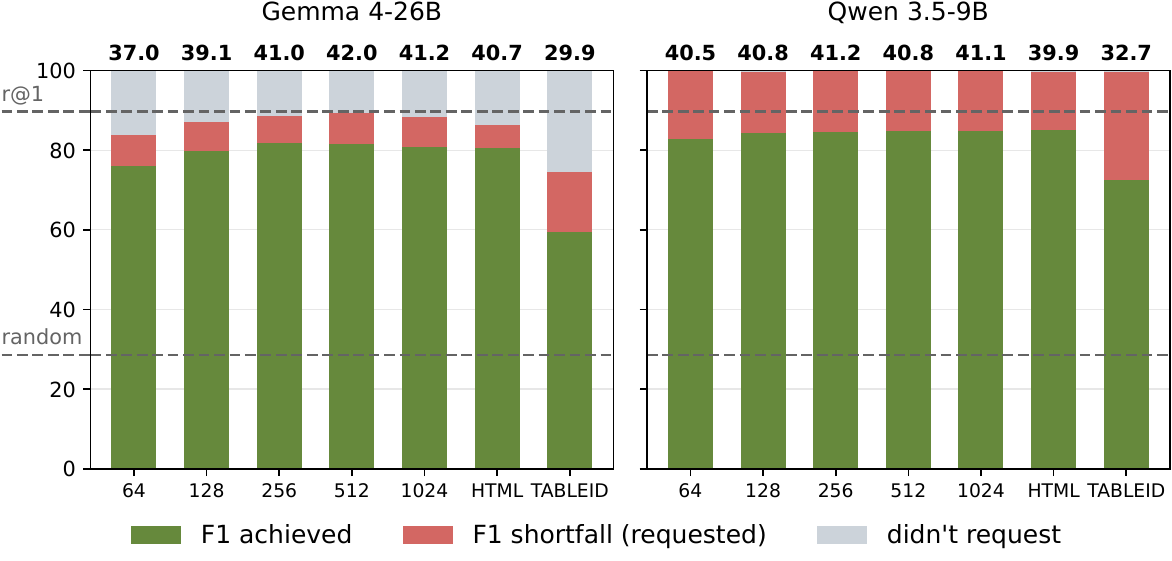}
 \caption{Relevant-table identification (F1) for \mh{} across token budgets and table representations (\arr{Gemma\,4 26B-A4B}, Qwen 3.5-9B). Bars (y~axis) decompose F1 into achieved vs.\ shortfall on requested tables; \textsc{html} and \textsc{tableid} are references; the horizontal lines mark random choice and ColQwen2@1, its best-F1 operating point. The percentage above each bar is the overall downstream QA accuracy, so identification gains carry through to the downstream task, not just retrieval (see  Appendix~\ref{sec:app-all-results} for more detail).
 }
 \vspace{-10pt}
  \label{fig:qa1}
\end{figure}

\begin{figure*}[t]
  \centering
  \setlength{\tabcolsep}{0pt}
  
  \begin{tabular}{@{} c c c c c @{}}
    & \multicolumn{2}{c}{Full-context} & \multicolumn{2}{c}{BGE-M3} \\
    
    & \small{Qwen 3.5-9B} & \small{Gemma\,4 26B-A4B} & \small{Qwen 3.5-9B} & \small{Gemma\,4 26B-A4B} \\[1mm] 
    
    \raisebox{-0.5\height}{\rotatebox{90}{MultiHiertt}} \hspace{2pt} \raisebox{-0.5\height}{\rotatebox{90}{\tiny Accuracy}} &
    \hspace{-2pt}\raisebox{-0.5\height}{\includegraphics[width=0.23\linewidth]{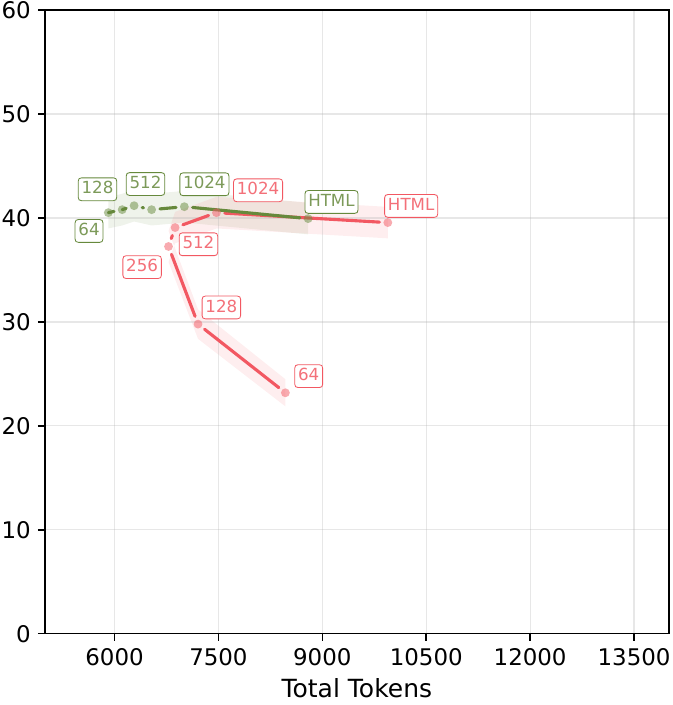}} &
    \hspace{-2pt}\raisebox{-0.5\height}{\includegraphics[width=0.23\linewidth]{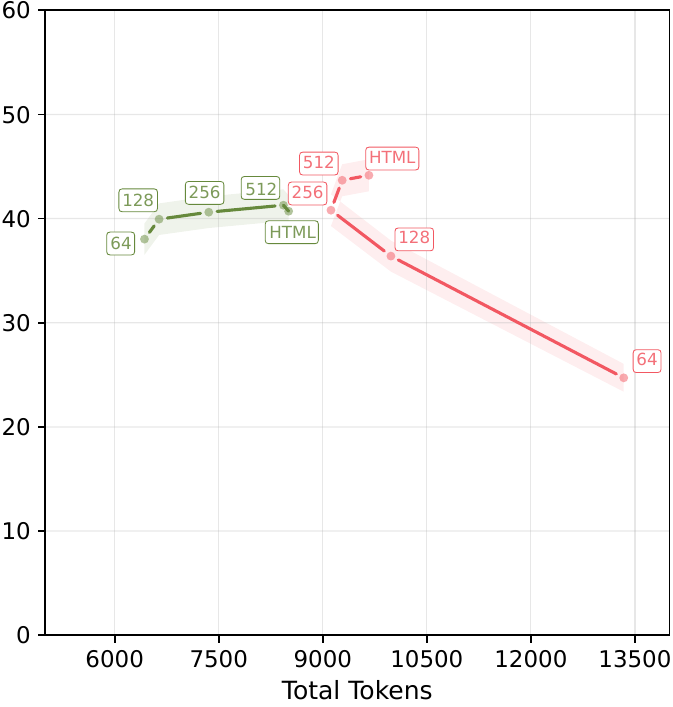}} &
    \raisebox{-0.5\height}{\includegraphics[width=0.24\linewidth]{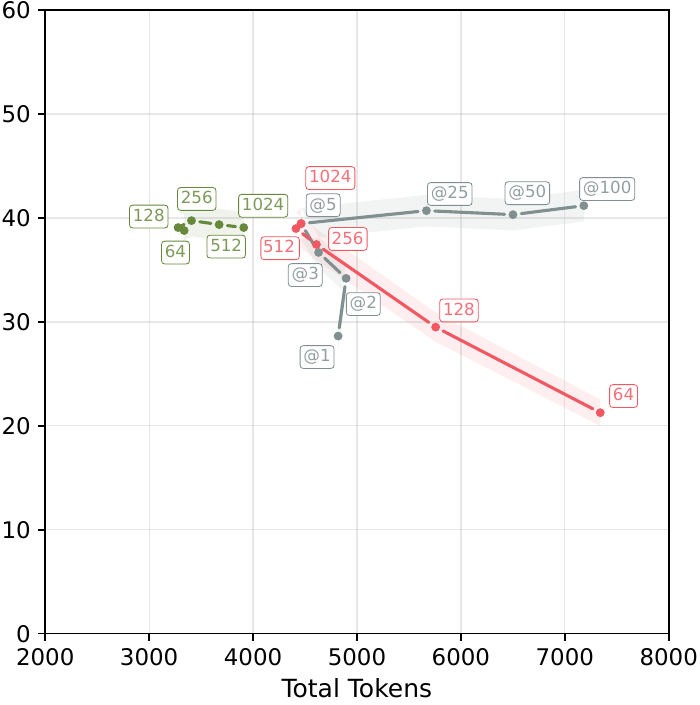}} &
    \raisebox{-0.5\height}{\includegraphics[width=0.24\linewidth]{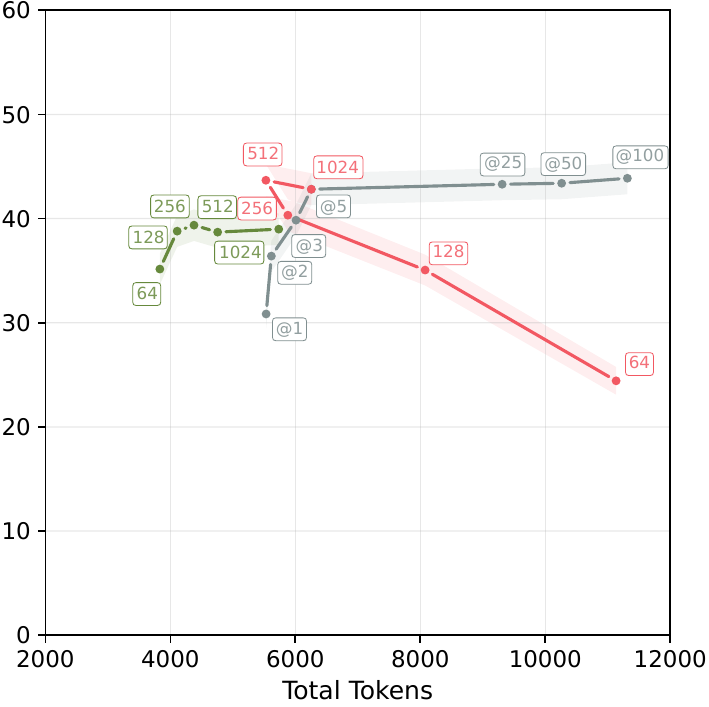}} \\
    
    \raisebox{-0.5\height}{\rotatebox{90}{FinLongDocQA}} \hspace{2pt} \raisebox{-0.5\height}{\rotatebox{90}{\tiny Accuracy}} &
    \raisebox{-0.5\height}{\includegraphics[width=0.24\linewidth]{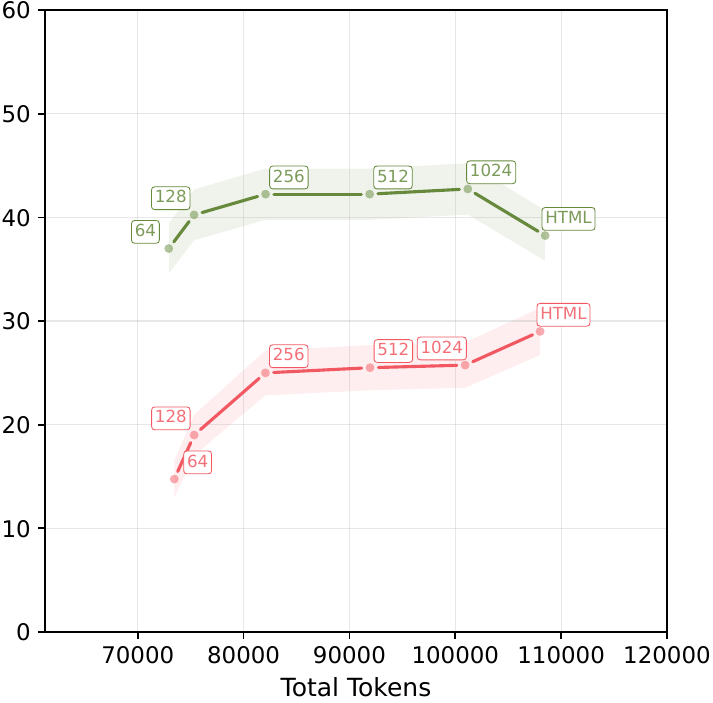}} &
    \raisebox{-0.5\height}{\includegraphics[width=0.24\linewidth]{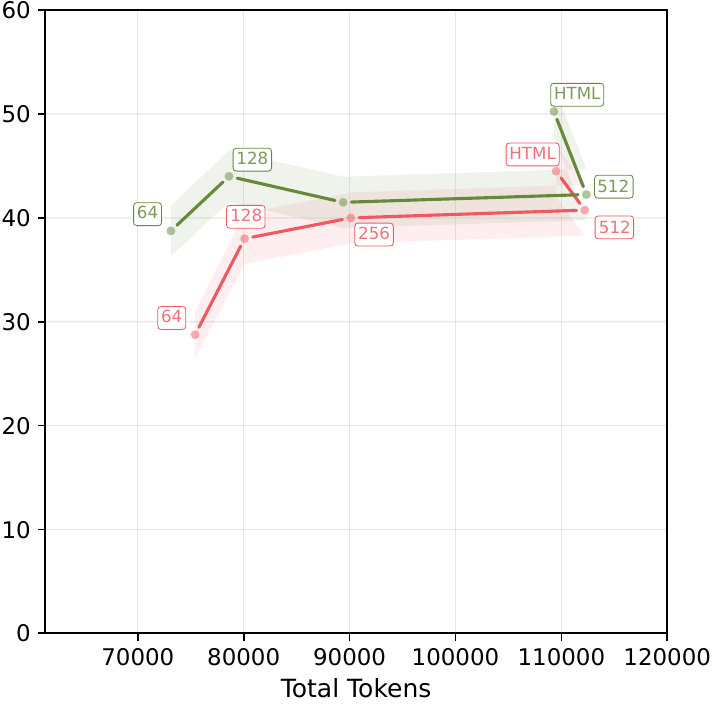}} &
    \raisebox{-0.5\height}{\includegraphics[width=0.24\linewidth]{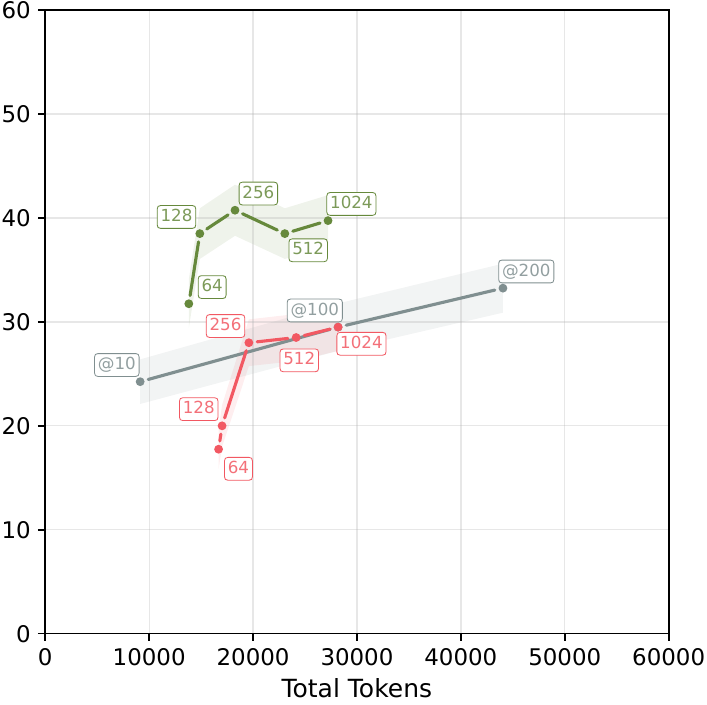}} &
    \raisebox{-0.5\height}{\includegraphics[width=0.24\linewidth]{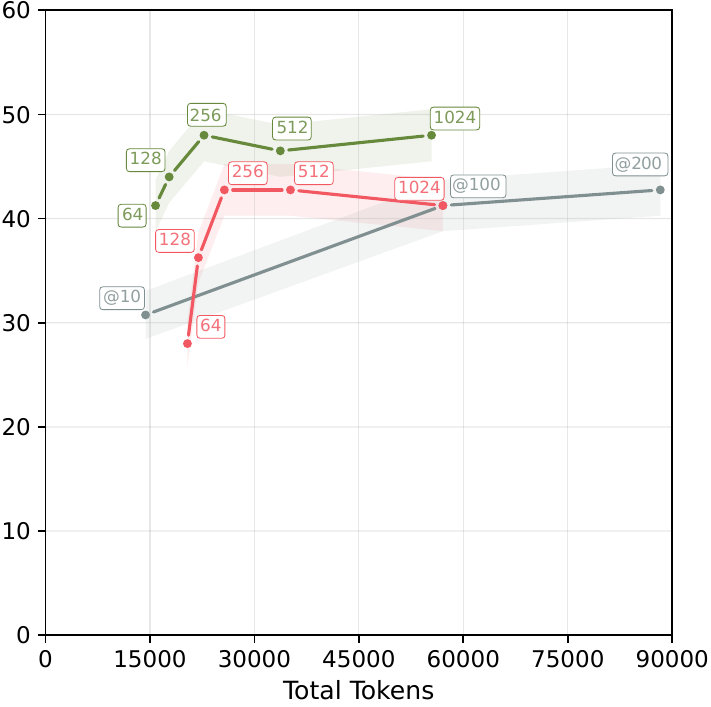}}
  \end{tabular}
  
  \vspace{1.5ex} 
  \textcolor[HTML]{F25862}{\Large $\bullet$} \small{Direct QA} \hspace{2em}
  \textcolor[HTML]{66893C}{\Large $\bullet$} \small{Two-Stage QA (ours)} \hspace{2em}
  \textcolor[HTML]{808F90}{\Large $\bullet$} \small{BGE-M3 @k}
  
  \caption{Accuracy against total tokens for direct QA and our two-stage method, on \mh{} (top) and \fldqa{} (bottom), for Gemma\,4~26B-A4B and Qwen 3.5-9B. The two left columns feed the full document; the two right columns feed a context retrieved by BGE-M3, with single-step QA over the retrieved units at increasing $k$ as reference. Markers along each line correspond to visual-token budgets, and bands show $\pm 1$~standard error of the mean across examples (Appendix~\ref{app:ci}). Results for all models are provided in Appendix~\ref{sec:app-all-results}.}
  \label{fig:method-results}
  \vspace{-10pt}
\end{figure*}

\section{Our Two-Stage Method: Identify \arr{First}, Reason Later}
\label{sec:method}


We test whether low-resolution table images retain enough information
for a model to identify the tables required to answer a question. Our
two-stage procedure is illustrated in Figure~\ref{fig:hero}. In the
\arr{first stage}, the model receives the document text together with all
tables compressed to a fixed visual-token budget. \arr{As in any
  pipeline that represents tables as images, we assume table
  boundaries are known (see Limitations).}  Each table is preceded by
a \texttt{[TABLE\_n]} marker \arr{(we ablate the effects of including
  this identifier in Appendix~\ref{sec:app-noid})}, and the model
identifies the tables it requires using
\texttt{\arr{\{}"request\_table": n\arr{\}}}. In the \arr{second stage}, the
requested tables are appended to the conversation at native
resolution, after which the model reasons and produces the answer.
The requested tables are returned at \arr{native resolution
  (Section~\ref{sec:budgets})}; we ablate this choice for Qwen models
in Appendix~\ref{sec:app-tool-budget}. \arr{The second stage keeps the
  full first-stage context and adds the requested tables}, while the
document text is provided as text. Our prompts are  in
Appendix~\ref{sec:app-prompts}.

The identification step is performed \emph{without} test-time reasoning, minimising 
its token overhead; in preliminary experiments, enabling reasoning produced
no significant improvement. The resulting method is training-free,
prompt-level, and model-agnostic. It builds on evidence that separating
identification from reasoning benefits long-context QA
\citep{guan2026evidence,zhao2026recontext}, while using pixel compression to
make identification over the full set of document tables inexpensive.

We include two controls to isolate the information supplied by the compressed
images. The \textsc{html} condition applies the same two-stage procedure with
tables represented as text. In \textsc{tableid}, each table is replaced by its
identifier, removing its contents while preserving its position and surrounding
context. Performance above \textsc{tableid} therefore indicates that compressed
table images provide useful evidence-identification signals.

\subsection{Stage 1: Relevant-Table Identification}
\label{sec:identification-results}

We evaluate evidence identification independently of downstream
reasoning, using F1 over the requested tables. Our main analysis uses
\mh{}, which provides direct table-level evidence annotations.
\fldqa{} instead annotates evidence pages, so treating every table on
an evidence page as relevant produces \arr{a superset of the true
  evidence set, and F1 on \fldqa{} is not directly
  comparable}. Figure~\ref{fig:qa1} reports results for Gemma\,4
26B-A4B and Qwen 3.5-9B; results for all models \arr{and for \fldqa{}}
are provided in Appendix~\ref{sec:app-identification-all}. A
random-selection baseline achieves 27.4\% F1.

\paragraph{Table identification is robust to compression.}
Identification performance varies little across visual-token budgets:
tables compressed to 64 tokens are identified nearly as accurately as
tables at higher resolutions or in \arr{\textsc{html}} form. This
contrasts with transcription and direct QA, both of which degrade at
the same budget (Figures~\ref{fig:ocr} and~\ref{fig:qa}). Aggressive
compression therefore removes the fine-grained information needed to
read cell values but preserves the broader cues that reveal what a
table is about and whether it is relevant to the question.
\arr{Gemma\,4 26B-A4B requests no table for 11--16\% of questions
  across budgets, and answers none of these correctly, whereas Qwen
  3.5-9B almost always requests one. When Gemma does request tables,
  its F1 exceeds Qwen's.}

\paragraph{Compressed images provide useful evidence.}
Adding \mbox{64-token} images improves F1 \arr{over the \textsc{tableid}
control} by~\arr{17 points} for Gemma\,4 26B-A4B and 10 points for Qwen 3.5-9B, confirming that the model uses information retained in
the compressed renderings rather than relying only on positional or
textual cues. The \textsc{tableid} condition nevertheless remains
above chance because surrounding passages often introduce or describe
the adjacent tables.

\subsection{Stage 2: Full-Resolution Reasoning}
\label{sec:perf-eff}

Figure~\ref{fig:method-results} (\arr{left two columns}) reports accuracy against
total decoder tokens, and Table~\ref{tab:perf-eff} summarises the token
savings of our method at budget~64 against single-step references:
the cheapest configuration it matches in accuracy (\emph{Matched}) and
single-step QA over native-resolution images (\emph{Image}). The
\emph{Retrieval} column is discussed in Section~\ref{sec:rag}.
All configurations include a fixed cost of $2{,}587{\pm}88$ tokens on
\mh{} and $60{,}842{\pm}202$ on \fldqa{} (median per example; mean
$\pm$ std.\ across models) for  
the encoding of the surrounding document text. We include this cost in all reported totals
but subtract it from the plot axes for readability.

\begin{table}[t]
  \centering
  \small
  \setlength{\tabcolsep}{4pt}
  \begin{tabular}{@{}l I@{~} r I@{~} r r I r r@{}}
    \toprule
    \textbf{Identify First (64)} & \textbf{Matched} & \multicolumn{2}{cI}{\textbf{Image}} & \multicolumn{2}{c@{}}{\textbf{Retrieval}} \\
     \textbf{Reason \arr{Later} vs.} & Tok. & Tok. & Acc. & Tok. & Acc. \\
    \midrule
    \multicolumn{6}{@{}l}{\textbf{\mh{}}} \\
    Qwen 3.5-9B & 20.9 & 20.9 & 0.0 & 25.2 & $-$0.7 \\
    Qwen 3 VL-8B & 13.1 & 14.6 & $-$4.4 & 14.8 & $-$5.0 \\
    Gemma\,4 26B-A4B & 27.2 & 43.2 & $-$6.9 & 38.7 & $-$7.6 \\
    Gemma\,4 E4B & 9.1 & 32.7 & $-$0.3 & 16.2 & $-$3.3 \\
    \arr{GLM-4.6V} & 13.9 & 18.8 & $-$1.5 & 14.4 & $-$1.1 \\
    \arr{Gemini 3.8 Flash} & 7.2 & 36.1 & +0.7 & 39.8 & $-$2.8 \\
    \cmidrule(lr){1-6}
    Average & 15.2 & 27.7 & $-$2.1 & 24.9 & $-$3.4 \\
    \midrule
    \multicolumn{6}{@{}l}{\textbf{\fldqa{} }} \\
    Qwen 3.5-9B & 32.5 & 27.7 & +11.2 & 51.0 & +2.2 \\
    Gemma\,4 26B-A4B & 28.2 & 53.5 & $-$1.4 & 72.3 & 0.0 \\
    \arr{Gemini 3.8 Flash} & 2.7 & 43.3 & +6.0 & 60.6 & $-$0.2 \\
    \cmidrule(lr){1-6}
    Average & 21.1 & 41.5 & +5.3 & 61.3 & +0.7 \\
    \bottomrule
  \end{tabular}
  \caption{
  Percentage of \textbf{tokens saved} (Tok.) and accuracy difference in points (Acc.; positive favours our method) of our two-stage method (at budget 64 and 280 for Gemini) against three single-step references. \emph{Matched} compares the cheapest configurations of each method at matched accuracy (Appendix~\ref{sec:app:matched}); \emph{Image} is single-step QA over table images at native resolution; \emph{Retrieval} \arr{runs both on} a context retrieved by BGE-M3 ($k{=}5$ on \mh{}, $k{=}100$ on \fldqa{}). 
  }
  \label{tab:perf-eff}
  \vspace{-10pt}
\end{table}

On \mh{}, the two-stage procedure avoids much of the reasoning-token
inflation observed in Section~\ref{sec:qa} at budgets of 512 tokens or
fewer. The \arr{best trade-off} is for Gemma\,4 26B-A4B: at 256 tokens, our
method uses \arr{35.1}\% fewer tokens than single-step QA over
native-resolution images, with a loss of \arr{4.3} accuracy points. It is
also \arr{more efficient than representing tables as \textsc{html} at
budgets of 512 or fewer}; for
Qwen 3.5-9B, it matches the accuracy of native-resolution images while
using 20.9\% fewer tokens. At budget~64, our method matches the
cheapest equivalent single-step configuration while saving \arr{15.2}\% of
tokens on average, and trails native-resolution images by only \arr{2.1}
accuracy points at \arr{27.7}\% lower cost (Table~\ref{tab:perf-eff}).

On \fldqa{}, the method improves \arr{over direct QA at the same budget
at every compressed budget}. 
At \arr{each model's best budget, 128} 
tokens \arr{for Gemma\,4 26B-A4B and 64 for Qwen 3.5-9B}, our method exceeds native-resolution single-step QA by \arr{3.8} accuracy 
points while using \arr{50}\% fewer tokens for Gemma\,4 26B-A4B,
and by 11.2 points while using 27.7\% fewer tokens for Qwen 3.5-9B. On
average across \arr{the three models}, it gains \arr{5.3} accuracy points over
native-resolution images while using \arr{41.5}\% fewer tokens
(Table~\ref{tab:perf-eff}).

\paragraph{Combining identification and compression.}
Our method combines two effects: identification, which improves
accuracy by restating the requested tables at native resolution next
to the question, and compression, which reduces cost by making
identification over all tables inexpensive.
Without compression, that is, at native resolution or with tables
serialised as \textsc{html}, any difference from direct QA stems from
the identification step alone. On the shorter \mh{} documents, this
step brings no consistent gain at native resolution (between $-$4.2
and $+$1.3 accuracy points across models;
Table~\ref{tab:app-full-mh}), whereas on the long \fldqa{} documents
it improves accuracy in five of the six uncompressed comparisons, by
up to 17.0 points; the only exception is Gemma\,4 26B-A4B at native
resolution ($-$1.0; Table~\ref{tab:app-full-fldqa}).
This is consistent with models
struggling to use evidence buried in long contexts
\citep{liu-etal-2024-lost,zhao2026recontext}: restating the relevant
tables next to the question matters when they sit among 80 others,
but not among three. Placing the relevant tables last is, however,
not available to single-step QA, as it requires knowing which
are relevant, \arr{the job of the first stage}.

\arr{Compression, in turn, reduces cost, and since the second stage always
reasons over native-resolution tables, accuracy is largely preserved
as the budget decreases: for Gemma\,4 26B-A4B on \fldqa{}, lowering
the budget from 1,024 to 64 tokens reduces total tokens from 158,204
to 73,124 at comparable accuracy (39.2 vs.\ 38.8), whereas direct QA
drops to 28.7 at the same budget.}
\arr{For Gemma\,4 26B-A4B, single-step \textsc{html} QA is more
accurate than ours at 64 tokens (44.5 vs.\ 38.8 on \fldqa{}),
but costs 50\% more.}

\subsection{Relationship to External Retrieval}
\label{sec:rag}
\arr{Our method and external retrieval are complementary. The two-stage procedure works on any context: when a document exceeds the context window, retrieval can first reduce it to its most relevant paragraphs and tables, and identification then runs on the result. We ask whether a dedicated retriever can replace compressed identification, and whether the procedure still helps after retrieval. All retrievers are used zero-shot, with the original question as the query and tables as atomic units.}

\paragraph{\arr{External retrieval identifies better but does not answer better.}}
We replace the identification stage with ColQwen2
\citep{faysse2024colpali}, a multi-vector retriever designed for
visually represented text and the stronger retriever in our
experiments (Appendix~\ref{sec:app-retrievers}).
Retrieved tables are supplied \arr{at native resolution} alongside the 
document text, and we evaluate \arr{$k \in \{1,2,3,5\}$ on \mh{} and
$k \in \{2,4,10,50,75,100\}$ on \fldqa{}}.

ColQwen2 identifies evidence tables more accurately than models reading a
compressed context (Figure~\ref{fig:qa1}), but this advantage does not carry through to downstream accuracy.
On \mh{}, retrieval plateaus at the accuracy our method already reaches: for
Qwen 3.5-9B, our 64-token configuration matches the accuracy of the \arr{most
accurate ColQwen2 configuration (at its lowest cost)}, using 19\% fewer
tokens. Only for Gemma\,4 26B-A4B does retrieval overtake our method, \arr{by up to
7.3 accuracy points, while our method uses 45.3\% of its tokens.} On \fldqa{} our method exceeds retrieval by 16 accuracy points at the same cost. External retrieval therefore identifies better without delivering a better end-to-end
trade-off. Full curves are in Appendix~\ref{sec:app-colqwen}.

\paragraph{Our two-stage method still helps after retrieval.}
We next apply our two-stage method to contexts retrieved by BGE-M3
\citep{chen-etal-2024-m3} from \arr{the current document alone, indexing
its paragraphs and tables together as candidate units.}
We use $k{=}5$ on \mh{} and $k{=}100$ on \fldqa{}, render the
retrieved tables at each visual-token budget, and compare against
single-step QA over the same retrieved context (see Figure~\ref{fig:method-results}, \arr{right two columns}).

On \mh{}, retrieval performance falls sharply below $k{=}5$,
limiting how far retrieval alone can reduce the context. 
\arr{Our method moves beyond this limit: Qwen 3.5-9B matches the accuracy
of single-step QA at $k{=}5$ using 25.2\% fewer tokens, while Gemma\,4
26B-A4B at a budget of 256 matches single-step QA at $k{=}3$ (39.4
vs.\ 39.8) using 27.2\% fewer tokens.}
The same general pattern holds for the remaining models except
Gemma\,4 E4B. As in Section~\ref{sec:perf-eff}, compression alone
does not achieve these gains; they arise from identifying and
restoring only the tables needed for reasoning.
On \fldqa{}, our method matches the accuracy of retrieval using \arr{61.3\% fewer tokens on average} (see Table~\ref{tab:perf-eff}, \emph{Retrieval} column).

\section{Conclusions}
\label{sec:conclusions}

We investigated whether pixel-level compression can reduce the cost of
question answering over multi-table documents. Moderate compression
allows table images to match their \textsc{html} serialisation \arr{in
accuracy on \mh{}} while
using fewer tokens. Under aggressive compression, however, models lose
the cell-level information needed for numerical reasoning and \arr{reason
longer}, eroding the efficiency gains.

Our central finding is that compression affects reading and
identification differently. Even when a table is too compressed to
support accurate reasoning, it retains enough information for identifying whether it is relevant. We exploit this
asymmetry with a training-free, two-stage method that first identifies
relevant tables in a compressed context and then restores only those
tables for \arr{native-resolution} reasoning. 
The method improves the \arr{accuracy--efficiency} trade-off in full-context QA, using 41\% fewer tokens while gaining \arr{5} accuracy points over single-step QA with native-resolution tables on \arr{\fldqa{}, and also saves tokens after retrieval}.

More broadly, our
results \arr{on financial documents} suggest that pixel compression is
\arr{best used not as a representation to reason over, but as a cheap
step for locating evidence.}

\section*{Limitations}

Our experiments are limited to English-language financial documents
and two datasets; moreover, \mh{} results are reported on the
development set for the reasons discussed in
Section~\ref{sec:datasets}.

We render tables from clean structured
representations using a uniform style in order to preserve information
parity between the textual and visual settings. Real-world table
images may contain varied fonts, complex layouts, scanning \arr{artefacts},
and recognition errors. Our experiments also assume that table
boundaries are known; applying the method to raw documents may require
an additional table-detection or extraction stage.

The effective compression level depends on table dimensions,
information density, rendering style, and the model's visual encoder.
The specific token budgets identified here may therefore not transfer
directly to other domains or document formats, although the evaluation
procedure can be used to determine suitable budgets in new settings.

\arr{Our method further relies on identifiers to refer to individual
tables; removing them leaves its advantage over single-step QA intact
but reduces accuracy under aggressive compression
(Appendix~\ref{sec:app-noid}).}

Finally, we use the number of decoder tokens as an
architecture-independent proxy for efficiency. This measure captures
both input and generated tokens, but not visual-encoder computation.
\arr{We complement it with wall-clock measurements on a single hardware
setup (Appendix~\ref{sec:app-runtime}), which include the overhead of
the additional inference pass; measuring memory use and FLOPs, and
timing on other hardware, would be necessary to establish the full
computational trade-off.}

\section*{Acknowledgments}
We thank the anonymous reviewers and area chair for their feedback.
This work was supported by the Edinburgh International Data Facility
(EIDF) and the Data-Driven Innovation Programme at the University of
Edinburgh. Access to EIDF was facilitated through the University of
Edinburgh's Generative AI Laboratory GAIL Fellow scheme. We gratefully acknowledge financial support from the Engineering and Physical Sciences Research Council (EPSRC; grant EP/W002876/1) and UK Research and Innovation (UKRI; grant UKRI/EP/D005028/1).


\bibliography{custom}

@article{liu-etal-2024-lost,
    title = "Lost in the Middle: How Language Models Use Long Contexts",
    author = "Liu, Nelson F.  and
      Lin, Kevin  and
      Hewitt, John  and
      Paranjape, Ashwin  and
      Bevilacqua, Michele  and
      Petroni, Fabio  and
      Liang, Percy",
    journal = "Transactions of the Association for Computational Linguistics",
    volume = "12",
    year = "2024",
    address = "Cambridge, MA",
    publisher = "MIT Press",
    url = "https://aclanthology.org/2024.tacl-1.9/",
    doi = "10.1162/tacl_a_00638",
    pages = "157--173"
}

@inproceedings{hsieh2024ruler,
  title={{RULER}: What's the Real Context Size of Your Long-Context Language Models?},
  author={Cheng-Ping Hsieh and Simeng Sun and Samuel Kriman and Shantanu Acharya and Dima Rekesh and Fei Jia and Yang Zhang and Boris Ginsburg},
  booktitle={First Conference on Language Modeling},
  year={2024},
}

@misc{liu2025comprehensivesurveylongcontext,
      title={A Comprehensive Survey on Long Context Language Modeling}, 
      author={Jiaheng Liu and Dawei Zhu and Zhiqi Bai and Yancheng He and Huanxuan Liao and Haoran Que and Zekun Wang and Chenchen Zhang and Ge Zhang and Jiebin Zhang and Yuanxing Zhang and Zhuo Chen and Hangyu Guo and Shilong Li and Ziqiang Liu and Yong Shan and Yifan Song and Jiayi Tian and Wenhao Wu and Zhejian Zhou and Ruijie Zhu and Junlan Feng and Yang Gao and Shizhu He and Zhoujun Li and Tianyu Liu and Fanyu Meng and Wenbo Su and Yingshui Tan and Zili Wang and Jian Yang and Wei Ye and Bo Zheng and Wangchunshu Zhou and Wenhao Huang and Sujian Li and Zhaoxiang Zhang},
      year={2025},
      eprint={2503.17407},
      archivePrefix={arXiv},
      primaryClass={cs.CL},
      url={https://arxiv.org/abs/2503.17407}, 
}

@inproceedings{yoon-etal-2024-compact,
    title = "{C}omp{A}ct: Compressing Retrieved Documents Actively for Question Answering",
    author = "Yoon, Chanwoong  and
      Lee, Taewhoo  and
      Hwang, Hyeon  and
      Jeong, Minbyul  and
      Kang, Jaewoo",
    editor = "Al-Onaizan, Yaser  and
      Bansal, Mohit  and
      Chen, Yun-Nung",
    booktitle = "Proceedings of the 2024 Conference on Empirical Methods in Natural Language Processing",
    month = nov,
    year = "2024",
    address = "Miami, Florida, USA",
    publisher = "Association for Computational Linguistics",
    url = "https://aclanthology.org/2024.emnlp-main.1194/",
    doi = "10.18653/v1/2024.emnlp-main.1194",
    pages = "21424--21439"
}

@misc{chen2024generativeadapter,
      title={Generative Adapter: Contextualizing Language Models in Parameters with A Single Forward Pass}, 
      author={Tong Chen and Hao Fang and Patrick Xia and Xiaodong Liu and Benjamin Van Durme and Luke Zettlemoyer and Jianfeng Gao and Hao Cheng},
      year={2024},
      eprint={2411.05877},
      archivePrefix={arXiv},
      primaryClass={cs.LG},
      url={https://arxiv.org/abs/2411.05877}, 
}

@misc{gunther2025latechunking,
      title={Late Chunking: Contextual Chunk Embeddings Using Long-Context Embedding Models}, 
      author={Michael Günther and Isabelle Mohr and Daniel James Williams and Bo Wang and Han Xiao},
      year={2025},
      eprint={2409.04701},
      archivePrefix={arXiv},
      primaryClass={cs.CL},
      url={https://arxiv.org/abs/2409.04701}, 
}

@misc{zhang2024chainagents,
      title={Chain of Agents: Large Language Models Collaborating on Long-Context Tasks}, 
      author={Yusen Zhang and Ruoxi Sun and Yanfei Chen and Tomas Pfister and Rui Zhang and Sercan Ö. Arik},
      year={2024},
      eprint={2406.02818},
      archivePrefix={arXiv},
      primaryClass={cs.CL},
      url={https://arxiv.org/abs/2406.02818}, 
}

@misc{zhao2026recontext,
      title={{ReContext}: Recursive Evidence Replay as {LLM} Harness for Long-Context Reasoning}, 
      author={Yanjun Zhao and Ruizhong Qiu and Tianxin Wei and Yuanchen Bei and Zhining Liu and Lingjie Chen and Ismini Lourentzou and Hanghang Tong and Jingrui He},
      year={2026},
      eprint={2607.02509},
      archivePrefix={arXiv},
      primaryClass={cs.AI},
      url={https://arxiv.org/abs/2607.02509}, 
}

@misc{rust2023pixels,
      title={Language Modelling with Pixels}, 
      author={Phillip Rust and Jonas F. Lotz and Emanuele Bugliarello and Elizabeth Salesky and Miryam de Lhoneux and Desmond Elliott},
      year={2023},
      eprint={2207.06991},
      archivePrefix={arXiv},
      primaryClass={cs.CL},
      url={https://arxiv.org/abs/2207.06991}, 
}

@inproceedings{lee2023pix2struct,
    author = {Lee, Kenton and Joshi, Mandar and Turc, Iulia and Hu, Hexiang and Liu, Fangyu and Eisenschlos, Julian and Khandelwal, Urvashi and Shaw, Peter and Chang, Ming-Wei and Toutanova, Kristina},
    title = {{Pix2Struct}: screenshot parsing as pretraining for visual language understanding},
    year = {2023},
    publisher = {JMLR.org},
    booktitle = {Proceedings of the 40th International Conference on Machine Learning},
    articleno = {780},
    numpages = {20},
    location = {Honolulu, Hawaii, USA},
    series = {ICML'23}
}

@inproceedings{lotz-etal-2023-text,
    title = "Text Rendering Strategies for Pixel Language Models",
    author = "Lotz, Jonas  and
      Salesky, Elizabeth  and
      Rust, Phillip  and
      Elliott, Desmond",
    editor = "Bouamor, Houda  and
      Pino, Juan  and
      Bali, Kalika",
    booktitle = "Proceedings of the 2023 Conference on Empirical Methods in Natural Language Processing",
    month = dec,
    year = "2023",
    address = "Singapore",
    publisher = "Association for Computational Linguistics",
    url = "https://aclanthology.org/2023.emnlp-main.628/",
    doi = "10.18653/v1/2023.emnlp-main.628",
    pages = "10155--10172"
}

@article{gao2024improving,
  title={Improving Language Understanding from Screenshots},
  author={Gao, Tianyu and Wang, Zirui and Bhaskar, Adithya and Chen, Danqi},
  journal={arXiv preprint arXiv:2402.14073},
  year={2024}
}

@article{wei2025deepseek,
  title={{DeepSeek-OCR}: Contexts Optical Compression},
  author={Wei, Haoran and Sun, Yaofeng and Li, Yukun},
  journal={arXiv preprint arXiv:2510.18234},
  year={2025}
}

@article{xing2025vision,
  title={Vision-centric Token Compression in Large Language Model},
  author={Xing, Ling and Wang, Alex Jinpeng and Yan, Rui and Shu, Xiangbo and Tang, Jinhui},
  journal={arXiv preprint arXiv:2502.00791},
  year={2025}
}

@inproceedings{cheng-etal-2026-glyph,
    title = "Glyph: Scaling Context Windows via Visual-Text Compression",
    author = "Cheng, Jiale  and
      Liu, Yusen  and
      Zhang, Xinyu  and
      Fei, Yulin  and
      Hong, Wenyi  and
      Lyu, Ruiliang  and
      Wang, Weihan  and
      Su, Zhe  and
      Gu, Xiaotao  and
      Liu, Xiao  and
      Bai, Yushi  and
      Tang, Jie  and
      Wang, Hongning  and
      Huang, Minlie",
    editor = "Liakata, Maria  and
      Moreira, Viviane P.  and
      Zhang, Jiajun  and
      Jurgens, David",
    booktitle = "Proceedings of the 64th Annual Meeting of the {A}ssociation for {C}omputational {L}inguistics (Volume 1: Long Papers)",
    month = jul,
    year = "2026",
    address = "San Diego, California, United States",
    publisher = "Association for Computational Linguistics",
    url = "https://aclanthology.org/2026.acl-long.1722/",
    doi = "10.18653/v1/2026.acl-long.1722",
    pages = "37145--37158",
    ISBN = "979-8-89176-390-6"
}

@inproceedings{li-etal-2025-text,
    title = "Text or Pixels? Evaluating Efficiency and Understanding of {LLM}s with Visual Text Inputs",
    author = "Li, Yanhong  and
      Lan, Zixuan  and
      Zhou, Jiawei",
    editor = "Christodoulopoulos, Christos  and
      Chakraborty, Tanmoy  and
      Rose, Carolyn  and
      Peng, Violet",
    booktitle = "Findings of the Association for Computational Linguistics: EMNLP 2025",
    month = nov,
    year = "2025",
    address = "Suzhou, China",
    publisher = "Association for Computational Linguistics",
    url = "https://aclanthology.org/2025.findings-emnlp.558/",
    doi = "10.18653/v1/2025.findings-emnlp.558",
    pages = "10564--10578",
    ISBN = "979-8-89176-335-7"
}

@misc{wang2024leveraging,
      title={Leveraging Visual Tokens for Extended Text Contexts in Multi-Modal Learning}, 
      author={Alex Jinpeng Wang and Linjie Li and Yiqi Lin and Min Li and Lijuan Wang and Mike Zheng Shou},
      year={2024},
      eprint={2406.02547},
      archivePrefix={arXiv},
      primaryClass={cs.CV},
      url={https://arxiv.org/abs/2406.02547}, 
}

@inproceedings{deng-etal-2024-tables,
    title = "Tables as Texts or Images: Evaluating the Table Reasoning Ability of {LLM}s and {MLLM}s",
    author = "Deng, Naihao  and
      Sun, Zhenjie  and
      He, Ruiqi  and
      Sikka, Aman  and
      Chen, Yulong  and
      Ma, Lin  and
      Zhang, Yue  and
      Mihalcea, Rada",
    editor = "Ku, Lun-Wei  and
      Martins, Andre  and
      Srikumar, Vivek",
    booktitle = "Findings of the Association for Computational Linguistics: ACL 2024",
    month = aug,
    year = "2024",
    address = "Bangkok, Thailand",
    publisher = "Association for Computational Linguistics",
    url = "https://aclanthology.org/2024.findings-acl.23/",
    doi = "10.18653/v1/2024.findings-acl.23",
    pages = "407--426"
}

@misc{xing2025tabledart,
    title={{TableDART}: Dynamic Adaptive Multi-Modal Routing for Table Understanding}, 
    author={Xiaobo Xing and Wei Yuan and Tong Chen and Quoc Viet Hung Nguyen and Xiangliang Zhang and Hongzhi Yin},
    year={2025},
    eprint={2509.14671},
    archivePrefix={arXiv},
    primaryClass={cs.CL},
    url={https://arxiv.org/abs/2509.14671}
}

@inproceedings{zheng-etal-2024-multimodal,
    title = "Multimodal Table Understanding",
    author = "Zheng, Mingyu  and
      Feng, Xinwei  and
      Si, Qingyi  and
      She, Qiaoqiao  and
      Lin, Zheng  and
      Jiang, Wenbin  and
      Wang, Weiping",
    editor = "Ku, Lun-Wei  and
      Martins, Andre  and
      Srikumar, Vivek",
    booktitle = "Proceedings of the 62nd Annual Meeting of the Association for Computational Linguistics (Volume 1: Long Papers)",
    month = aug,
    year = "2024",
    address = "Bangkok, Thailand",
    publisher = "Association for Computational Linguistics",
    url = "https://aclanthology.org/2024.acl-long.493/",
    doi = "10.18653/v1/2024.acl-long.493",
    pages = "9102--9124"
}

@inproceedings{sui2023table,
  title     = {Table Meets {LLM}: Can Large Language Models Understand Structured Table Data? A Benchmark and Empirical Study},
  author    = {Yuan Sui and Mengyu Zhou and Mingjie Zhou and Shi Han and Dongmei Zhang},
  booktitle = {Proceedings of the 17th {ACM} International Conference on Web Search and Data Mining},
  pages     = {645--654},
  publisher = {ACM},
  year      = {2024},
  doi       = {10.1145/3616855.3635752},
  bibSource = {Semantic Scholar https://www.semanticscholar.org/paper/f534f566535f4e0fd2b72b1db3b18c47479e5092}
}

@misc{wu2025vtoolr1vlmslearnthink,
      title={VTool-R1: VLMs Learn to Think with Images via Reinforcement Learning on Multimodal Tool Use}, 
      author={Mingyuan Wu and Jingcheng Yang and Jize Jiang and Meitang Li and Kaizhuo Yan and Hanchao Yu and Minjia Zhang and Chengxiang Zhai and Klara Nahrstedt},
      year={2025},
      eprint={2505.19255},
      archivePrefix={arXiv},
      primaryClass={cs.LG},
      url={https://arxiv.org/abs/2505.19255}, 
}

@misc{alonso2026tablet,
      title={TABLET: A Large-Scale Dataset for Robust Visual Table Understanding}, 
      author={Iñigo Alonso and Imanol Miranda and Eneko Agirre and Mirella Lapata},
      year={2026},
      eprint={2509.21205},
      archivePrefix={arXiv},
      primaryClass={cs.CV},
      url={https://arxiv.org/abs/2509.21205}, 
}

@misc{kim2024tablevqabench,
      title={TableVQA-Bench: A Visual Question Answering Benchmark on Multiple Table Domains}, 
      author={Yoonsik Kim and Moonbin Yim and Ka Yeon Song},
      year={2024},
      eprint={2404.19205},
      archivePrefix={arXiv},
      primaryClass={cs.CV},
      url={https://arxiv.org/abs/2404.19205}, 
}

@inproceedings{singh-etal-2025-mtabvqa,
    title = "{MT}ab{VQA}: Evaluating Multi-Tabular Reasoning of Language Models in Visual Space",
    author = "Singh, Anshul  and
      Biemann, Chris  and
      Strich, Jan",
    editor = "Christodoulopoulos, Christos  and
      Chakraborty, Tanmoy  and
      Rose, Carolyn  and
      Peng, Violet",
    booktitle = "Findings of the Association for Computational Linguistics: EMNLP 2025",
    month = nov,
    year = "2025",
    address = "Suzhou, China",
    publisher = "Association for Computational Linguistics",
    url = "https://aclanthology.org/2025.findings-emnlp.1083/",
    doi = "10.18653/v1/2025.findings-emnlp.1083",
    pages = "19866--19891",
    ISBN = "979-8-89176-335-7"
}

@misc{wang2025needleinatable,
      title={NeedleInATable: Exploring Long-Context Capability of Large Language Models towards Long-Structured Tables}, 
      author={Lanrui Wang and Mingyu Zheng and Hongyin Tang and Zheng Lin and Yanan Cao and Jingang Wang and Xunliang Cai and Weiping Wang},
      year={2025},
      eprint={2504.06560},
      archivePrefix={arXiv},
      primaryClass={cs.CL},
      url={https://arxiv.org/abs/2504.06560}, 
}

@misc{zou2025tattootoo,
      title={TaTToo: Tool-Grounded Thinking PRM for Test-Time Scaling in Tabular Reasoning}, 
      author={Jiaru Zou and Soumya Roy and Vinay Kumar Verma and Ziyi Wang and David Wipf and Pan Lu and Sumit Negi and James Zou and Jingrui He},
      year={2025},
      eprint={2510.06217},
      archivePrefix={arXiv},
      primaryClass={cs.AI},
      url={https://arxiv.org/abs/2510.06217}, 
}

@inproceedings{herzig-etal-2020-tapas,
    title = "{T}a{P}as: Weakly Supervised Table Parsing via Pre-training",
    author = {Herzig, Jonathan  and
      Nowak, Pawel Krzysztof  and
      M{\"u}ller, Thomas  and
      Piccinno, Francesco  and
      Eisenschlos, Julian},
    editor = "Jurafsky, Dan  and
      Chai, Joyce  and
      Schluter, Natalie  and
      Tetreault, Joel",
    booktitle = "Proceedings of the 58th Annual Meeting of the Association for Computational Linguistics",
    month = jul,
    year = "2020",
    address = "Online",
    publisher = "Association for Computational Linguistics",
    url = "https://aclanthology.org/2020.acl-main.398/",
    doi = "10.18653/v1/2020.acl-main.398",
    pages = "4320--4333"
}

@inproceedings{wang2024chain,
  title={{Chain-of-Table}: Evolving Tables in the Reasoning Chain for Table Understanding},
  author={Wang, Zilong and Zhang, Hao and Li, Chun-Liang and Eisenschlos, Julian Martin and Perot, Vincent and Wang, Zifeng and Miculicich, Lesly and Fujii, Yasuhisa and Shang, Jingbo and Lee, Chen-Yu and Pfister, Tomas},
  booktitle={The Twelfth International Conference on Learning Representations},
  year={2024}
}

@misc{choi2026docprune,
      title={DocPrune:Efficient Document Question Answering via Background, Question, and Comprehension-aware Token Pruning}, 
      author={Joonmyung Choi and Sanghyeok Lee and Jongha Kim and Sehyung Kim and Dohwan Ko and Jihyung Kil and Hyunwoo J. Kim},
      year={2026},
      eprint={2604.22281},
      archivePrefix={arXiv},
      primaryClass={cs.CV},
      url={https://arxiv.org/abs/2604.22281}, 
}

@misc{chen2024fastv,
      title={An Image is Worth 1/2 Tokens After Layer 2: Plug-and-Play Inference Acceleration for Large Vision-Language Models}, 
      author={Liang Chen and Haozhe Zhao and Tianyu Liu and Shuai Bai and Junyang Lin and Chang Zhou and Baobao Chang},
      year={2024},
      eprint={2403.06764},
      archivePrefix={arXiv},
      primaryClass={cs.CV},
      url={https://arxiv.org/abs/2403.06764}, 
}

@inproceedings{shang2024PruMerge,
  title={{LLaVA-PruMerge}: Adaptive Token Reduction for Efficient Large Multimodal Models},
  author={Shang, Yuzhang and Cai, Mu and Xu, Bingxin and Lee, Yong Jae and Yan, Yan},
  booktitle={Proceedings of the IEEE/CVF International Conference on Computer Vision (ICCV)},
  year={2025}
}

@inproceedings{alonso-etal-2024-pixt3,
    title = "{P}ix{T}3: Pixel-based Table-To-Text Generation",
    author = "Alonso, I{\~n}igo  and
      Agirre, Eneko  and
      Lapata, Mirella",
    editor = "Ku, Lun-Wei  and
      Martins, Andre  and
      Srikumar, Vivek",
    booktitle = "Proceedings of the 62nd Annual Meeting of the Association for Computational Linguistics (Volume 1: Long Papers)",
    month = aug,
    year = "2024",
    address = "Bangkok, Thailand",
    publisher = "Association for Computational Linguistics",
    url = "https://aclanthology.org/2024.acl-long.364",
    pages = "6721--6736",
}

@inproceedings{parikh-etal-2020-totto,
    title = "{ToTTo}: A Controlled Table-To-Text Generation Dataset",
    author = "Parikh, Ankur  and
      Wang, Xuezhi  and
      Gehrmann, Sebastian  and
      Faruqui, Manaal  and
      Dhingra, Bhuwan  and
      Yang, Diyi  and
      Das, Dipanjan",
    booktitle = "Proceedings of the 2020 Conference on Empirical Methods in Natural Language Processing (EMNLP)",
    month = nov,
    year = "2020",
    address = "Online",
    publisher = "Association for Computational Linguistics",
    url = "https://aclanthology.org/2020.emnlp-main.89",
    doi = "10.18653/v1/2020.emnlp-main.89",
    pages = "1173--1186",
}

@inproceedings{2019TabFactA,
  title     = {{TabFact}: A Large-scale Dataset for Table-based Fact Verification},
  author    = {Chen, Wenhu and Wang, Hongmin and Chen, Jianshu and Zhang, Yunkai and Wang, Hong and Li, Shiyang and Zhou, Xiyou and Wang, William Yang},
  booktitle = {International Conference on Learning Representations (ICLR)},
  address   = {Addis Ababa, Ethiopia},
  month     = apr,
  year      = {2020}
}

@inproceedings{zhao-etal-2022-multihiertt,
    title = "{M}ulti{H}iertt: Numerical Reasoning over Multi Hierarchical Tabular and Textual Data",
    author = "Zhao, Yilun  and
      Li, Yunxiang  and
      Li, Chenying  and
      Zhang, Rui",
    editor = "Muresan, Smaranda  and
      Nakov, Preslav  and
      Villavicencio, Aline",
    booktitle = "Proceedings of the 60th Annual Meeting of the Association for Computational Linguistics (Volume 1: Long Papers)",
    month = may,
    year = "2022",
    address = "Dublin, Ireland",
    publisher = "Association for Computational Linguistics",
    url = "https://aclanthology.org/2022.acl-long.454/",
    doi = "10.18653/v1/2022.acl-long.454",
    pages = "6588--6600"
}

@article{wang2026finlongdocqa,
  title={Document-Level Numerical Reasoning across Single and Multiple Tables in Financial Reports},
  author={Wang, Yi-Cheng and Wang, Wei-An and Chen, Chu-Song},
  journal={arXiv preprint arXiv:2604.03664},
  year={2026}
}

@misc{zhong2020image,
      title={Image-based table recognition: data, model, and evaluation}, 
      author={Xu Zhong and Elaheh ShafieiBavani and Antonio Jimeno Yepes},
      year={2020},
      eprint={1911.10683},
      archivePrefix={arXiv},
      primaryClass={cs.CV},
      url={https://arxiv.org/abs/1911.10683}, 
}

@misc{kwok2026tabqaworld,
      title={{TABQAWORLD}: Optimizing Multimodal Reasoning for Multi-Turn Table Question Answering}, 
      author={Tung Sum Thomas Kwok and Xinyu Wang and Xiaofeng Lin and Peng Lu and Chunhe Wang and Changlun Li and Hanwei Wu and Nan Tang and Elisa Kreiss and Guang Cheng},
      year={2026},
      eprint={2604.03393},
      archivePrefix={arXiv},
      primaryClass={cs.AI},
      url={https://arxiv.org/abs/2604.03393}, 
}

@misc{gemmateam2026gemma4,
      title={Gemma 4 Technical Report}, 
      author={{Gemma Team}},
      year={2026},
      eprint={2607.02770},
      archivePrefix={arXiv},
      primaryClass={cs.CL},
      url={https://arxiv.org/abs/2607.02770}, 
}

@misc{bai2025qwen3vltechnicalreport,
      title={Qwen3-VL Technical Report}, 
      author={Shuai Bai and Yuxuan Cai and Ruizhe Chen and Keqin Chen and Xionghui Chen and Zesen Cheng and Lianghao Deng and Wei Ding and Chang Gao and Chunjiang Ge and Wenbin Ge and Zhifang Guo and Qidong Huang and Jie Huang and Fei Huang and Binyuan Hui and Shutong Jiang and Zhaohai Li and Mingsheng Li and Mei Li and Kaixin Li and Zicheng Lin and Junyang Lin and Xuejing Liu and Jiawei Liu and Chenglong Liu and Yang Liu and Dayiheng Liu and Shixuan Liu and Dunjie Lu and Ruilin Luo and Chenxu Lv and Rui Men and Lingchen Meng and Xuancheng Ren and Xingzhang Ren and Sibo Song and Yuchong Sun and Jun Tang and Jianhong Tu and Jianqiang Wan and Peng Wang and Pengfei Wang and Qiuyue Wang and Yuxuan Wang and Tianbao Xie and Yiheng Xu and Haiyang Xu and Jin Xu and Zhibo Yang and Mingkun Yang and Jianxin Yang and An Yang and Bowen Yu and Fei Zhang and Hang Zhang and Xi Zhang and Bo Zheng and Humen Zhong and Jingren Zhou and Fan Zhou and Jing Zhou and Yuanzhi Zhu and Ke Zhu},
      year={2025},
      eprint={2511.21631},
      archivePrefix={arXiv},
      primaryClass={cs.CV},
      url={https://arxiv.org/abs/2511.21631}, 
}

@misc{qwen35blog,
    title = {Qwen3.5: Towards Native Multimodal Agents},
    url = {https://qwen.ai/blog?id=qwen3.5},
    author = {Qwen Team},
    month = {February},
    year = {2026}
}

@misc{vteam2025glm45vglm41vthinkingversatilemultimodal,
      title={{GLM-4.5V} and {GLM-4.1V-Thinking}: Towards Versatile Multimodal Reasoning with Scalable Reinforcement Learning}, 
      author={V Team and Wenyi Hong and Wenmeng Yu and Xiaotao Gu and Guo Wang and Guobing Gan and Haomiao Tang and Jiale Cheng and Ji Qi and Junhui Ji and Lihang Pan and Shuaiqi Duan and Weihan Wang and Yan Wang and Yean Cheng and Zehai He and Zhe Su and Zhen Yang and Ziyang Pan and Aohan Zeng and Baoxu Wang and Bin Chen and Boyan Shi and Changyu Pang and Chenhui Zhang and Da Yin and Fan Yang and Guoqing Chen and Jiazheng Xu and Jiale Zhu and Jiali Chen and Jing Chen and Jinhao Chen and Jinghao Lin and Jinjiang Wang and Junjie Chen and Leqi Lei and Letian Gong and Leyi Pan and Mingdao Liu and Mingde Xu and Mingzhi Zhang and Qinkai Zheng and Sheng Yang and Shi Zhong and Shiyu Huang and Shuyuan Zhao and Siyan Xue and Shangqin Tu and Shengbiao Meng and Tianshu Zhang and Tianwei Luo and Tianxiang Hao and Tianyu Tong and Wenkai Li and Wei Jia and Xiao Liu and Xiaohan Zhang and Xin Lyu and Xinyue Fan and Xuancheng Huang and Yanling Wang and Yadong Xue and Yanfeng Wang and Yanzi Wang and Yifan An and Yifan Du and Yiming Shi and Yiheng Huang and Yilin Niu and Yuan Wang and Yuanchang Yue and Yuchen Li and Yutao Zhang and Yuting Wang and Yu Wang and Yuxuan Zhang and Zhao Xue and Zhenyu Hou and Zhengxiao Du and Zihan Wang and Peng Zhang and Debing Liu and Bin Xu and Juanzi Li and Minlie Huang and Yuxiao Dong and Jie Tang},
      year={2025},
      eprint={2507.01006},
      archivePrefix={arXiv},
      primaryClass={cs.CV},
      url={https://arxiv.org/abs/2507.01006}, 
}

@misc{gemini38flash2026,
  author       = {{Google DeepMind}},
  title        = {Gemini 3.8 Flash Model Card},
  year         = {2026},
  month        = sep,
  howpublished = {\url{https://deepmind.google/models/model-cards/gemini-3-8-flash/}},
  note         = {Model ID: \texttt{gemini-3.8-flash}. Accessed: 2026-11-23}
}

@misc{tschannen2025siglip2,
      title={SigLIP 2: Multilingual Vision-Language Encoders with Improved Semantic Understanding, Localization, and Dense Features}, 
      author={Michael Tschannen and Alexey Gritsenko and Xiao Wang and Muhammad Ferjad Naeem and Ibrahim Alabdulmohsin and Nikhil Parthasarathy and Talfan Evans and Lucas Beyer and Ye Xia and Basil Mustafa and Olivier Hénaff and Jeremiah Harmsen and Andreas Steiner and Xiaohua Zhai},
      year={2025},
      eprint={2502.14786},
      archivePrefix={arXiv},
      primaryClass={cs.CV},
      url={https://arxiv.org/abs/2502.14786}, 
}

@inproceedings{chen-etal-2024-m3,
    title = "{M}3-Embedding: Multi-Linguality, Multi-Functionality, Multi-Granularity Text Embeddings Through Self-Knowledge Distillation",
    author = "Chen, Jianlyu  and
      Xiao, Shitao  and
      Zhang, Peitian  and
      Luo, Kun  and
      Lian, Defu  and
      Liu, Zheng",
    editor = "Ku, Lun-Wei  and
      Martins, Andre  and
      Srikumar, Vivek",
    booktitle = "Findings of the Association for Computational Linguistics: ACL 2024",
    month = aug,
    year = "2024",
    address = "Bangkok, Thailand",
    publisher = "Association for Computational Linguistics",
    url = "https://aclanthology.org/2024.findings-acl.137/",
    doi = "10.18653/v1/2024.findings-acl.137",
    pages = "2318--2335"
}

@inproceedings{faysse2024colpali,
      title={{ColPali}: Efficient Document Retrieval with Vision Language Models},
      author={Manuel Faysse and Hugues Sibille and Tony Wu and Bilel Omrani and Gautier Viaud and C{\'e}line Hudelot and Pierre Colombo},
      booktitle={The Thirteenth International Conference on Learning Representations},
      year={2025},
      url={https://arxiv.org/abs/2407.01449},
}

@misc{guan2026evidence,
      title={Evidence-Augmented Policy Optimization with Reward Co-Evolution for Long-Context Reasoning}, 
      author={Xin Guan and Zijian Li and Shen Huang and Pengjun Xie and Jingren Zhou and Jiuxin Cao},
      year={2026},
      eprint={2601.10306},
      archivePrefix={arXiv},
      primaryClass={cs.AI},
      url={https://arxiv.org/abs/2601.10306}, 
}

@inproceedings{mathew2021docvqa,
    title = "{DocVQA}: A Dataset for {VQA} on Document Images",
    author = "Mathew, Minesh and Karatzas, Dimosthenis and Jawahar, C. V.",
    booktitle = "Proceedings of the IEEE/CVF Winter Conference on Applications of Computer Vision (WACV)",
    pages = "2200--2209",
    year = "2021"
}

\newpage

\appendix
\setcounter{topnumber}{2}
\setcounter{dbltopnumber}{2}
\renewcommand{\topfraction}{0.9}
\renewcommand{\dbltopfraction}{0.9}
\renewcommand{\floatpagefraction}{0.7}
\renewcommand{\dblfloatpagefraction}{0.7}

\section{Prompts}
\label{sec:app-prompts}

\begin{figure*}[p]
\centering
\includegraphics[scale=.80]{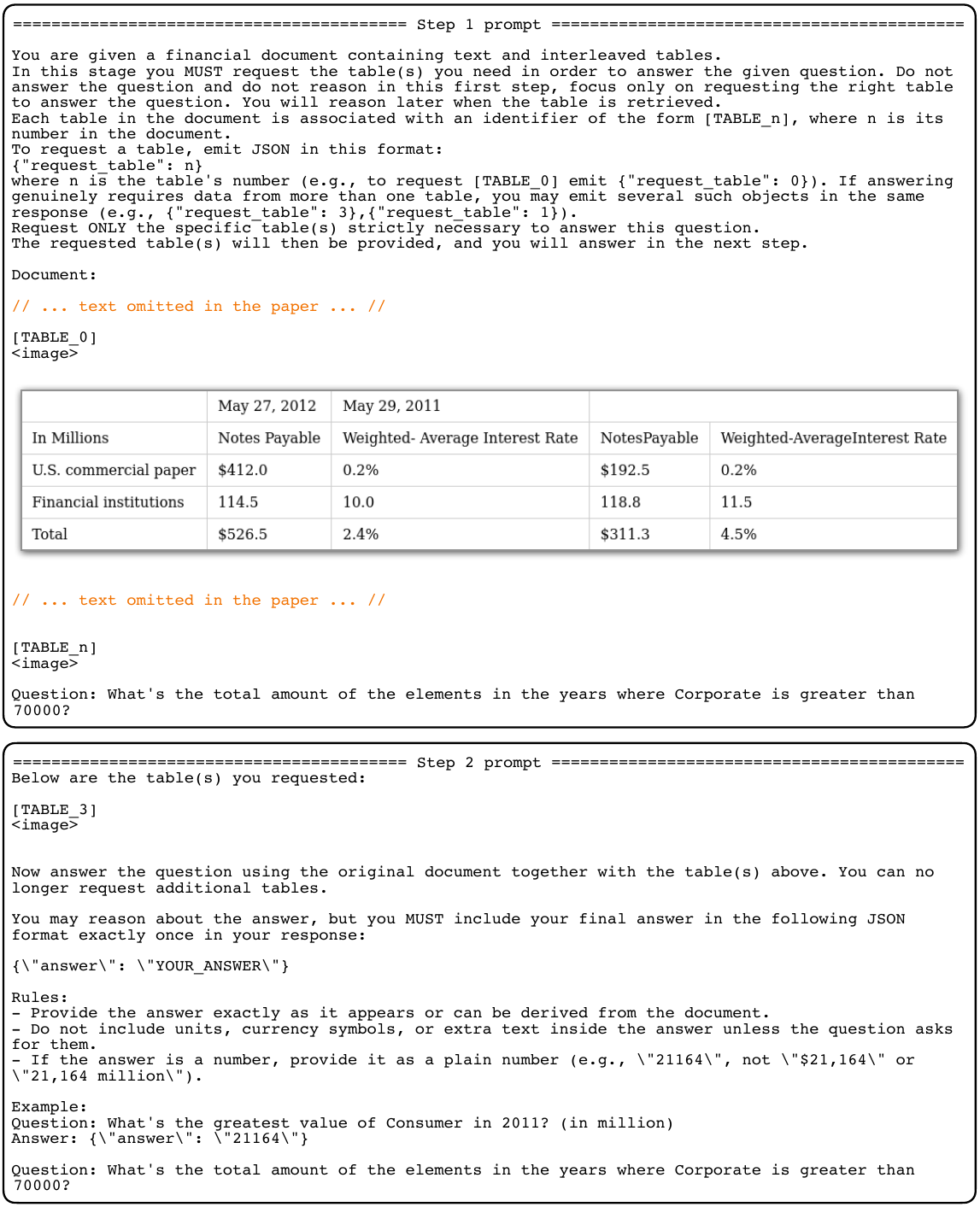}
\caption{Prompt used for our two-stage method.}
\label{fig:prompt_qactm}
\end{figure*}

This appendix shows the prompts used in our experiments \arr{(}Figures~\ref{fig:prompt_ocr}--\ref{fig:prompt_qactm}).
For readability, each prompt shows a single table image, but every \texttt{<image>} placeholder in the actual prompts is accompanied by its corresponding table image.
The tables shown here are included for reference only and are not rendered at the pixel dimensions of any particular visual-token budget.

\subsection{Transcription (OCR)}
\label{sec:ocr-example}

\begin{figure*}[p]
\centering
\includegraphics[scale=.80]{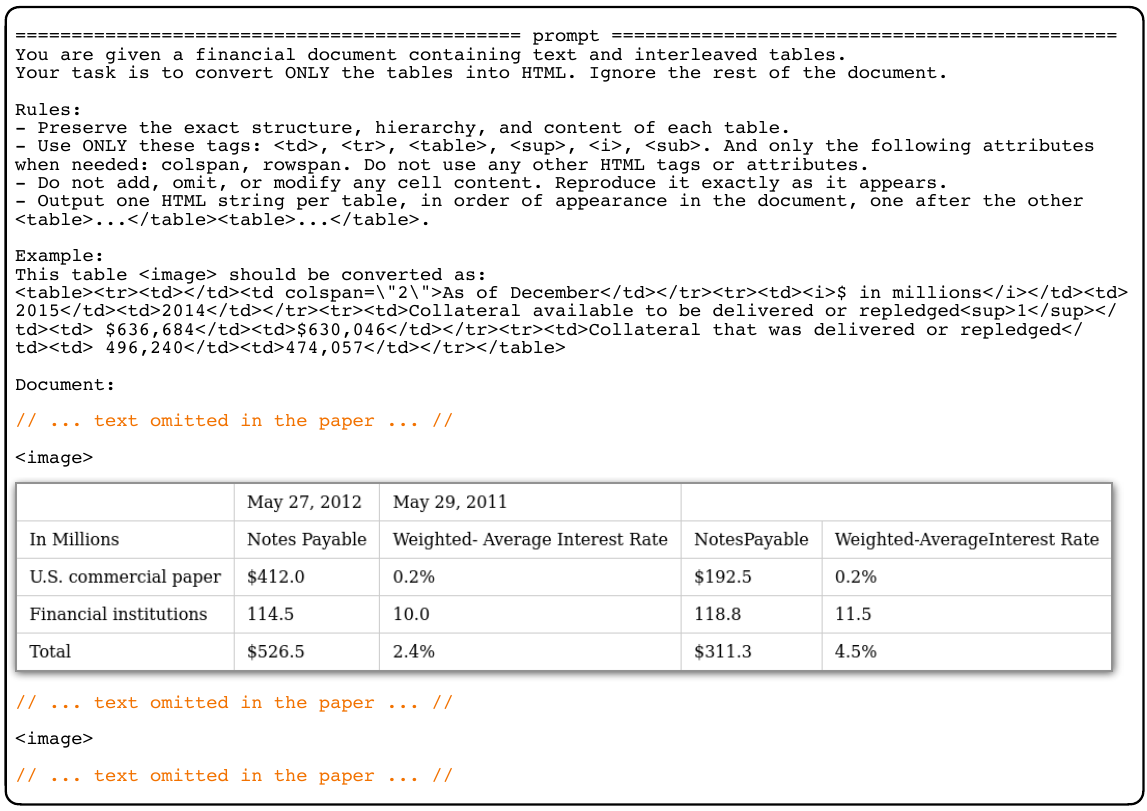}
\caption{Prompt used for table transcription.}
\label{fig:prompt_ocr}

\vspace{1em}
\includegraphics[scale=.80]{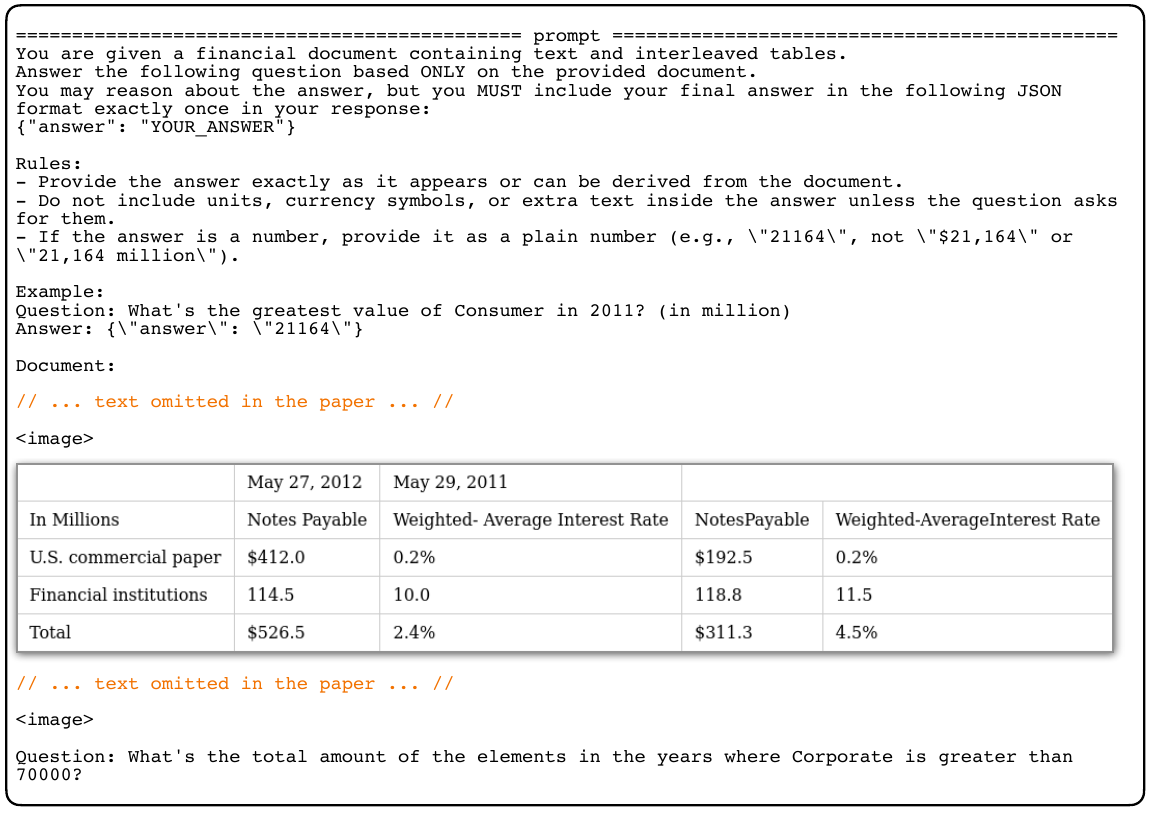}
\caption{Prompt used for direct QA.}
\label{fig:prompt_qa}
\end{figure*}

Figure~\ref{fig:prompt_ocr} shows the prompt used for table transcription, in the \textsc{img}$\rightarrow$\textsc{html} setting.
In the \textsc{html}$\rightarrow$\textsc{html} and \textsc{latex}$\rightarrow$\textsc{html} settings, each \texttt{<image>} placeholder is replaced by the \arr{\textsc{html} or \LaTeX{}} serialisation of the same table\arr{,} respectively, and the rest of the prompt remains unchanged.

\subsection{Question Answering (QA)}
\label{sec:app-prompts-qa}

Figures~\ref{fig:prompt_qa} and~\ref{fig:prompt_qactm} show the prompts used for direct QA and for our two-stage method, both  in the hybrid setting.
The \textsc{html} setting replaces each \texttt{<image>} placeholder with the \arr{\textsc{html}} serialisation of the same table, and the rest of the prompt remains unchanged.

\section{\arr{Confidence Intervals}}
\label{app:ci}

Every score we report is a mean over a fixed set of $N$ test examples.
We take the test example as the unit of analysis and report the standard error of the mean (s.e.m.).
Let $s_i\in\{0,1\}$ be a model's exact-match accuracy on example $i$.
We report the mean $\hat{\mu}=\frac{1}{N}\sum_i s_i$ (shown as a percentage) with a band of $\pm 1$~s.e.m., a $68\%$ interval under a normal approximation, which captures the sampling uncertainty of each plotted mean:
\begin{equation}
\mathrm{s.e.m.}=\frac{\hat{\sigma}}{\sqrt{N}},
\hspace{-15pt}\qquad
\hat{\sigma}^{2}=\frac{1}{N-1}\sum_{i=1}^{N}\bigl(s_i-\hat{\mu}\bigr)^{2}.
\end{equation}

\section{Quantifying Token Savings}
\label{sec:app:matched}

Each configuration in this work gives us two numbers, accuracy and token cost.
We know that the Pareto frontier lies in the upper-left corner of our accuracy per token graphs, but defining its precise gradient requires choosing which of the two to prioritise.
When selecting a point of reference to compare our method against, we prioritise accuracy.
That is, we compare the cheapest and most accurate configuration of each method, first selecting the most accurate and then, among the results at equivalent performance, the cheapest.
\arr{We define equivalence with a paired bootstrap over examples: since all
configurations of a model answer the same questions, we resample the
examples with replacement (10{,}000 resamples) and compute the
difference in accuracy between two configurations on each resample. A
configuration is considered equivalent to a target when it is not
significantly worse, that is, when the 95\% interval of this difference
does not lie entirely below zero.}

We also assume that the baseline can represent tables as images and that pixel-level compression is available to it as well.
Note that this strips from our results the improvements that come from those two techniques alone, so what remains comes solely from the two-stage procedure itself, making this a conservative estimate and, in our view, a fairer and more realistic comparison.
Algorithm~\ref{alg:token-savings} states the procedure.

\begin{algorithm}[t]
\caption{Token savings at matched accuracy}
\label{alg:token-savings}
\begin{algorithmic}[1]
\State Order the baseline results by accuracy, highest first, and take each in turn as the target level.
\State Collect every configuration, from either method, whose accuracy is \arr{not significantly below the target under the paired bootstrap}.
\State If none of our method's configurations qualifies, the target is out of reach; skip to the next.
\State Otherwise take the cheapest qualifying configuration of our method and the cheapest qualifying baseline.
\State Report the percentage of tokens the former saves over the latter.
\end{algorithmic}
\end{algorithm}

\section{Effect of Test-Time Reasoning}
\label{sec:app-nothink}

\begin{figure*}[t]
  \centering
  \setlength{\tabcolsep}{0pt}
  
  \begin{tabular}{@{} c c c c c @{}}
    
    \small{Qwen 3 VL-8B} & \small{Qwen 3.5-9B} & \small{Gemma\,4 E4B} & \small{Gemma\,4 26B-A4B} \\[1mm] 
    
    \raisebox{-0.5\height}{\includegraphics[width=0.23\linewidth]{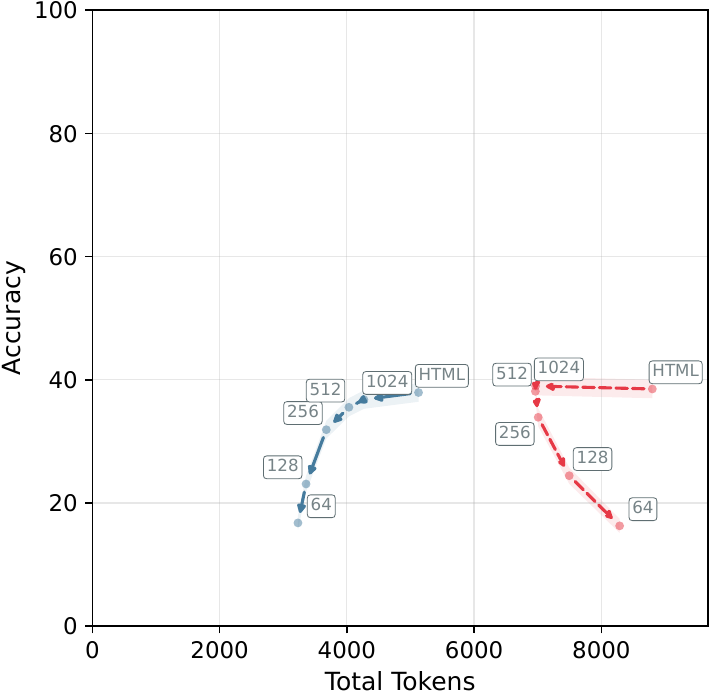}} &
    \raisebox{-0.5\height}{\includegraphics[width=0.23\linewidth]{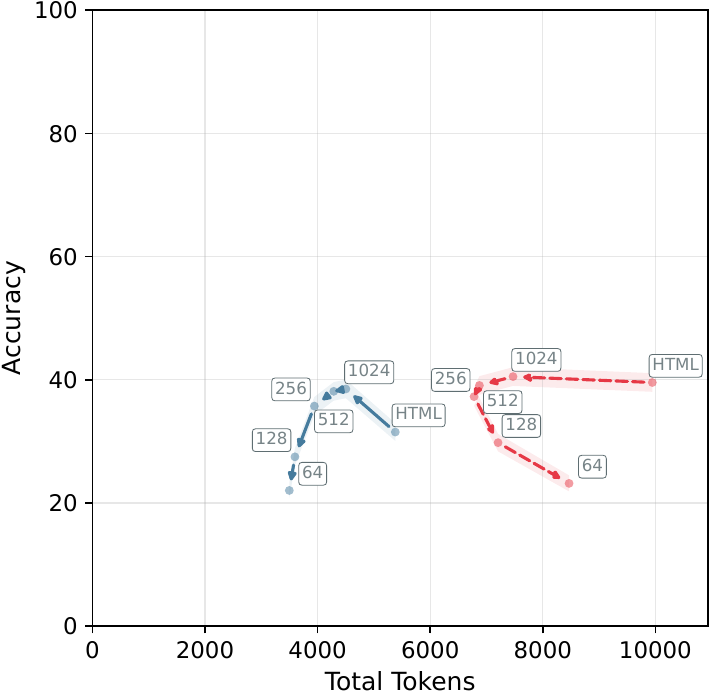}} &
    \raisebox{-0.5\height}{\includegraphics[width=0.23\linewidth]{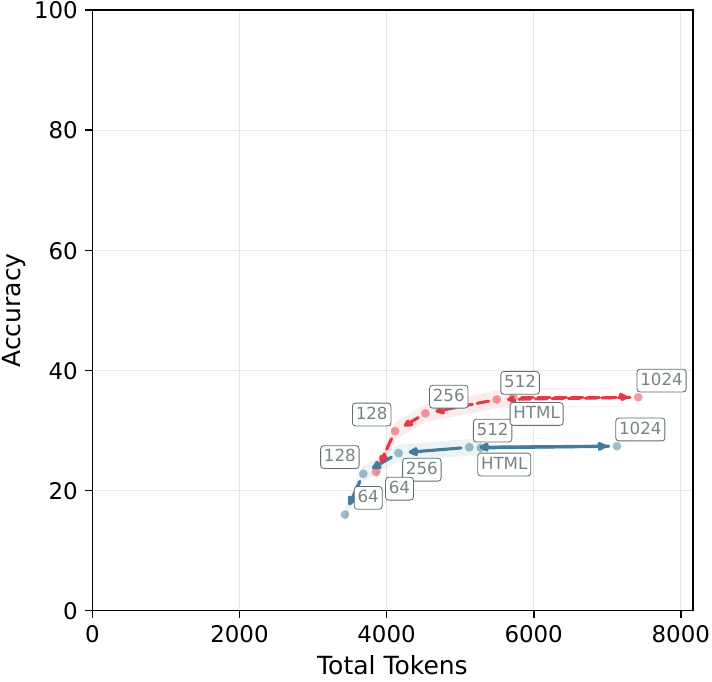}} &
    \raisebox{-0.5\height}{\includegraphics[width=0.23\linewidth]{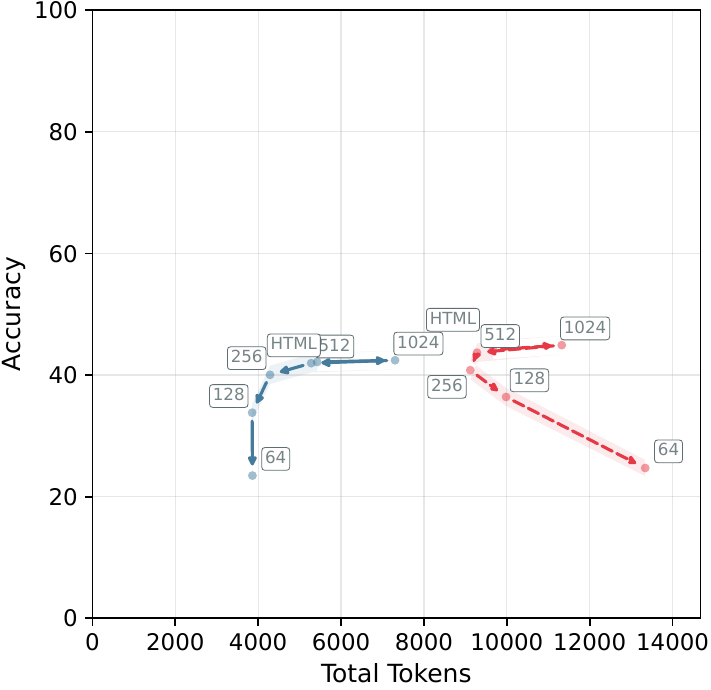}} &
  \end{tabular}
  
  \vspace{1.5ex} 
  \textcolor[HTML]{457B9D}{\Large $\bullet$} \small{Without test-time reasoning} \hspace{2em}
  \textcolor[HTML]{E63A46}{\Large $\bullet$} \small{With test-time reasoning} \hspace{2em}
  
  \caption{Direct QA with and without test-time reasoning across visual-token budgets on \mh{}, \arr{for the four reasoning-capable Gemma and Qwen models.} Compression widens the gap between the two conditions, as lower budgets induce longer reasoning traces.}
    \label{fig:app-nothink}
\end{figure*}

Figure~\ref{fig:app-nothink} compares direct QA with and without test-time reasoning across visual-token budgets.
Two patterns emerge.
First, the token cost of reasoning grows as tables are compressed, so the gap between the two conditions widens towards the lower budgets, which is the effect discussed in Section~\ref{sec:qa}.
Second, the accuracy benefit of reasoning varies by model, \arr{but is positive for most models}.
We enable test-time reasoning in the main experiments because both datasets emphasise numerical reasoning, and because it is the stronger of the two conditions for most models.

\section{Full Results}
\label{sec:app-all-results}

This appendix reports the complete results for Sections~\ref{sec:qa}, \ref{sec:perf-eff}, and~\ref{sec:rag}, for all models and both datasets.
Figures show accuracy against total tokens; tables give the underlying numbers, including the breakdown of token counts into their textual, visual, and generated tokens.
On \fldqa{} we evaluate \arr{Qwen 3.5-9B, Gemma\,4 26B-A4B, and Gemini 3.8 Flash as the 128k context windows of Gemma\,4 E4B and GLM-4.6V} cannot accommodate these documents alongside reasoning traces, and Qwen 3 VL-8B is omitted as each configuration is considerably more expensive to run at this length.

\subsection{Full-Context QA}

\begin{figure*}[p]
  \centering
  \setlength{\tabcolsep}{0pt}
  \resizebox{0.8\linewidth}{!}{%
    \begin{tabular}{@{} c c c c @{}}
      & \multicolumn{2}{c}{\small \mh{}} & \multicolumn{1}{c}{\small \fldqa{}} \\
      \cmidrule(lr){2-3} \cmidrule(lr){4-4}
      & \small{Qwen 3 VL-8B} & \small{Qwen 3.5-9B} & \small{Qwen 3.5-9B} \\
      &
      \raisebox{-0.5\height}{\includegraphics[width=0.30\linewidth]{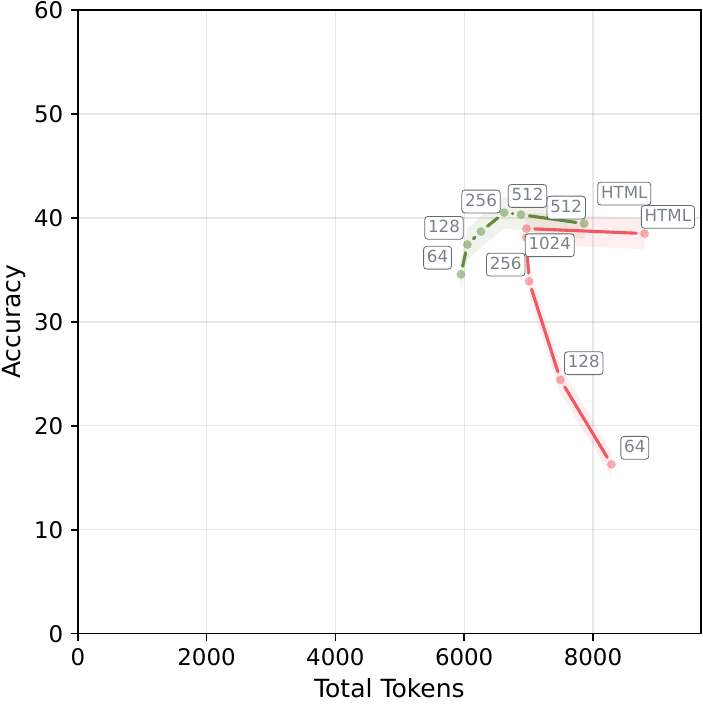}} &
      \raisebox{-0.5\height}{\includegraphics[width=0.30\linewidth]{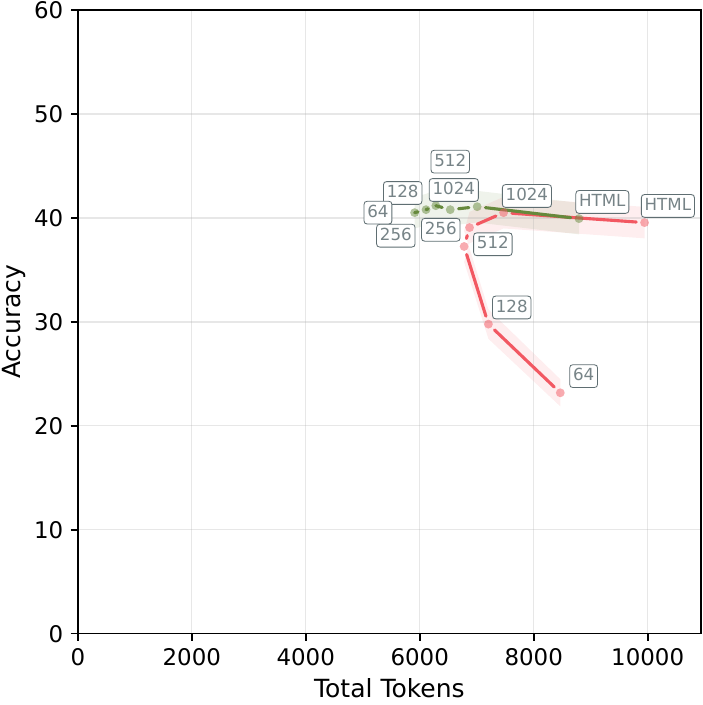}} &
      \raisebox{-0.5\height}{\includegraphics[width=0.30\linewidth]{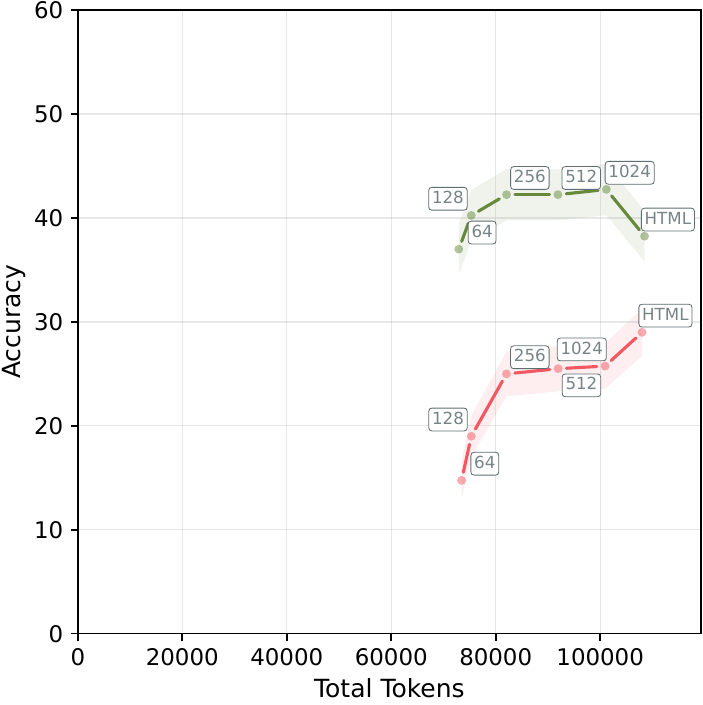}} \\[2mm]
      & \small{Gemma\,4 E4B} & \small{Gemma\,4 26B-A4B} & \small{Gemma\,4 26B-A4B} \\
      &
      \raisebox{-0.5\height}{\includegraphics[width=0.30\linewidth]{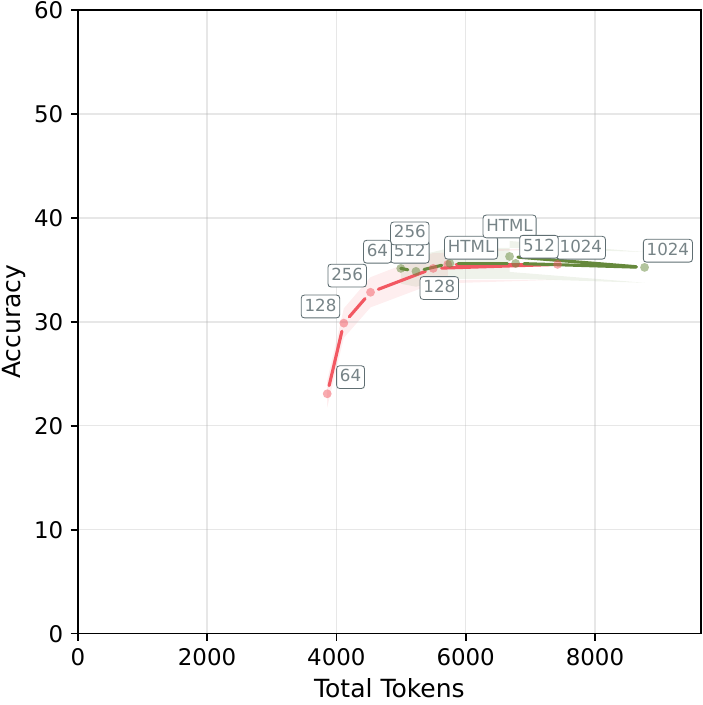}} &
      \raisebox{-0.5\height}{\includegraphics[width=0.30\linewidth]{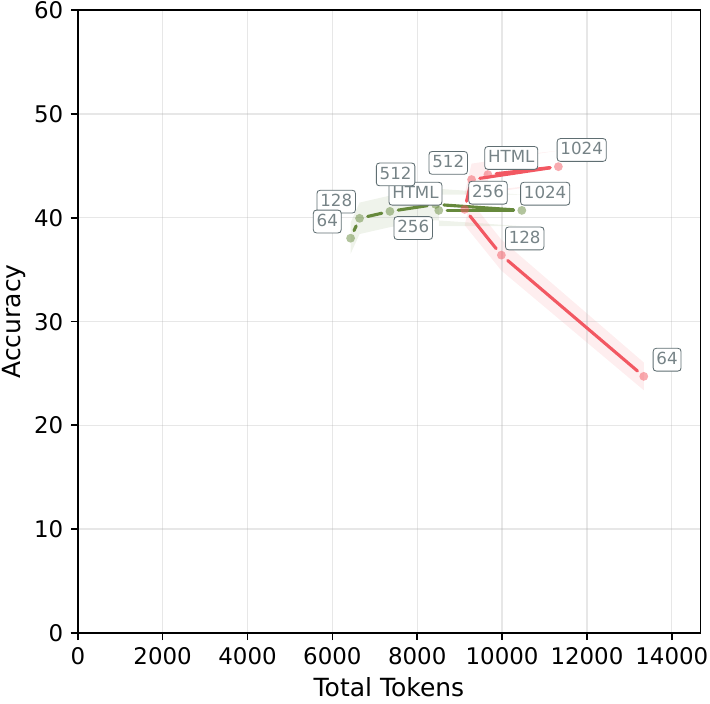}} &
      \raisebox{-0.5\height}{\includegraphics[width=0.30\linewidth]{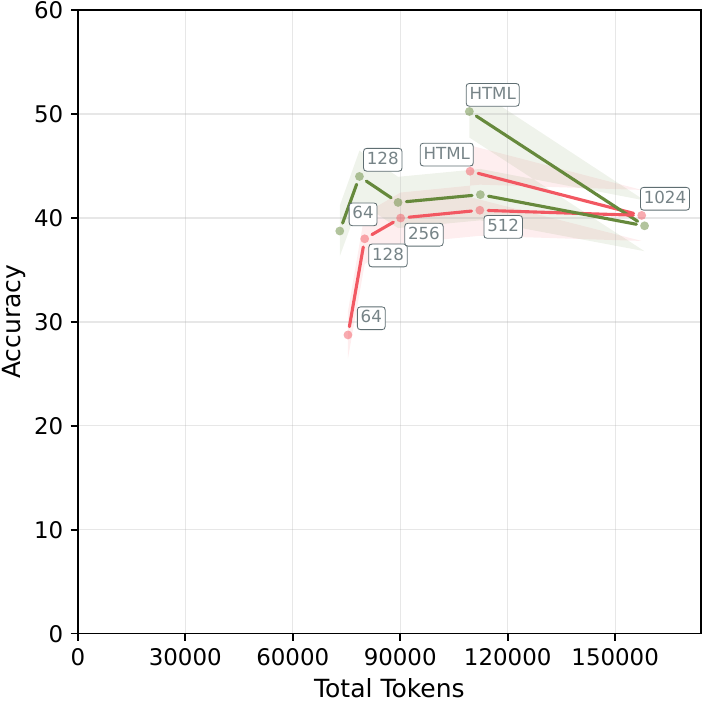}} \\
      & \small{GLM-4.6V} & \small{Gemini 3.8 Flash} & \small{Gemini 3.8 Flash} \\
      &
      \raisebox{-0.5\height}{\includegraphics[width=0.30\linewidth]{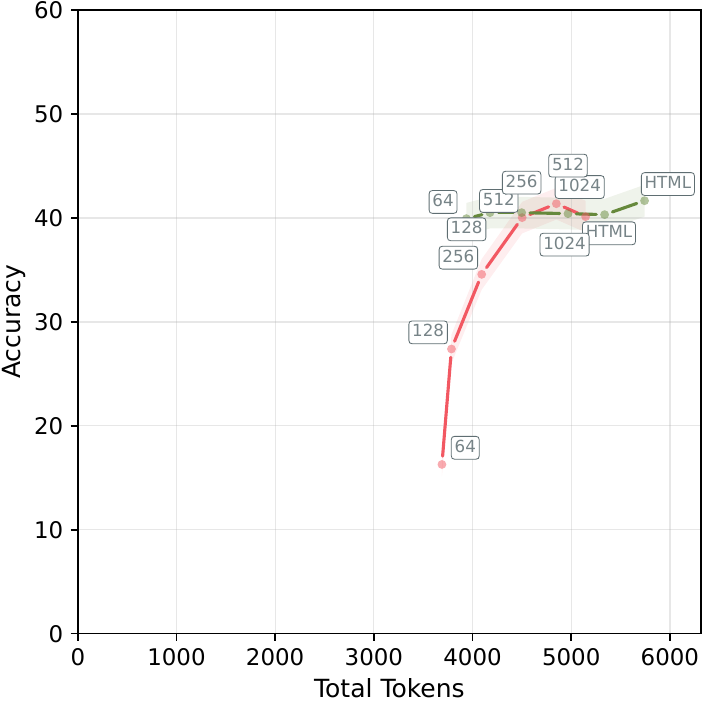}} &
      \raisebox{-0.5\height}{\includegraphics[width=0.30\linewidth]{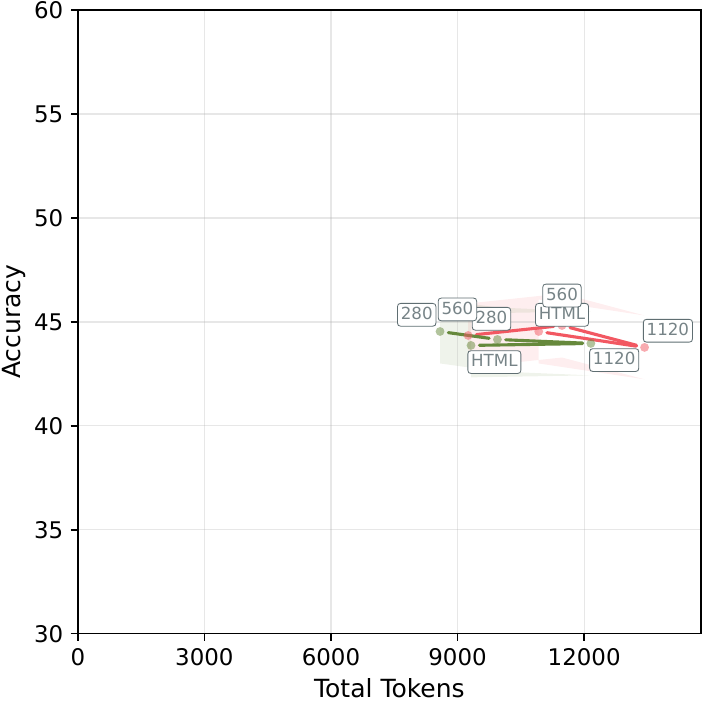}} &
      \raisebox{-0.5\height}{\includegraphics[width=0.30\linewidth]{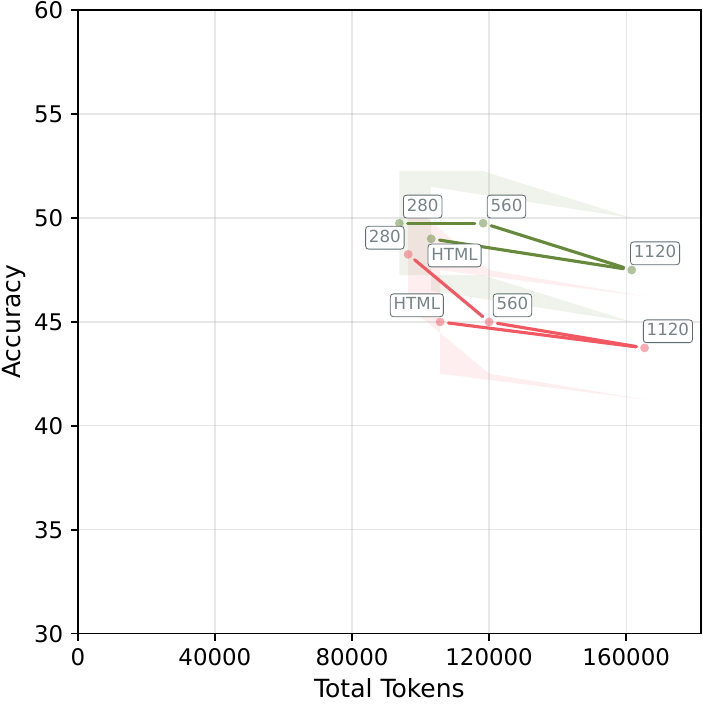}} \\
    \end{tabular}}
  
  \vspace{1.5ex} 
  \textcolor[HTML]{F25862}{\Large $\bullet$} \small{Direct QA} \hspace{2em}
  \textcolor[HTML]{66893C}{\Large $\bullet$} \small{Two-Stage QA (ours)} \hspace{2em}
  
  \caption{Accuracy against total tokens for direct QA and our two-stage method, on \mh{} \arr{(left two columns)} and \fldqa{} \arr{(right column)}, for \arr{Qwen 3 VL-8B, Qwen 3.5-9B, Gemma\,4 E4B, Gemma\,4 26B-A4B}, GLM-4.6V, and Gemini 3.8 Flash \arr{with the full document}. Markers along each line correspond to visual-token budgets, and bands show $\pm 1$~s.e.m.\ across examples (Appendix~\ref{app:ci}).}
  \label{fig:app-nothink3}

\vspace{1em}
  \includegraphics[width=\linewidth]{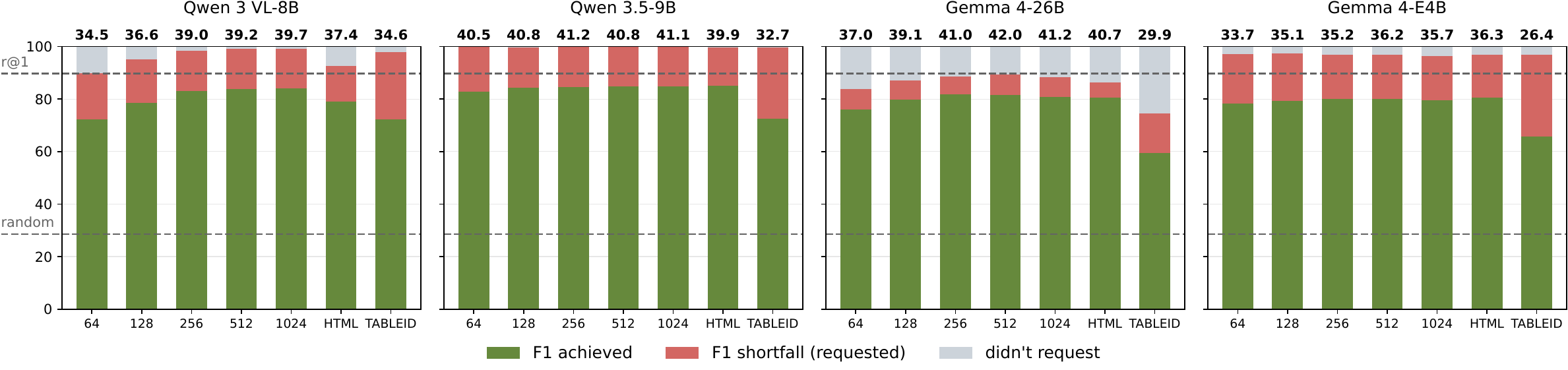}
  \caption{Relevant-table identification (F1) on \mh{} across token budgets and table representations, for all models. Bars show identification F1 (y~axis), decomposed into F1 achieved vs.\ shortfall on requested tables; \textsc{html} and \textsc{tableid} are references, and the horizontal lines mark random choice and ColQwen2@1, its best-F1 operating point. The percentage above each bar is the overall downstream QA accuracy.}
  \label{fig:app-ident-mh}
\end{figure*}

In this setting,  the complete document is provided to the model, with tables represented as images at each visual-token budget (Figure~\ref{fig:app-nothink3}).
Direct QA answers from this context in a single step, while our two-stage method first identifies the tables it needs and then reasons over them at native resolution.

\begin{figure*}[t]
  \centering
  \setlength{\tabcolsep}{0pt}
  \resizebox{0.8\linewidth}{!}{%
    \begin{tabular}{@{} c c c c @{}}
      & \multicolumn{2}{c}{\small \mh{}} & \multicolumn{1}{c}{\small \fldqa{}} \\
      \cmidrule(lr){2-3} \cmidrule(lr){4-4}
      & \small{Qwen 3 VL-8B} & \small{Qwen 3.5-9B} & \small{Qwen 3.5-9B} \\
       &
      \raisebox{-0.5\height}{\includegraphics[width=0.30\linewidth]{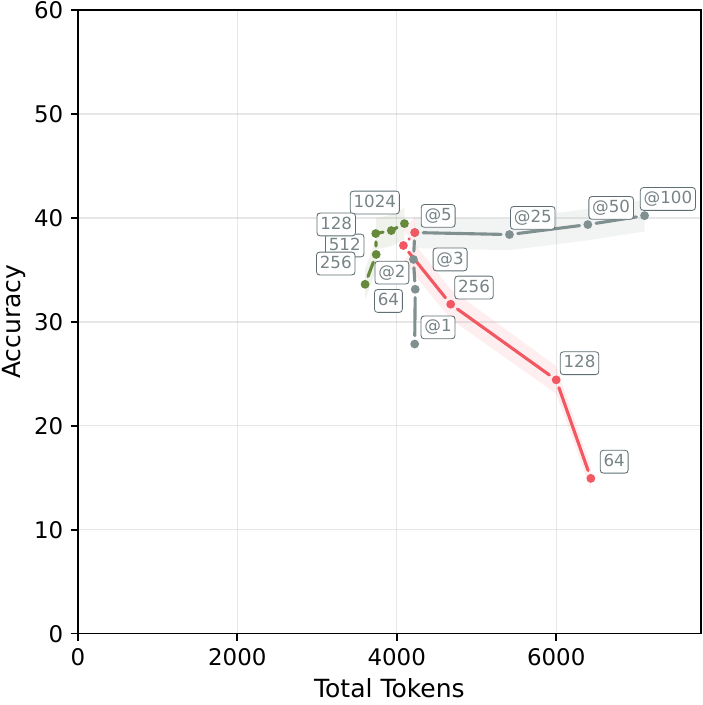}} &
      \raisebox{-0.5\height}{\includegraphics[width=0.30\linewidth]{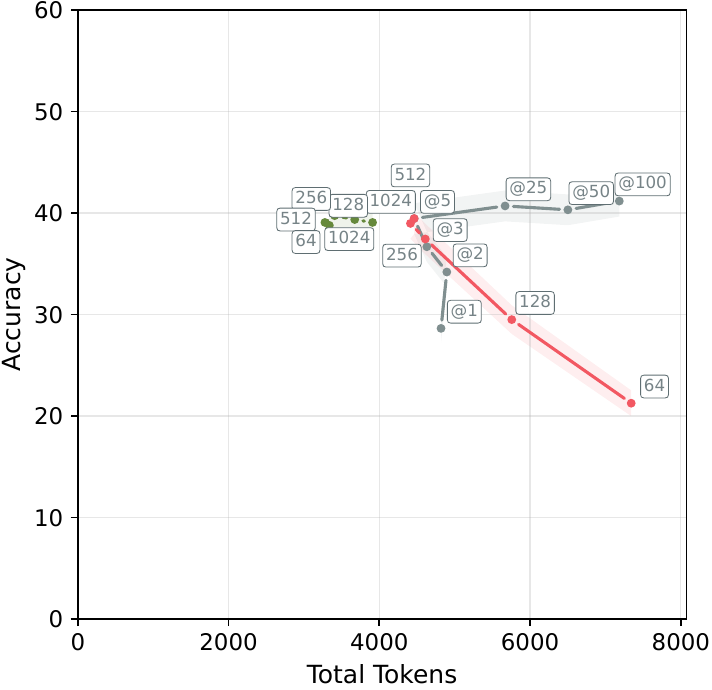}} &
      \raisebox{-0.5\height}{\includegraphics[width=0.30\linewidth]{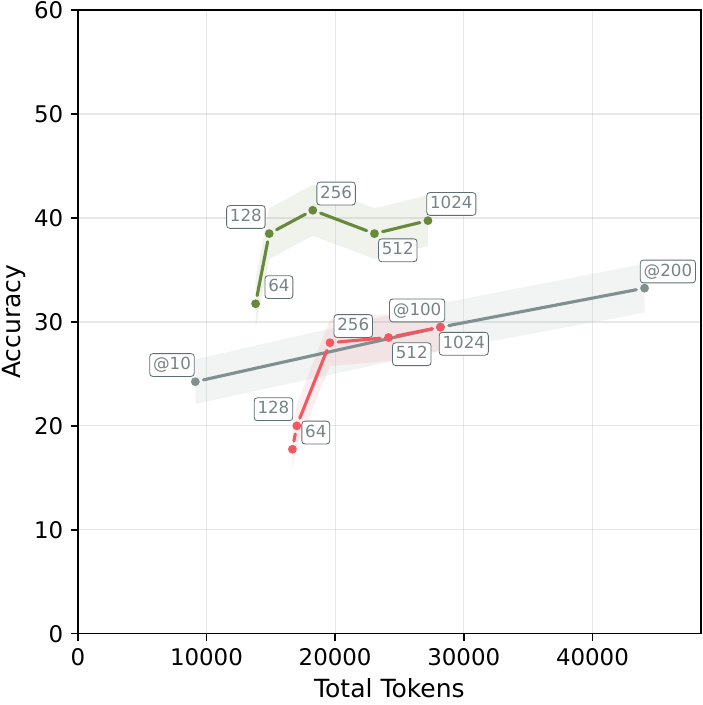}} \\[2mm]
      & \small{Gemma\,4 E4B} & \small{Gemma\,4 26B-A4B} & \small{Gemma\,4 26B-A4B} \\
       &
      \raisebox{-0.5\height}{\includegraphics[width=0.30\linewidth]{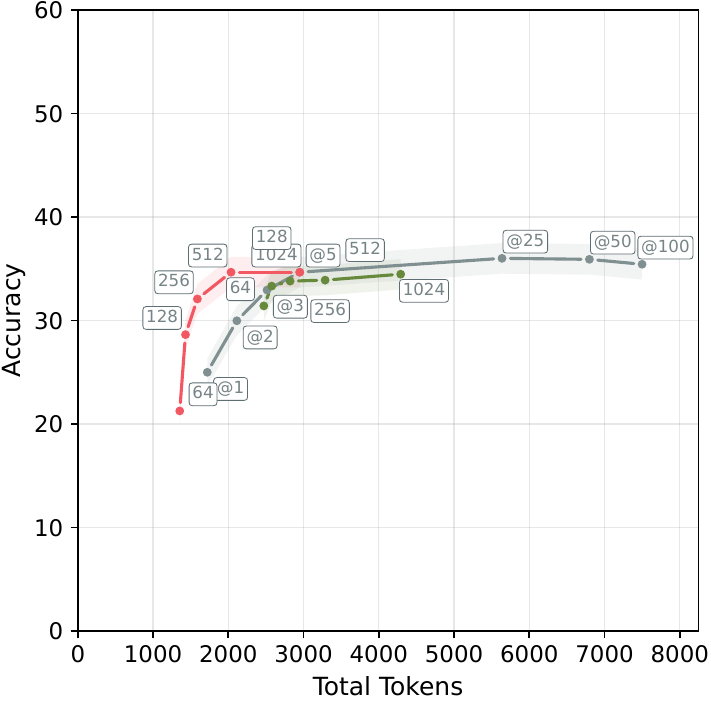}} &
      \raisebox{-0.5\height}{\includegraphics[width=0.30\linewidth]{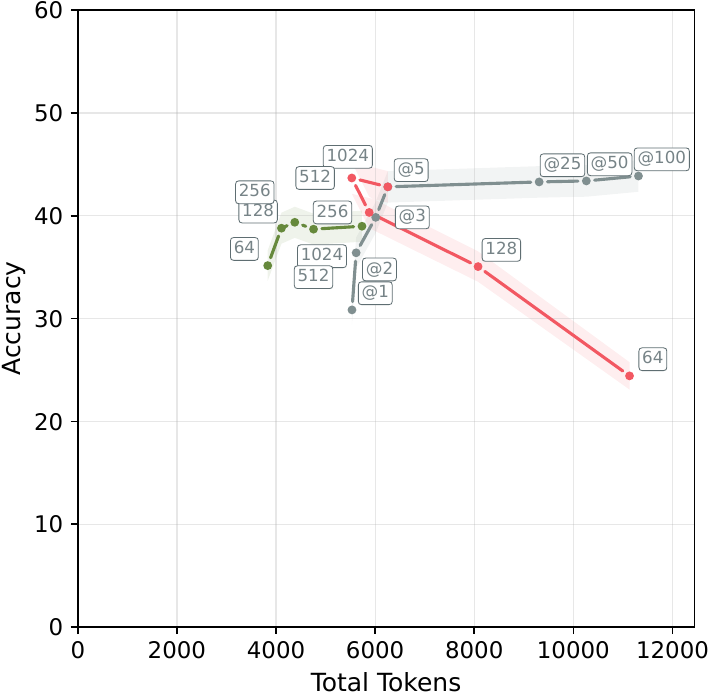}} &
      \raisebox{-0.5\height}{\includegraphics[width=0.30\linewidth]{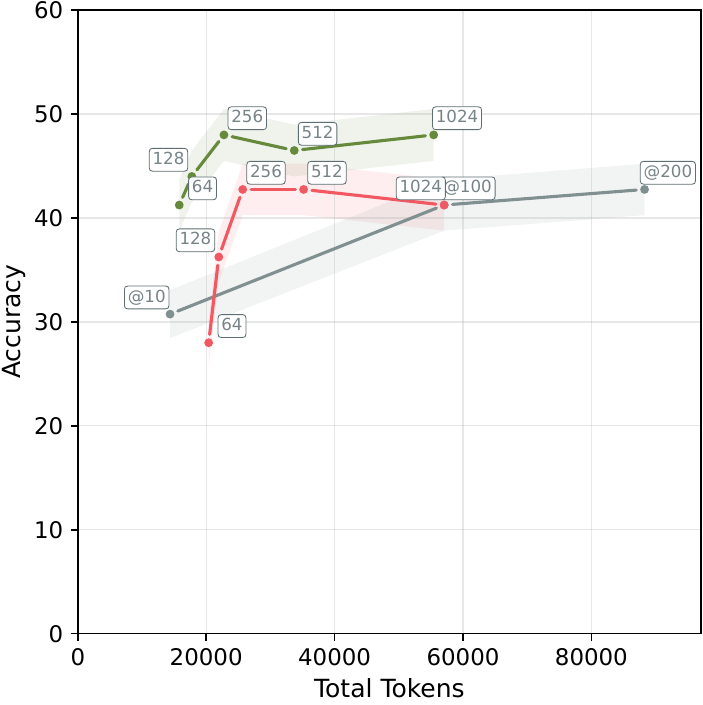}} \\
    \end{tabular}}
  
  \vspace{1.5ex} 
  \textcolor[HTML]{F25862}{\Large $\bullet$} \small{Direct QA} \hspace{2em}
  \textcolor[HTML]{66893C}{\Large $\bullet$} \small{Two-Stage QA (ours)} \hspace{2em}
  \textcolor[HTML]{808F90}{\Large $\bullet$} \small{BGE-M3 @k}
  
  \caption{Accuracy against total tokens for direct QA \arr{and} our two-stage method, on \mh{} \arr{(left two columns)} and \fldqa{} \arr{(right column)}, for \arr{Qwen 3 VL-8B, Qwen 3.5-9B, Gemma\,4 E4B, and Gemma\,4 26B-A4B}. We feed a context retrieved by BGE-M3, with single-step QA over the retrieved units at increasing $k$ as reference. Markers along each line correspond to visual-token budgets, and bands show $\pm 1$~s.e.m.\ across examples (Appendix~\ref{app:ci}).}
  \label{fig:app-nothink2}
\end{figure*}

\begin{table*}[t]
\centering
  \fontsize{8}{9.6}\selectfont
  \setlength{\tabcolsep}{3.5pt}
    \begin{tabular}{@{}ll rrrrr rrrrr@{}}
    \toprule
     & & \multicolumn{5}{c}{\textbf{Direct QA}} & \multicolumn{5}{c@{}}{\textbf{Two-stage QA (ours)}} \\
    \cmidrule(lr){3-7} \cmidrule(l){8-12}
    Model & Budget & Acc. & Text & Visual & Output & Total & Acc. & Text & Visual & Output & Total \\
    \midrule
    Qwen 3 VL-8B & -- & 38.5 & 4696 & 0 & 3766 & 8802 & 39.5 & 4981 & 0 & 1924 & 7401 \\
     & 1024 & 39.0 & 2583 & 1358 & 2731 & 6966 & 40.3 & 2876 & 1830 & 1854 & 6882 \\
     & 512 & 38.1 & 2583 & 1070 & 3106 & 6964 & 40.5 & 2876 & 1620 & 1916 & 6621 \\
     & 256 & 33.9 & 2583 & 717 & 3619 & 7008 & 38.7 & 2876 & 1265 & 1944 & 6262 \\
     & 128 & 24.4 & 2583 & 425 & 4368 & 7496 & 37.5 & 2876 & 894 & 1961 & 6050 \\
     & 64 & 16.3 & 2583 & 292 & 5252 & 8288 & 34.6 & 2876 & 720 & 2120 & 5950 \\
    \midrule
    Qwen 3.5-9B & -- & 39.6 & 4765 & 0 & 4958 & 9946 & 39.9 & 6064 & 0 & 2190 & 8795 \\
     & 1024 & 40.5 & 2650 & 1358 & 2974 & 7472 & 41.1 & 2959 & 1891 & 1772 & 7008 \\
     & 512 & 39.1 & 2650 & 1070 & 2818 & 6874 & 40.8 & 2962 & 1716 & 1695 & 6536 \\
     & 256 & 37.3 & 2650 & 717 & 3140 & 6780 & 41.2 & 2960 & 1338 & 1824 & 6282 \\
     & 128 & 29.8 & 2650 & 425 & 3804 & 7206 & 40.8 & 2960 & 984 & 1882 & 6110 \\
     & 64 & 23.2 & 2650 & 292 & 5238 & 8466 & 40.5 & 2959 & 822 & 1868 & 5910 \\
    \midrule
    \arr{Gemma\,4 E4B} & -- & 35.5 & 4756 & 0 & 790 & 5723 & 36.3 & 5978 & 0 & 560 & 6679 \\
     & 1024 & 35.5 & 2632 & 4292 & 688 & 7422 & 35.2 & 2946 & 5372 & 540 & 8768 \\
     & 512 & 35.2 & 2632 & 2114 & 758 & 5498 & 35.6 & 2945 & 3199 & 545 & 6773 \\
     & 256 & 32.9 & 2632 & 1028 & 786 & 4528 & 35.6 & 2945 & 2105 & 536 & 5755 \\
     & 128 & 29.9 & 2632 & 494 & 839 & 4115 & 34.9 & 2940 & 1576 & 539 & 5232 \\
     & 64 & 23.1 & 2632 & 235 & 902 & 3858 & 35.2 & 2948 & 1312 & 540 & 4996 \\
    \midrule
    \arr{Gemma\,4 26B-A4B} & -- & 44.2 & 4756 & 0 & 4332 & 9666 & 40.7 & 5983 & 0 & 1869 & 8508 \\
     & 1024 & 44.9 & 2632 & 4292 & 3874 & 11326 & 40.7 & 2930 & 5378 & 1718 & 10465 \\
     & 512 & 43.7 & 2632 & 2114 & 4112 & 9280 & 41.3 & 2942 & 3206 & 1874 & 8432 \\
     & 256 & 40.8 & 2632 & 1028 & 5138 & 9118 & 40.6 & 2940 & 2106 & 1836 & 7356 \\
     & 128 & 36.4 & 2632 & 494 & 6739 & 9982 & 39.9 & 2936 & 1576 & 1764 & 6640 \\
     & 64 & 24.7 & 2632 & 235 & 10288 & 13339 & 38.0 & 2934 & 1310 & 1734 & 6430 \\
    \midrule
    GLM-4.6V & -- & 40.1 & 4312 & 0 & 517 & 5146 & 41.7 & 4602 & 0 & 267 & 5044 \\
     & 1024 & 41.4 & 2413 & 1600 & 606 & 4850 & 40.3 & 2704 & 2340 & 259 & 5338 \\
     & 512 & 40.0 & 2413 & 1244 & 612 & 4503 & 40.4 & 2704 & 1976 & 260 & 4968 \\
     & 256 & 34.6 & 2413 & 746 & 642 & 4093 & 40.5 & 2702 & 1522 & 261 & 4498 \\
     & 128 & 27.4 & 2413 & 435 & 744 & 3786 & 40.5 & 2704 & 1146 & 268 & 4177 \\
     & 64 & 16.3 & 2413 & 216 & 820 & 3690 & 39.9 & 2704 & 908 & 262 & 3938 \\
    \midrule
    Gemini 3.8 Flash & -- & 44.5 & 4056 & 0 & 6316 & 10922 & 43.9 & 4389 & 0 & 2866 & 8230 \\
     & 1120 & 43.8 & 2608 & 4292 & 5711 & 13431 & 44.0 & 2956 & 5394 & 3012 & 12158 \\
     & 560 & 44.8 & 2608 & 2114 & 6408 & 11474 & 44.2 & 2939 & 3216 & 3040 & 9944 \\
     & 280 & 44.3 & 2608 & 1028 & 5223 & 9250 & 44.5 & 2940 & 2125 & 2764 & 8584 \\
    \bottomrule
    \end{tabular}
  \caption{Full-context results on \mh{}, for direct QA (left) and our two-stage method (right). \emph{Budget} is the visual-token budget applied to each table image, with `--' denoting the \textsc{html} setting in which tables are serialised as text. \emph{Acc.} is exact-match accuracy; the remaining columns report median token counts per example, split into input text tokens, input visual tokens, and generated tokens\arr{; \emph{Total} is the median per-example total}. \arr{For Gemini, 1{,}120, 560 and 280 are the API's resolution levels; the visual-token counts it reports coincide with Gemma\,4's.}}
\label{tab:app-full-mh}
\end{table*}

\begin{table*}[t]
\centering
  \fontsize{8}{9.6}\selectfont
  \setlength{\tabcolsep}{3pt}
    \begin{tabular}{@{}ll rrrrr rrrrr@{}}
    \toprule
     & & \multicolumn{5}{c}{\textbf{Direct QA}} & \multicolumn{5}{c@{}}{\textbf{Two-stage QA (ours)}} \\
    \cmidrule(lr){3-7} \cmidrule(l){8-12}
    Model & Budget & Acc. & Text & Visual & Output & Total & Acc. & Text & Visual & Output & Total \\
    \midrule
    Qwen 3.5-9B & -- & 29.0 & 102782 & 0 & 4760 & 108016 & 38.2 & 104222 & 0 & 3949 & 108494 \\
     & 1024 & 25.8 & 61050 & 33444 & 3690 & 100936 & 42.8 & 61973 & 33732 & 2342 & 101184 \\
     & 512 & 25.5 & 61050 & 24916 & 3844 & 91947 & 42.2 & 61974 & 25564 & 2434 & 91910 \\
     & 256 & 25.0 & 61050 & 15950 & 4116 & 82064 & 42.2 & 61974 & 16652 & 2336 & 82077 \\
     & 128 & 19.0 & 61050 & 8914 & 5006 & 75332 & 40.2 & 61974 & 9398 & 2404 & 75324 \\
     & 64 & 14.8 & 61050 & 6064 & 5252 & 73452 & 37.0 & 61974 & 6562 & 2923 & 72930 \\
    \midrule
    \arr{Gemma\,4 26B-A4B} & -- & 44.5 & 102562 & 0 & 4984 & 109496 & 50.2 & 104154 & 0 & 2796 & 109296 \\
     & 1024 & 40.2 & 60829 & 85892 & 4642 & 157397 & 39.2 & 61526 & 87358 & 2424 & 158204 \\
     & 512 & 40.8 & 60829 & 42340 & 5398 & 112222 & 42.2 & 61576 & 43566 & 2528 & 112364 \\
     & 256 & 40.0 & 60829 & 20630 & 5686 & 90110 & 41.5 & 61574 & 22242 & 2550 & 89392 \\
     & 128 & 38.0 & 60829 & 9990 & 6712 & 80083 & 44.0 & 61592 & 11422 & 2864 & 78616 \\
     & 64 & 28.7 & 60829 & 4814 & 8182 & 75411 & 38.8 & 61554 & 6226 & 2660 & 73124 \\
    \midrule
    Gemini 3.8 Flash & -- & 45.0 & 91196 & 0 & 13225 & 105726 & 49.0 & 92268 & 0 & 7196 & 102022 \\
     & 1120 & 43.8 & 60647 & 85892 & 14590 & 165356 & 47.5 & 61506 & 87382 & 7578 & 161604 \\
     & 560 & 45.0 & 60647 & 42340 & 14808 & 120001 & 49.8 & 61507 & 43597 & 7658 & 118259 \\
     & 280 & 48.2 & 60647 & 20630 & 13784 & 96386 & 49.8 & 61530 & 21990 & 7248 & 93814 \\
    \bottomrule
    \end{tabular}
  \caption{Full-context results on \fldqa{}\arr{, for direct QA (left) and our two-stage method (right)}. \emph{Budget} is the visual-token budget applied to each table image, with `--' denoting the \textsc{html} setting in which tables are serialised as text. \emph{Acc.} is exact-match accuracy; the remaining columns report median token counts per example, split into input text tokens, input visual tokens, and generated tokens\arr{; \emph{Total} is the median per-example total}. Gemma\,4~E4B\arr{, GLM-4.6V} and Qwen 3 VL-8B are omitted, as this dataset exceeds the context window of the \arr{first two} and makes every configuration considerably more expensive to run \arr{for the last}. \arr{Gemini's visual-token counts are explained in Table~\ref{tab:app-full-mh}.}}
\label{tab:app-full-fldqa}
\end{table*}

\subsection{QA over Retrieved Context}

In this setting, the context is assembled by BGE-M3 rather than fed in full, using $k{=}5$ on \mh{} and $k{=}100$ on \fldqa{} (Figure~\ref{fig:app-nothink2}).
Direct QA and our two-stage method operate over this reduced context, with single-step QA over retrieved units at increasing $k$ included as a reference for what retrieval alone achieves.

\begin{table*}[t]
\centering
  \fontsize{8}{9.6}\selectfont
  \setlength{\tabcolsep}{3pt}
    \begin{tabular}{lllrrrrr}
    \toprule
    Model & Budget & Task & Acc. & Input text & Input visual & Output text & Total \\
    \midrule
    Qwen 3 VL-8B & 1024@100 & Re. QA & 40.2 & 2583 & 1358 & 2848 & 7108 \\
    Qwen 3 VL-8B & 1024@50 & Re. QA & 39.4 & 1927 & 1348 & 2994 & 6396 \\
    Qwen 3 VL-8B & 1024@25 & Re. QA & 38.4 & 1091 & 1260 & 2983 & 5414 \\
    Qwen 3 VL-8B & 1024@5 & Re. QA & 38.6 & 358 & 768 & 3140 & 4226 \\
    Qwen 3 VL-8B & 512@5 & Re. QA & 37.4 & 358 & 539 & 3142 & 4084 \\
    Qwen 3 VL-8B & 256@5 & Re. QA & 31.7 & 358 & 350 & 3952 & 4676 \\
    Qwen 3 VL-8B & 128@5 & Re. QA & 24.4 & 358 & 216 & 5402 & 6000 \\
    Qwen 3 VL-8B & 64@5 & Re. QA & 14.9 & 358 & 144 & 5928 & 6434 \\
    Qwen 3 VL-8B & 1024@3 & Re. QA & 36.0 & 290 & 616 & 3235 & 4210 \\
    Qwen 3 VL-8B & 1024@2 & Re. QA & 33.1 & 255 & 531 & 3420 & 4232 \\
    Qwen 3 VL-8B & 1024@1 & Re. QA & 27.9 & 221 & 336 & 3752 & 4225 \\
    \midrule
    Qwen 3.5-9B & 1024@100 & Re. QA & 41.2 & 2650 & 1358 & 2889 & 7184 \\
    Qwen 3.5-9B & 1024@50 & Re. QA & 40.3 & 1978 & 1348 & 2991 & 6502 \\
    Qwen 3.5-9B & 1024@25 & Re. QA & 40.7 & 1122 & 1260 & 3201 & 5668 \\
    Qwen 3.5-9B & 1024@5 & Re. QA & 39.5 & 372 & 768 & 3341 & 4462 \\
    Qwen 3.5-9B & 512@5 & Re. QA & 39.0 & 372 & 539 & 3478 & 4413 \\
    Qwen 3.5-9B & 256@5 & Re. QA & 37.5 & 372 & 350 & 3811 & 4610 \\
    Qwen 3.5-9B & 128@5 & Re. QA & 29.5 & 372 & 216 & 5169 & 5757 \\
    Qwen 3.5-9B & 64@5 & Re. QA & 21.3 & 372 & 144 & 6850 & 7342 \\
    Qwen 3.5-9B & 1024@3 & Re. QA & 36.7 & 302 & 616 & 3775 & 4630 \\
    Qwen 3.5-9B & 1024@2 & Re. QA & 34.2 & 267 & 531 & 4208 & 4896 \\
    Qwen 3.5-9B & 1024@1 & Re. QA & 28.6 & 232 & 336 & 4318 & 4818 \\
    \midrule
    \arr{Gemma\,4 E4B} & 1024@100 & Re. QA & 35.4 & 2632 & 4292 & 744 & 7498 \\
    \arr{Gemma\,4 E4B} & 1024@50 & Re. QA & 35.9 & 1979 & 4284 & 728 & 6798 \\
    \arr{Gemma\,4 E4B} & 1024@25 & Re. QA & 36.0 & 1126 & 3276 & 758 & 5637 \\
    \arr{Gemma\,4 E4B} & 1024@5 & Re. QA & 34.7 & 381 & 2144 & 719 & 2950 \\
    \arr{Gemma\,4 E4B} & 512@5 & Re. QA & 34.7 & 381 & 1048 & 726 & 2037 \\
    \arr{Gemma\,4 E4B} & 256@5 & Re. QA & 32.1 & 381 & 510 & 751 & 1589 \\
    \arr{Gemma\,4 E4B} & 128@5 & Re. QA & 28.6 & 381 & 243 & 836 & 1431 \\
    \arr{Gemma\,4 E4B} & 64@5 & Re. QA & 21.3 & 381 & 116 & 855 & 1355 \\
    \arr{Gemma\,4 E4B} & 1024@3 & Re. QA & 33.0 & 311 & 1100 & 686 & 2514 \\
    \arr{Gemma\,4 E4B} & 1024@2 & Re. QA & 30.0 & 276 & 1085 & 671 & 2114 \\
    \arr{Gemma\,4 E4B} & 1024@1 & Re. QA & 25.0 & 241 & 1078 & 656 & 1720 \\
    \midrule
    \arr{Gemma\,4 26B-A4B} & 1024@100 & Re. QA & 43.9 & 2632 & 4292 & 4117 & 11314 \\
    \arr{Gemma\,4 26B-A4B} & 1024@50 & Re. QA & 43.4 & 1979 & 4284 & 4036 & 10262 \\
    \arr{Gemma\,4 26B-A4B} & 1024@25 & Re. QA & 43.3 & 1126 & 3276 & 4257 & 9310 \\
    \arr{Gemma\,4 26B-A4B} & 1024@5 & Re. QA & 42.8 & 381 & 2144 & 3955 & 6256 \\
    \arr{Gemma\,4 26B-A4B} & 512@5 & Re. QA & 43.7 & 381 & 1048 & 4049 & 5529 \\
    \arr{Gemma\,4 26B-A4B} & 256@5 & Re. QA & 40.3 & 381 & 510 & 5063 & 5880 \\
    \arr{Gemma\,4 26B-A4B} & 128@5 & Re. QA & 35.1 & 381 & 243 & 7495 & 8077 \\
    \arr{Gemma\,4 26B-A4B} & 64@5 & Re. QA & 24.4 & 381 & 116 & 10678 & 11133 \\
    \arr{Gemma\,4 26B-A4B} & 1024@3 & Re. QA & 39.8 & 311 & 1100 & 4368 & 6011 \\
    \arr{Gemma\,4 26B-A4B} & 1024@2 & Re. QA & 36.4 & 276 & 1085 & 4201 & 5616 \\
    \arr{Gemma\,4 26B-A4B} & 1024@1 & Re. QA & 30.8 & 241 & 1078 & 4766 & 5532 \\
    \midrule
    GLM-4.6V & 1024@100 & Re. QA & 41.2 & 2413 & 1600 & 570 & 4856 \\
    GLM-4.6V & 1024@50 & Re. QA & 40.9 & 1792 & 1589 & 566 & 4168 \\
    GLM-4.6V & 1024@25 & Re. QA & 41.1 & 996 & 1518 & 613 & 3346 \\
    GLM-4.6V & 1024@5 & Re. QA & 39.8 & 323 & 972 & 597 & 2022 \\
    GLM-4.6V & 512@5 & Re. QA & 38.5 & 323 & 611 & 662 & 1759 \\
    GLM-4.6V & 256@5 & Re. QA & 33.5 & 323 & 384 & 699 & 1483 \\
    GLM-4.6V & 128@5 & Re. QA & 26.7 & 323 & 218 & 773 & 1304 \\
    GLM-4.6V & 64@5 & Re. QA & 15.0 & 323 & 104 & 898 & 1331 \\
    GLM-4.6V & 1024@3 & Re. QA & 36.6 & 261 & 808 & 614 & 1794 \\
    GLM-4.6V & 1024@2 & Re. QA & 33.4 & 231 & 688 & 630 & 1670 \\
    GLM-4.6V & 1024@1 & Re. QA & 27.8 & 200 & 443 & 693 & 1428 \\
    \midrule
    Gemini 3.8 Flash & 1120@5 & Re. QA & 43.1 & 363 & 2143 & 5972 & 8288 \\
    Gemini 3.8 Flash & 560@5 & Re. QA & 42.8 & 363 & 1048 & 6211 & 7383 \\
    Gemini 3.8 Flash & 280@5 & Re. QA & 42.1 & 363 & 510 & 5950 & 6665 \\
    \bottomrule
    \end{tabular}
    
  \caption{Results on \mh{} with a context retrieved by BGE-M3. \emph{Budget} reports the visual-token budget applied to each table image and the number of retrieved units $k$, in the form budget@$k$; \emph{Task} is the QA setting. \emph{Acc.} is exact-match accuracy; the remaining columns report median token counts per example, split into input text tokens, input visual tokens, and generated tokens\arr{; \emph{Total} is the median per-example total}.}
\label{tab:app-retr-mh}
\end{table*}

\begin{table*}[t]
\centering
  \fontsize{8}{9.6}\selectfont
  \setlength{\tabcolsep}{3pt}
    \begin{tabular}{lllrrrrr}
    \toprule
    Model & Budget & Task & Acc. & Input text & Input visual & Output text & Total \\
    \midrule
    Qwen 3.5-9B & 1024@200 & Re. QA & 33.2 & 14574 & 24480 & 3188 & 44049 \\
    Qwen 3.5-9B & 1024@100 & Re. QA & 29.5 & 6266 & 16518 & 3908 & 28185 \\
    Qwen 3.5-9B & 512@100 & Re. QA & 28.5 & 6266 & 12732 & 4194 & 24146 \\
    Qwen 3.5-9B & 256@100 & Re. QA & 28.0 & 6266 & 8208 & 4534 & 19597 \\
    Qwen 3.5-9B & 128@100 & Re. QA & 20.0 & 6266 & 4434 & 5842 & 17019 \\
    Qwen 3.5-9B & 64@100 & Re. QA & 17.8 & 6266 & 3006 & 6892 & 16682 \\
    Qwen 3.5-9B & 1024@10 & Re. QA & 24.2 & 782 & 1870 & 6042 & 9134 \\
    \midrule
    \arr{Gemma\,4 26B-A4B} & 1024@200 & Re. QA & 42.8 & 14688 & 64844 & 5413 & 88282 \\
    \arr{Gemma\,4 26B-A4B} & 1024@100 & Re. QA & 41.2 & 6370 & 42202 & 6561 & 57069 \\
    \arr{Gemma\,4 26B-A4B} & 512@100 & Re. QA & 42.8 & 6370 & 20856 & 7146 & 35174 \\
    \arr{Gemma\,4 26B-A4B} & 256@100 & Re. QA & 42.8 & 6370 & 10221 & 8605 & 25687 \\
    \arr{Gemma\,4 26B-A4B} & 128@100 & Re. QA & 36.2 & 6370 & 4969 & 9766 & 21960 \\
    \arr{Gemma\,4 26B-A4B} & 64@100 & Re. QA & 28.0 & 6370 & 2372 & 11360 & 20384 \\
    \arr{Gemma\,4 26B-A4B} & 1024@10 & Re. QA & 30.8 & 804 & 4351 & 8650 & 14367 \\
    \midrule
    Gemini 3.8 Flash & 1120@100 & Re. QA & 43.0 & 6265 & 42202 & 19219 & 68896 \\
    Gemini 3.8 Flash & 560@100 & Re. QA & 46.5 & 6265 & 20856 & 17093 & 45346 \\
    Gemini 3.8 Flash & 280@100 & Re. QA & 45.0 & 6265 & 10221 & 16401 & 33174 \\
    \bottomrule
    \end{tabular}
  \caption{Results on \fldqa{} with a context retrieved by BGE-M3. \emph{Budget} reports the visual-token budget applied to each table image and the number of retrieved units $k$, in the form budget@$k$; \emph{Task} is the QA setting. \emph{Acc.} is exact-match accuracy; the remaining columns report median token counts per example, split into input text tokens, input visual tokens, and generated tokens\arr{; \emph{Total} is the median per-example total}. Gemma\,4~E4B\arr{, GLM-4.6V} and Qwen 3 VL-8B are omitted for the reasons given in Table~\ref{tab:app-full-fldqa}. \arr{Gemini's visual-token counts are explained in Table~\ref{tab:app-full-mh}.}}
\label{tab:app-retr-fldqa}
\end{table*}

\section{Table Identification: All Models}
\label{sec:app-identification-all}

\begin{figure}[t]
  \centering
  \includegraphics[width=\linewidth]{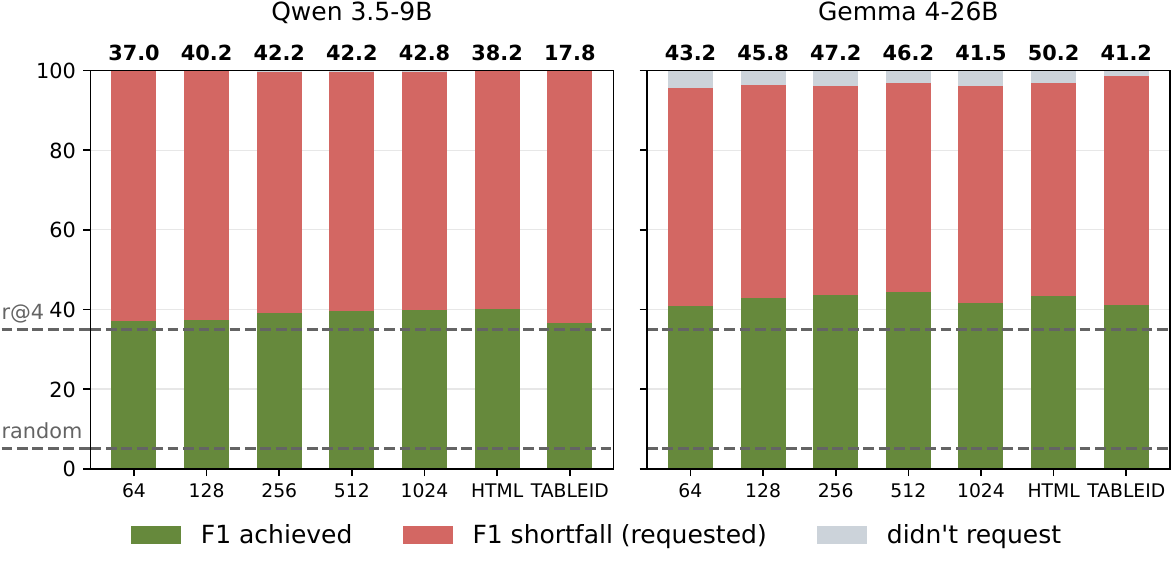}
  \caption{Relevant-table identification (F1) on \fldqa{} for \arr{Qwen 3.5-9B and Gemma\,4 26B-A4B}. Bars and references are as in Figure~\ref{fig:app-ident-mh}, except that the retrieval line marks ColQwen2@4, its best-F1 operating point on this dataset. \fldqa{} annotates evidence at the page level, so every table on an evidence page counts as relevant\arr{, and the resulting scores are therefore not directly comparable to those on \mh{}}. These results are shown for reference and are not used to support any claim in the paper.}
  \label{fig:app-ident-fldqa}
\end{figure}

Figures~\ref{fig:app-ident-mh} and~\ref{fig:app-ident-fldqa} extend the identification results of Section~\ref{sec:identification-results} to every model.
Robustness to compression holds across the board: identification varies little across visual-token budgets for all models.
The gain over \textsc{tableid} is clear for Qwen 3.5 and the Gemma\,4 models, but marginal for Qwen 3 VL.

On \fldqa{}, identification F1 is almost identical between low image budgets and \textsc{tableid}.
As noted in Section~\ref{sec:identification-results}, this dataset annotates evidence at the page level, so treating every table on an evidence page as relevant yields \arr{a superset of} the true evidence set. 
The downstream accuracies printed above each bar tell a different story.
Removing table contents costs Qwen 3.5 a large amount of accuracy (37.0 at 64 tokens against 17.8 for \textsc{tableid}) and Gemma\,4 26B-A4B a smaller one (38.8 against 37.8), even though the two conditions are indistinguishable under F1.
This is consistent with the annotation being too coarse to separate them: a model that locates the right page scores well whether or not it can tell which of its tables matters, while the answer still depends on getting that distinction right.

\section{At What Resolution to Return the Requested Table}
\label{sec:app-tool-budget}

\begin{figure}[t]
  \centering
  \includegraphics[width=0.49\linewidth]{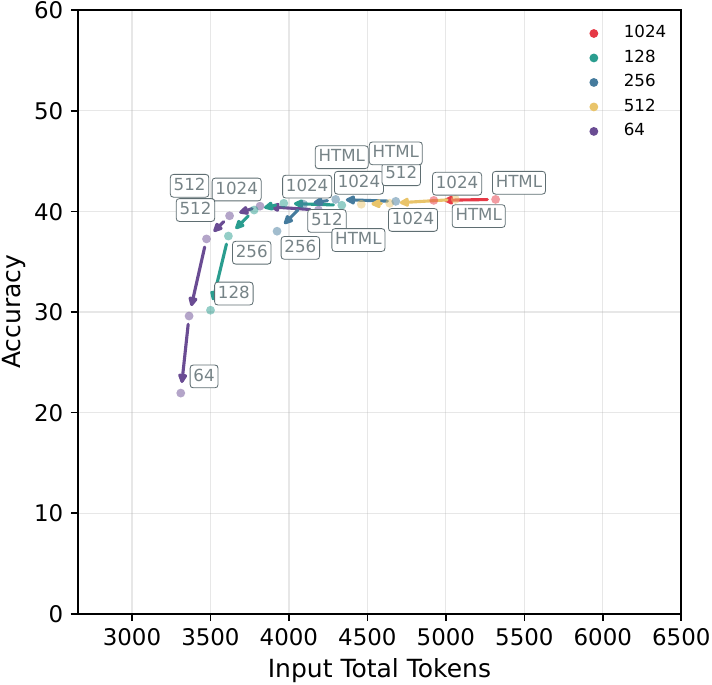}\hfill
  \includegraphics[width=0.49\linewidth]{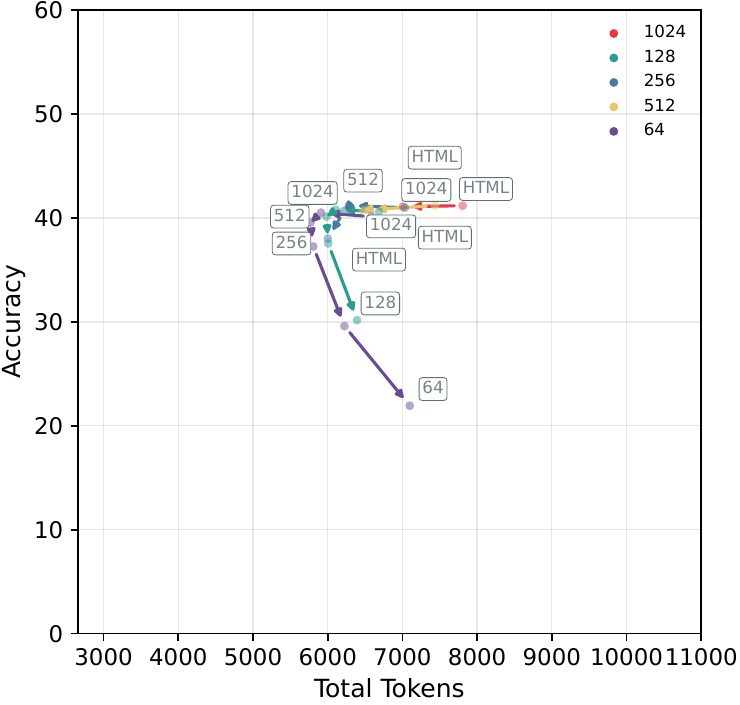}
  \caption{Accuracy against input tokens (left) and generated tokens (right) as a function of the budget at which the requested table is returned, for Qwen 3.5-9B on \mh{}. Colours distinguish the visual-token budget applied to the context, and each labelled point is the budget at which the requested table is returned (\textsc{html}, or images at 1{,}024 to 64 tokens). }
  \label{fig:app-tool-budget}
\end{figure}

The two stages of our method have independent budgets: the first compresses every table in the context, while the second returns the requested table at a chosen resolution. The main text fixes the return budget at 1{,}024 visual tokens, which amounts to no compression on our datasets. Figure~\ref{fig:app-tool-budget} justifies this choice on Qwen 3.5-9B, sweeping the return budget over \arr{\textsc{html}} and the five visual-token budgets for each of the five context budgets.

Accuracy holds up as long as the returned table is legible: \arr{\textsc{html}} and 1{,}024 visual tokens perform equally, and accuracy begins to degrade from 512 tokens onwards—slightly at first, then more clearly at the lower budgets. Generated tokens mirror the finding of Section~\ref{sec:qa}: returning the evidence at a low budget reintroduces the longer reasoning traces that compression induces, so the second stage loses the very property that motivates it. We therefore return the requested table at 1{,}024 tokens, which preserves accuracy and keeps reasoning traces short, and which on our datasets is equivalent to returning it uncompressed.

\section{\arr{Removing Table Identifiers}}
\label{sec:app-noid}

\begin{figure}[t]
  \centering
  \includegraphics[width=0.49\linewidth]{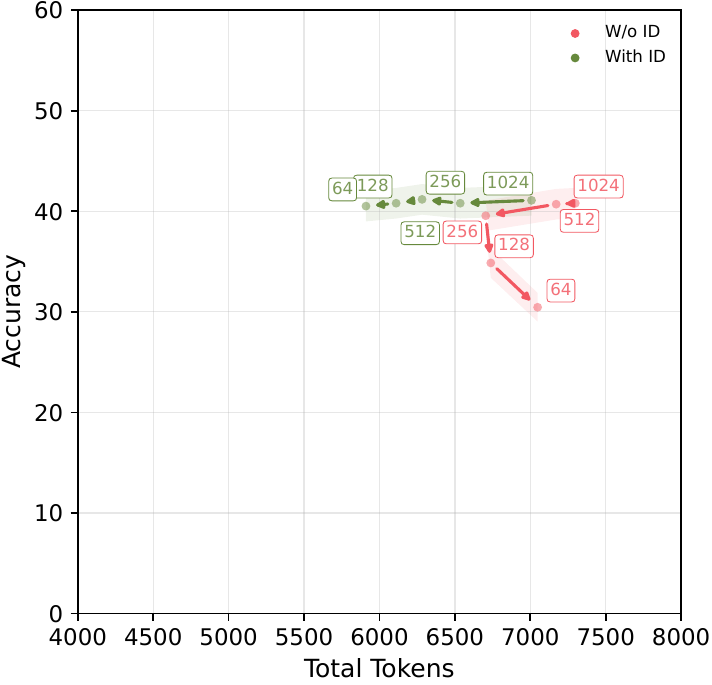}\hfill
  \includegraphics[width=0.49\linewidth]{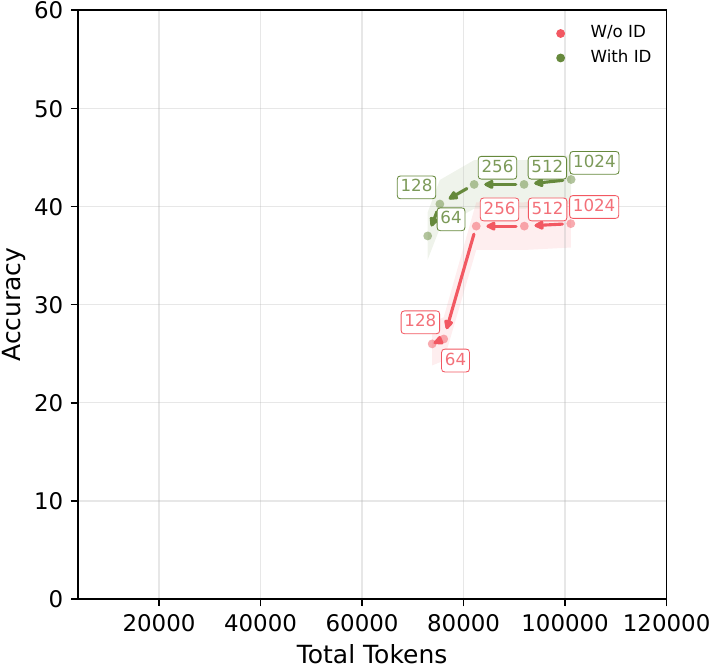}
  \caption{Accuracy against total tokens for our two-stage method with
  and without \texttt{[TABLE\_n]} identifiers, for Qwen 3.5-9B on \mh{}
  (left) and \fldqa{} (right). Markers correspond to visual token
  budgets.}
  \label{fig:app-noid}
\end{figure}

Our method precedes each table with a \texttt{[TABLE\_n]} marker, which
the model uses to request tables in the \arr{first stage}. Such markers come at
negligible cost: any pipeline that feeds tables as images must already
locate them and insert an image placeholder for each, and in
\textsc{html} table boundaries are explicit. To test how much the method
depends on them, we remove the markers and let the model refer to tables
by their order of appearance, leaving everything else unchanged. We run
this ablation with Qwen 3.5-9B on both datasets
(Figure~\ref{fig:app-noid}).

\paragraph{Identifiers matter most under heavy compression.}
On \mh{}, removing the identifiers barely affects accuracy at 512 and
1{,}024 tokens (0.1 and 0.3 points), and costs 1.6 points at 256. The
drop grows as the tables become harder to read, reaching 5.9 points at
128 tokens and 10.0 at 64. On \fldqa{}, the same pattern holds at a
higher level: the drop is around 4 points from 256 to 1{,}024 tokens
(4.2--4.6) and rises to 13.7 at 128 and 11.0 at 64. Unlike on \mh{},
the identifiers help even at native resolution. This is consistent with
Section~\ref{sec:perf-eff}, where restating the relevant tables matters
most in long documents, and with the difficulty of referring to a table
by its position among around 80 rather than three.

\paragraph{Identifiers do not add cost.}
Without identifiers, the model generates 37--82\% more tokens across
datasets and budgets. On \mh{}, this raises total cost at every budget,
from 4.1\% at 1{,}024 tokens to 19.3\% at 64. On \fldqa{}, where the
document text dominates, the extra generation offsets the roughly 680
input tokens saved by removing the markers, leaving total cost within
1.2\% of the original.

\paragraph{The method remains effective without identifiers.}
Even without identifiers, our method is more accurate than single-step
QA at every budget on both datasets: by 0.3 to 7.3 points on \mh{}
(30.5 vs.\ 23.2 at 64 tokens), and by 7.5 to 13.0 points on \fldqa{}
(26.0 vs.\ 14.8 at 64 tokens). It does so using fewer total tokens at
every budget except 512 on \mh{}, and within 1\% of the single-step
cost on \fldqa{}. The identifiers therefore strengthen the method,
particularly under aggressive compression and in long documents, but
its advantage over single-step QA does not depend on them.

\section{Retriever Comparison}
\label{sec:app-retrievers}

\begin{figure*}[t]
  \centering
  \includegraphics[width=0.58\linewidth]{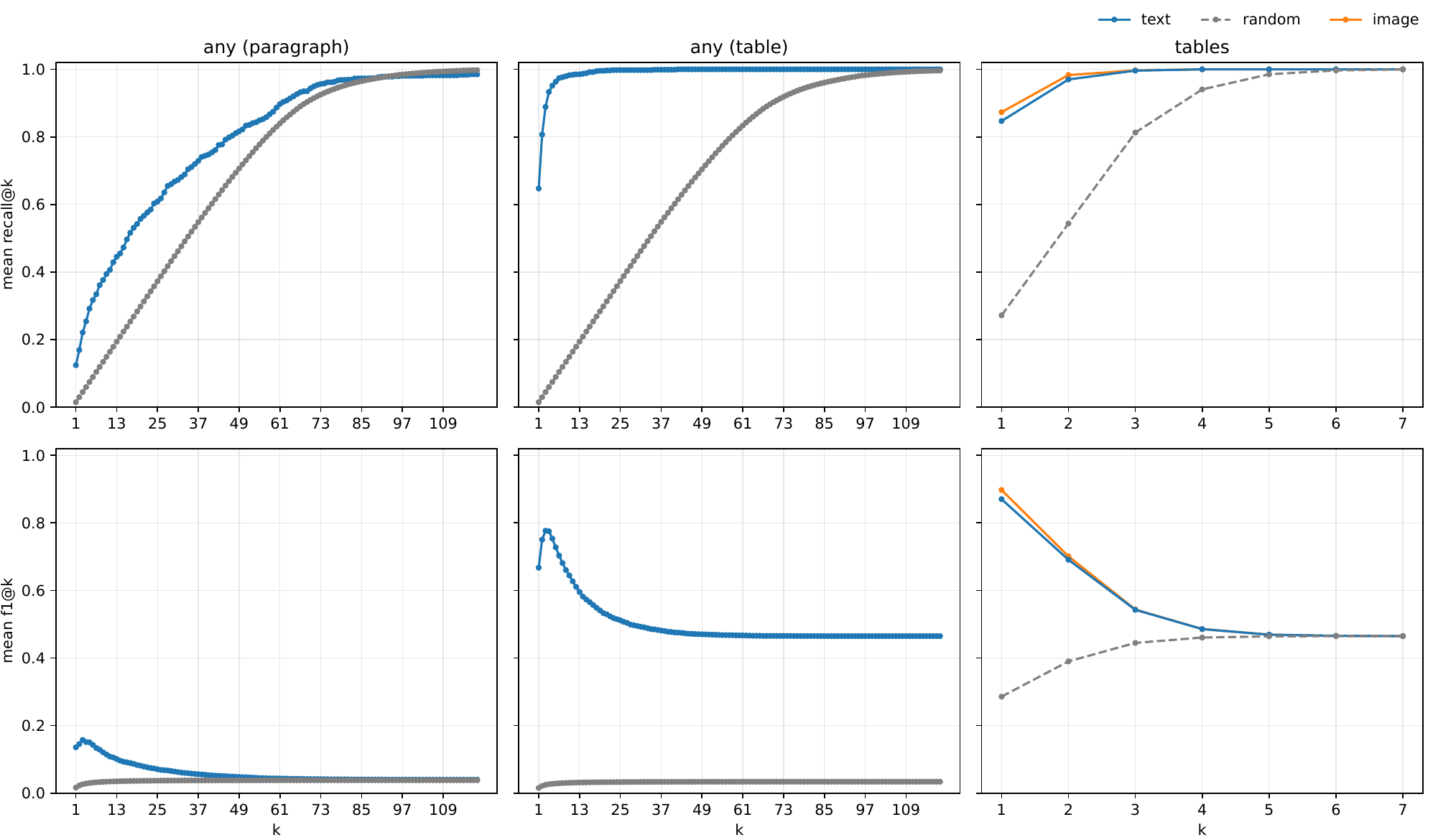}\hfill
  \includegraphics[width=0.40\linewidth]{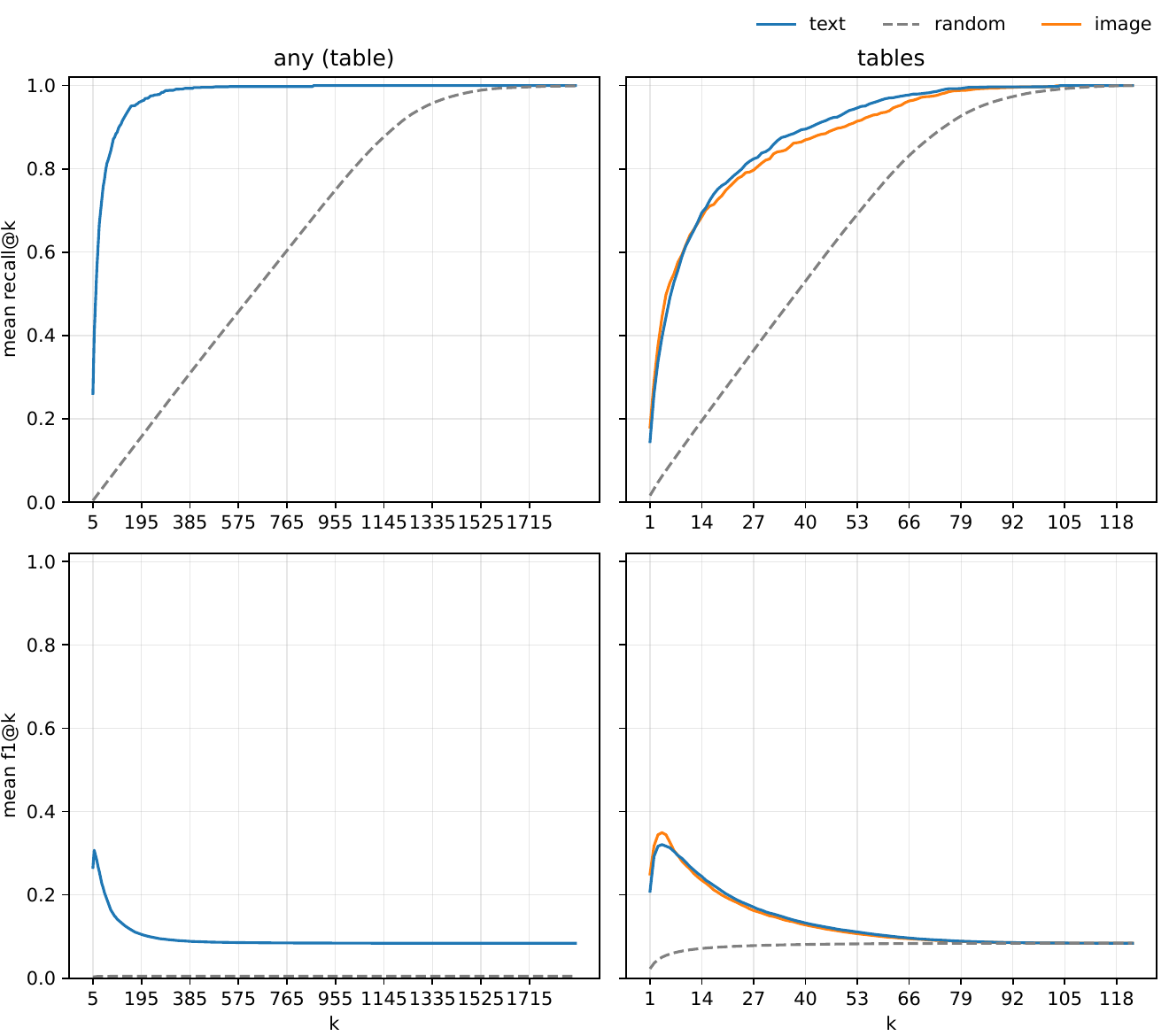}
  \caption{Retrieval performance on \mh{} (left) and \fldqa{} (right), reporting average recall (top row) and average F1 (bottom row) against the number of retrieved units. \emph{Text} is BGE-M3 and \emph{image} is ColQwen2. In \emph{any (paragraph)} and \emph{any (table)} the pool contains both paragraphs and tables, and only the text retriever applies; in \emph{tables} the pool contains tables alone and both retrievers are compared. \fldqa{} omits \emph{any (paragraph)}, as its evidence is annotated at the page level.}
  \label{fig:app-retrievers}
\end{figure*}

This appendix compares the two retrievers used in the paper and justifies where each is applied.
BGE-M3 \citep{chen-etal-2024-m3} indexes linearised text, while ColQwen2 \citep{faysse2024colpali} indexes rendered images, so only the former can operate over a pool that mixes paragraphs and tables.
Figure~\ref{fig:app-retrievers} reports average recall and F1 against the number of retrieved units, under three settings.
In \emph{any (paragraph)} and \emph{any (table)}, the pool contains both the document's paragraphs and its tables, and we measure how well the retriever recovers paragraph evidence and table evidence respectively; only BGE-M3 applies here.
In \emph{tables}, the pool contains tables alone, which allows both retrievers to be compared directly, BGE-M3 over their textual serialisation and ColQwen2 over their rendered images.

ColQwen2 outperforms BGE-M3 at retrieving tables from a table-only pool, which is why we use it in the table-retrieval setting of Section~\ref{sec:rag} and as the reference operating point in the identification results of Section~\ref{sec:identification-results}.
BGE-M3 is used wherever the pool mixes modalities, since ColQwen2 cannot index text.

On \fldqa{} we omit the \emph{any (paragraph)} setting, as the dataset annotates evidence at the page level rather than per passage.
For the same reason, we treat every table on an evidence page as an evidence table, which inflates the size of the evidence set and is part of why recall approaches 1.0 more slowly than on \mh{}; the far larger candidate pool, around 80 tables per document against three, accounts for the rest.

\section{Retriever for Identification}
\label{sec:app-colqwen}

\begin{figure*}[t]
  \centering
  \setlength{\tabcolsep}{0pt}
    \begin{tabular}{@{} c c c c @{}}
      & \multicolumn{2}{c}{\small \mh{}} & \multicolumn{1}{c}{\small \fldqa{}} \\
      \cmidrule(lr){2-3} \cmidrule(lr){4-4}
      & \small{Qwen 3 VL-8B} & \small{Qwen 3.5-9B} & \small{Qwen 3.5-9B} \\
      &
      \raisebox{-0.5\height}{\includegraphics[width=0.30\linewidth]{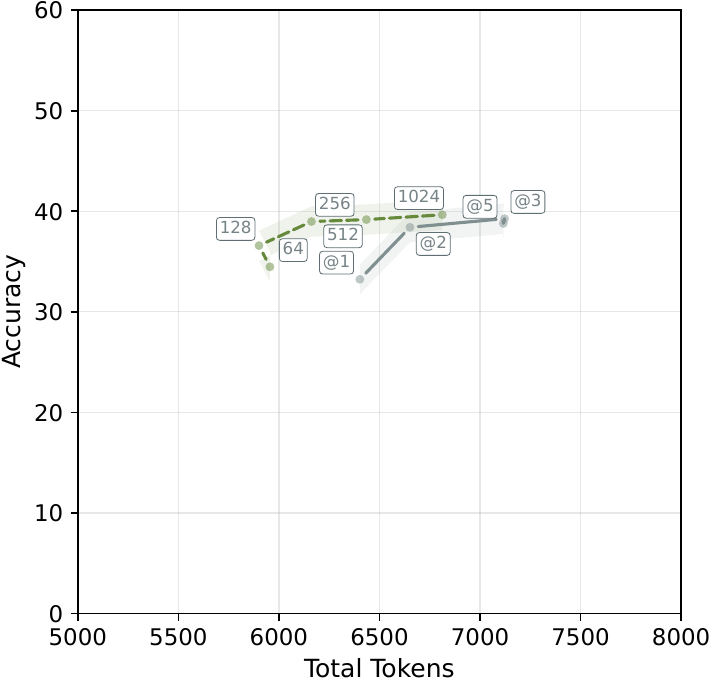}} &
      \raisebox{-0.5\height}{\includegraphics[width=0.30\linewidth]{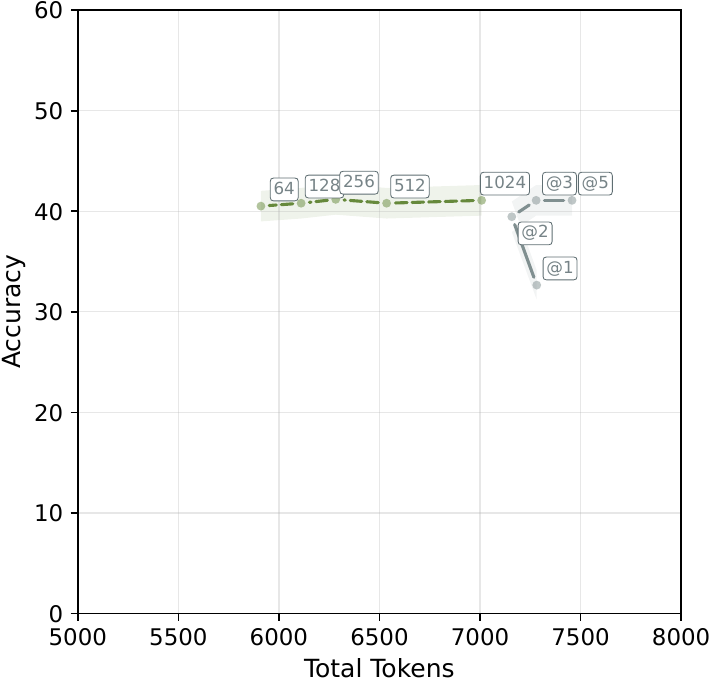}} &
      \raisebox{-0.5\height}{\includegraphics[width=0.30\linewidth]{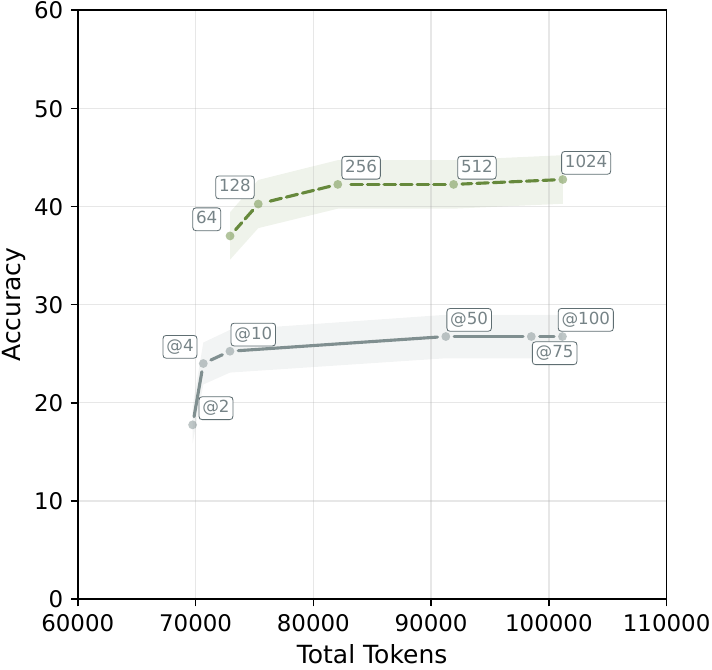}} \\[2mm]
      & \small{Gemma\,4 E4B} & \small{Gemma\,4 26B-A4B} & \small{Gemma\,4 26B-A4B} \\
      &
      \raisebox{-0.5\height}{\includegraphics[width=0.30\linewidth]{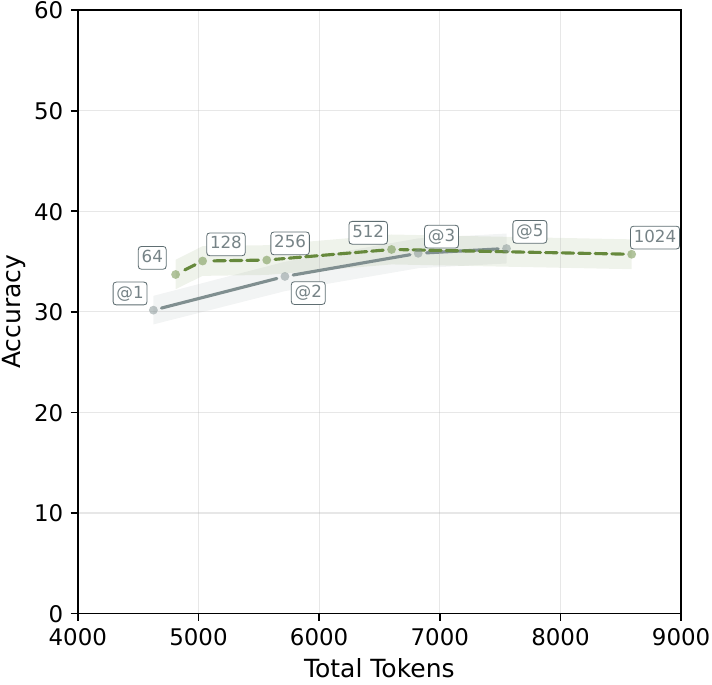}} &
      \raisebox{-0.5\height}{\includegraphics[width=0.30\linewidth]{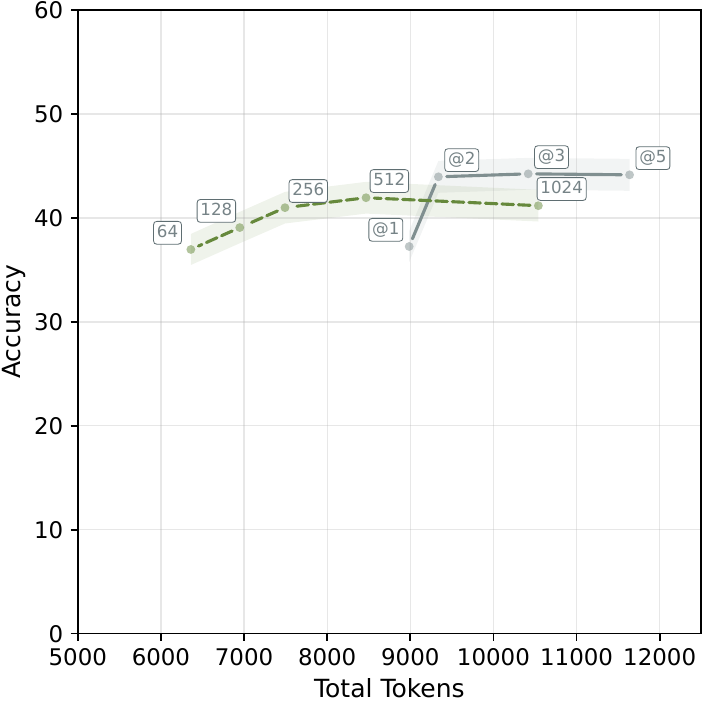}} &
      \raisebox{-0.5\height}{\includegraphics[width=0.30\linewidth]{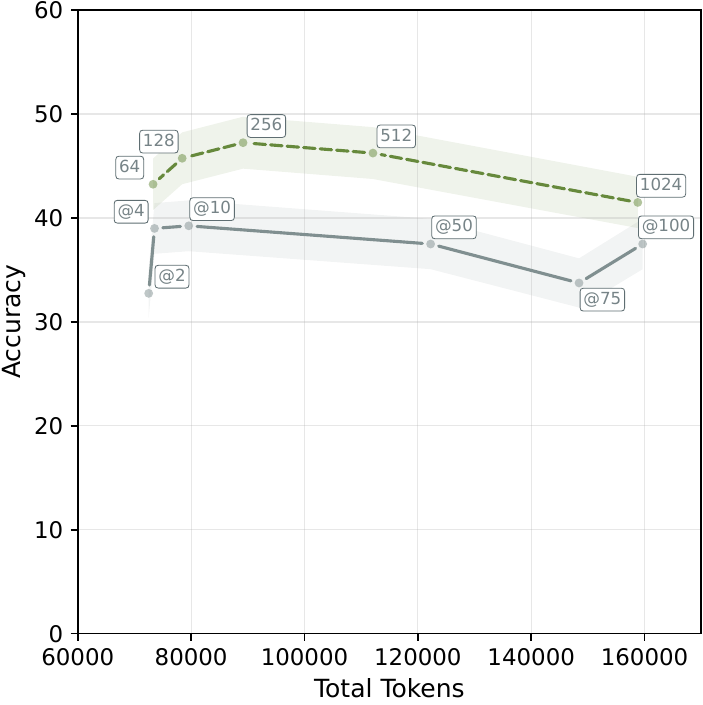}} \\
    \end{tabular}

  \vspace{1.5ex}
  \textcolor[HTML]{66893C}{\Large $\bullet$} \small{Two-Stage QA with model-based evidence identification (ours)} \hspace{2em}
  \textcolor[HTML]{808F90}{\Large $\bullet$} \small{ColQwen2 @k}

  \caption{Accuracy against total tokens for our two-stage method with model-based table identification, and for direct QA over a context in which the tables are supplied by ColQwen2 at increasing $k$, on \mh{} \arr{(left two columns)} and \fldqa{} \arr{(right column)}. Both conditions receive the full document text; they differ only in how the tables reach the model. Markers along our curves correspond to visual-token budgets, and bands show $\pm 1$~s.e.m.\ across examples (Appendix~\ref{app:ci}).}
  \label{fig:app-colqwen}
\end{figure*}

Section~\ref{sec:rag} asks whether a dedicated retriever can replace the identification stage of our method.
Figure~\ref{fig:app-colqwen} reports the full curves behind that comparison, \arr{for the Gemma and Qwen models on both datasets.}
In the retrieval condition the document text is fed in full, as in our hybrid setting, but the tables are supplied by ColQwen2 at increasing $k$ and left uncompressed, so the retriever assumes the role our first stage otherwise plays.

\section{\arr{Token Accounting Across Stages}}
\label{sec:app-token-accounting}

Single-step configurations run a single inference pass, so their cost
is the number of input tokens (textual and visual) plus the number of
generated tokens. Our two-stage method runs two passes, but the second
continues the first: the requested tables and the second-stage prompt
are appended to the first-stage conversation, which already contains
the document text, the compressed tables, the question, the
first-stage instructions, and the model's table request. We therefore
count
\begin{equation}
T_{\text{ours}} =
\underbrace{T^{(1)}_{\text{in}} + T^{(1)}_{\text{gen}}}_{\text{Stage 1}}
+
\underbrace{T_{\text{tab}} + T^{(2)}_{\text{prompt}} + T^{(2)}_{\text{gen}}}_{\text{Stage 2}},
\end{equation}
where $T^{(1)}_{\text{in}}$ is the first-stage input (document text,
compressed tables, question, and instructions), $T^{(1)}_{\text{gen}}$
the generated table request, $T_{\text{tab}}$ the visual tokens of the
requested tables at native resolution, $T^{(2)}_{\text{prompt}}$ the
second-stage instructions, and $T^{(2)}_{\text{gen}}$ the reasoning
and final answer. In the token breakdowns of
Appendix~\ref{sec:app-all-results}, $T^{(1)}_{\text{gen}}$ and
$T^{(2)}_{\text{gen}}$ are reported together as generated tokens.

Counting the shared prefix once corresponds to standard prefix
caching. Decoders use causal attention, so the key-value
(KV) states of each token depend only on the tokens preceding it.
Appending content to a conversation therefore leaves the KV states of
the existing prefix unchanged, and serving systems with prefix caching
reuse them instead of reprocessing the prefix. Only $T_{\text{tab}}$ and
$T^{(2)}_{\text{prompt}}$ require new prefill in the second stage.

Our implementation does not retain KV states between stages and
instead re-encodes the full conversation in the second pass. Since the
recomputed states are identical to cached ones (up to floating-point
nondeterminism), this affects runtime but not the model's outputs, the
reported token counts, or any of our conclusions. As in
Section~\ref{sec:budgets}, the cost of the visual encoder is not
included in these counts.

\section{\arr{Wall-Clock Time and Cost}}
\label{sec:app-runtime}

Decoder tokens (Section~\ref{sec:budgets}) are independent of hardware
and serving stack, but do not directly capture runtime. We therefore
measure the wall-clock generation time of Qwen 3.5-9B on \mh{}, for
direct QA and our two-stage method, at every visual-token budget and
with tables serialised as \textsc{html}. All runs use a single NVIDIA
H100 80GB HBM3 GPU and vLLM under identical settings; generation time
excludes model loading. Table~\ref{tab:runtime} reports total
generation time, the average time per example (total time divided by
the 1,044 examples), and the prefix-cache hit rate, i.e., the fraction
of prompt tokens whose KV states are served from vLLM's automatic
prefix cache instead of being recomputed during prefill.

\begin{table*}[t]
\centering
\small
\begin{tabular}{@{}l ccc ccc c@{}}
\toprule
& \multicolumn{3}{c}{\textbf{Direct QA}} & \multicolumn{3}{c}{\textbf{Two-stage (ours)}} & \\
\cmidrule(lr){2-4} \cmidrule(lr){5-7}
\textbf{Budget} & Gen.\ time & sec/ex & Cache & Gen.\ time & sec/ex & Cache & $\Delta$ sec/ex \\
\midrule
64           & 4:33:56 & 15.74 & 65.7\% & 1:00:36 & 3.48 & 78.8\% & $-$77.9\% \\
128          & 2:42:16 &  9.33 & 65.2\% & 1:00:40 & 3.49 & 78.8\% & $-$62.6\% \\
256          & 1:41:59 &  5.86 & 67.1\% & 0:57:37 & 3.31 & 78.8\% & $-$43.5\% \\
512          & 1:23:34 &  4.80 & 64.1\% & 0:47:50 & 2.75 & 78.5\% & $-$42.7\% \\
1024         & 1:20:05 &  4.60 & 64.3\% & 0:50:28 & 2.90 & 78.8\% & $-$37.0\% \\
\textsc{html} & 1:11:19 &  4.10 & 64.2\% & 1:00:45 & 3.49 & 76.6\% & $-$14.9\% \\
\bottomrule
\end{tabular}
\caption{Wall-clock generation time (h:mm:ss), average seconds per
example, and prefix-cache hit rate for Qwen 3.5-9B on \mh{} (1,044
examples) on a single NVIDIA H100 80GB HBM3 GPU. $\Delta$ is the
relative change in seconds per example of our method with respect to
direct QA at the same configuration.}
\label{tab:runtime}
\end{table*}

\paragraph{Our method is consistently faster.}
Our method reduces the time per example in every configuration, by
37.0\% at native resolution (1,024 tokens) and by up to 77.9\% at 64
tokens. The gap widens as the budget decreases because direct QA slows
down sharply under compression, from 4.60\,s per example at 1,024
tokens to 15.74\,s at 64, mirroring the longer reasoning traces
observed in Section~\ref{sec:qa}. Our method instead stays between
2.75\,s and 3.49\,s across all budgets, since the second stage always
reasons over native-resolution tables. Even its slowest configuration
is faster than the fastest direct QA configuration (\textsc{html},
4.10\,s), and its fastest (512 tokens, 2.75\,s) is 40.2\% faster than
direct QA at native resolution. The additional inference pass
therefore does not offset the savings: generation time is dominated by
decoding, and our method produces substantially shorter outputs
(Appendix~\ref{sec:app-all-results}).

\paragraph{Prefix caching.}
Direct QA already reuses a share of its prompt tokens, as prefixes
such as the instructions are shared across examples, and \mh{} poses
several questions over the same document. Our method raises the hit
rate from roughly 65\% to 79\%, reflecting that the second stage
continues the first and reuses the KV states of the shared prefix
(Appendix~\ref{sec:app-token-accounting}).

\section{\arr{Markdown Serialisation}}
\label{sec:app-markdown}

We serialise tables as \textsc{html} because it represents
hierarchical structure, such as multi-level headers and spanning
cells, without loss (Section~\ref{sec:settings}). Markdown is a more
compact alternative that flattens this structure.
Table~\ref{tab:app-markdown} compares the two serialisations with our
method for the open-weight models.

\begin{table*}[t]
\centering
\small
\begin{tabular}{@{}l rr rr rr rr@{}}
\toprule
& \multicolumn{2}{c}{\textbf{\textsc{html} QA}} & \multicolumn{2}{c}{\textbf{MD QA}} & \multicolumn{2}{c}{\textbf{MD \arr{two-stage}}} & \multicolumn{2}{c@{}}{\textbf{Ours (64)}} \\
\cmidrule(lr){2-3} \cmidrule(lr){4-5} \cmidrule(lr){6-7} \cmidrule(l){8-9}
\textbf{Model} & Acc. & Tok. & Acc. & Tok. & Acc. & Tok. & Acc. & Tok. \\
\midrule
\multicolumn{9}{@{}l}{\textbf{\mh{}}} \\
GLM-4.6V        & 40.1 & 5146  & 40.6 & 4281 & 42.2 & 4916 & 39.9 & 3938 \\
Qwen 3 VL-8B    & 38.5 & 8802  & 39.2 & 7859 & 40.1 & 7010 & 34.6 & 5950 \\
Qwen 3.5-9B     & 39.6 & 9946  & 40.0 & 9124 & 41.3 & 7260 & 40.5 & 5910 \\
Gemma\,4 E4B    & 35.5 & 5723  & 36.7 & 4896 & 36.7 & 6103 & 35.2 & 4996 \\
\arr{Gemma\,4 26B-A4B} & 44.2 & 9666  & 43.5 & 8442 & 40.4 & 7695 & 38.0 & 6430 \\
\midrule
\multicolumn{9}{@{}l}{\textbf{\fldqa{}}} \\
Qwen 3.5-9B     & 29.0 & 108016 & 28.5 & 94668 & 43.0 & 94056 & 37.0 & 72930 \\
\arr{Gemma\,4 26B-A4B} & 44.5 & 109496 & 49.2 & 94365 & 44.8 & 95537 & 38.8 & 73124 \\
\bottomrule
\end{tabular}
\caption{Accuracy and median total tokens per example with tables
serialised as \textsc{html} or Markdown (MD), and for our method with
tables compressed to 64 visual tokens. \emph{QA} is single-step
direct QA; \emph{\arr{two-stage}} applies our method with tables as
text.}
\label{tab:app-markdown}
\end{table*}

\paragraph{Markdown is cheaper than \textsc{html}, but not compressed images.}
Markdown uses 8--17\% fewer tokens than \textsc{html} at similar
accuracy (between $-$0.7 and \arr{$+$1.2} points on \mh{}, and $-$0.5 and
\arr{$+$4.7} on \fldqa{}). Compressed images, however, remain the more efficient
representation for Qwen 3.5-9B and GLM-4.6V: at 64 tokens, our
method matches or exceeds Markdown direct QA (between $-$0.7 and
$+$8.5 points) while using 8--35\% fewer tokens. The advantage is
largest on long documents, where Qwen 3.5-9B gains 8.5 points over
Markdown on \fldqa{} while using 23\% fewer tokens. For Qwen 3 VL-8B and the Gemma
models, the two representations offer a trade-off: Markdown direct QA
is more accurate, by 4.6 points for Qwen 3 VL-8B, 1.5 for Gemma\,4 E4B, and 5.5 (\mh{})
and \arr{10.4} (\fldqa{}) for Gemma\,4 26B-A4B, while our method
uses 24\% fewer tokens for Qwen 3 VL-8B and around 23\% fewer for Gemma\,4 26B-A4B.

\end{document}